\documentclass[manuscript,screen]{acmart}

\AtBeginDocument{%
  \providecommand\BibTeX{{%
    \normalfont B\kern-0.5em{\scshape i\kern-0.25em b}\kern-0.8em\TeX}}}

\setcopyright{acmcopyright}
\copyrightyear{2021}
\acmYear{2021}
\acmDOI{10.1145/1122445.1122456}

\acmConference[Woodstock '18]{Woodstock '18: ACM Symposium on Neural
  Gaze Detection}{June 03--05, 2018}{Woodstock, NY}
\acmBooktitle{Woodstock '18: ACM Symposium on Neural Gaze Detection,
  June 03--05, 2018, Woodstock, NY}
\acmPrice{15.00}
\acmISBN{978-1-4503-XXXX-X/18/06}

\usepackage{xcolor}
\usepackage{cite}
\usepackage{geometry}
\usepackage{graphicx}
\usepackage{pdflscape}
\usepackage{dirty talk}
\usepackage{hyperref}
\usepackage{booktabs}
\usepackage{multirow}
\usepackage{wrapfig}
\usepackage{longtable}

\newtheorem{definition}{Definition}

\definecolor{green}{RGB}{3,200,15}

\begin{document}

\title{Conflict Avoidance in Social Navigation - a Survey}

\author{Reuth Mirsky}

\author{Xuesu Xiao}

\author{Justin Hart}

\author{Peter Stone}

\renewcommand{\shortauthors}{Mirsky et al.}

\begin{abstract}
  A major goal in robotics is to enable intelligent mobile robots to operate smoothly in shared human-robot environments.  One of the most fundamental capabilities in service of this goal is competent navigation in this ``social" context.  As a result, there has been a recent surge of research on social navigation; and especially as it relates to the handling of conflicts between agents during social navigation. These developments introduce a variety of models and algorithms, however as this research area is inherently interdisciplinary, many of the relevant papers are not comparable and there is no shared standard vocabulary.
    This survey aims to bridge this gap by introducing such a common language, using it to survey existing work, and highlighting open problems. It starts by defining the boundaries of this survey to a limited, yet highly common type of social navigation -- conflict avoidance. Within this proposed scope, this survey introduces a detailed taxonomy of the conflict avoidance components. This survey then maps existing work into this taxonomy, while discussing papers using its framing. Finally, this paper proposes some future research directions and open problems that are currently on the frontier of social navigation to aid ongoing and future research.
\end{abstract}



\maketitle

\section{Introduction}
\label{sec:intro}
Enabling autonomous robots to navigate in the presence of people and/or other robots has been studied for the past 70 years. One of the first examples of social navigation is Grey Walter's work, who built robotic ``turtles'' that could navigate on their own \citep{walter1950imitation}. These robots, named Elmer and Elsie, were an exercise in minimalism and demonstrated that a small number of brain cells could give rise to complex behaviors. They each consisted of ``two miniature radio tubes, two sense organs, one for light and the other for touch, and two effectors or motors, one for crawling and the other for steering''. Their power supply was a hearing-aid battery. Nevertheless, these robots could navigate freely in an enclosed space and change their trajectory in response to light and touch.

Modern mobile robots are much more sophisticated and complex. Most feature a variety of sensors, intricate steering systems, and several layers of hardware and software to control their movement. Despite these improvements, mobile robots are still not prevalent in our homes and offices. One of the main reasons for this deficit is that comprehensive autonomy is still achievable only in controlled environments and is usually induced by hard-coded rules or learned from a relatively clean dataset \citep{cassandra1996acting,kaelbling2020foundation, siegwart2011introduction}.
The problem of navigation in the presence of other robots and humans is complex and cross disciplinary in nature. Solutions draw from robotics, artificial intelligence, engineering, psychology, biology, and other areas of study. As such, each of these communities has defined social navigation differently. In the multi-robot community \citep{verma2021multi}$^{R1}$, 
social navigation usually refers to robot navigation in the presence of additional robots. In human-robot interaction (HRI), social navigation refers strictly to the task of navigating in a shared space with people.
\citet{rios2015proxemics} gave a compact description of socially-aware navigation:
\emph{Socially-aware navigation is the strategy exhibited by a social robot which identifies and follows social conventions (in terms of management of space) in order to preserve a comfortable interaction with humans. The resulting behavior is predictable, adaptable, and easily understood by humans. This definition implies, from the robot’s point of view, that humans are no longer perceived only as dynamic obstacles but also as social entities.}

In the general social navigation setting, a social agent is an agent (either human or robot) that is aware of the objectives of others (human or robot) and considers them in its behavior, either by adjusting its policy or by indicating why it chose a potentially “anti-social” behavior. This general definition is quite broad, encompassing a wide variety of multi-agent navigation scenarios, including those that involve only robots. In practice, the term ``social navigation'' usually refers to a more human-centric perspective. Thus, this survey focuses on three requirements that separate human-centric social navigation from more general social navigation. These requirements are:
\begin{enumerate}
    \item There exists an autonomously navigating agent. The agent has a specific, reachable navigational goal.
    \item There exists one (or more) humans or animals in the environment.
    \item 
     The interaction takes place in the real world (either a controlled or natural environment), not in simulation.
\end{enumerate}

Many papers have discussed challenges that occur when only one or two of these requirements are met. Teleoperation of robots is widely investigated within HRI, but it is not consistent with (1). The multi-agent systems (MAS) and distributed planning communities focus on constructing algorithms for multi-robot navigation, which do not meet requirement (2). Even within the HRI community, many works describe progress in social navigation in simulations rather than in real world environments, so that requirement (3) does not hold. Significant work has been done in the graphics community to model crowds and swarms, but these works also do not meet requirement (3). 
 Our main focus is on papers that meet all three requirements. This survey also cites some papers for which not all of the above requirements hold, due to their contributions to our understanding of social navigation. In cases where the underlying scope of a paper is not fully aligned with this survey, we indicate the requirements that do hold on the first occasion that the paper is referenced. For example: \citet{walter1950imitation}$^{R1,R3}$ is a work in which there is an autonomous agent (R1) --- a mechanical ``turtle'' that navigates in the wild (R3) --- but no human pedestrians are present (R2). 

Even within the context of the three requirements discussed above, there are many behaviors that could be considered ``social'': following, giving navigational instructions, waiting in line, and others; as discussed later in this section. To limit the scope of this survey, we focus on one specific type of social interaction with people which requires the robot to reason about an encounter, specifically \emph{conflicts}. A conflict is a short-term interaction between a robot and a human in which there is a chance that the robot and the human will collide. Note that this potential event can be objective, meaning that if no party changes its course they will collide; or it could be that the passing of the robot is perceived as being on a collision course by the human. Additionally, not all interactions in social navigation are conflict avoidance. For example when a robot is designed to carry a person's luggage and follow them, the task is a social navigation task in which the robot needs to detect the person, reason about the proper distance from them, and drive at a safe and comfortable speed. These challenges, however, are orthogonal to the challenge of avoiding conflicts with other pedestrians. 
%
Understanding conflicts in social navigation requires a definition of what a conflict is in this context:

\begin{definition}
A \textbf{conflict} between a robot and other mobile robots or pedestrians is a situation in which if there is no change of direction or a change in speed by at least one of the parties, they will collide.
\end{definition}

By this definition, not all conflicts end in a physical collision, but every collision is preceded by a conflict. Moreover, as the interacting parties can falsely predict an upcoming collision (e.g. a human feels that the robot will come too close and is risking a collision), the presence of a conflict is a subjective matter that depends on the interpretation of the interacting parties.
%
This survey is not the first to identify navigational conflicts as being separate from collisions. A footnote from \citet{van2011reciprocal} implies a difference between reasoning about conflicts in motion planning and avoiding collisions:
\addtolength\leftmargini{-0.1in}\begin{quote}
    Note that the problem of (local) collision-avoidance differs from motion planning, where the global environment of the robot is considered to be known and a complete path towards a goal configuration is planned at once, and collision detection, which simply determines if two geometric objects intersect or not.
\end{quote}
However, in \citet{van2011reciprocal} they do not elaborate on this idea. 
Based on this scope, the contributions of this survey are as follows.
\begin{enumerate}
    \item It surveys work in which the authors include conflict avoidance in their models. 
    \item It introduces a taxonomy of the attributes that vary between models and algorithms for conflict avoidance.
    \item Based on this taxonomy, it identifies the attributes of existing works and categorizes these works into tables.
    \item It summarizes the current state of the art in conflict avoidance in social navigation, including a practical checklist to follow when introducing a new contribution the body of literature.
\end{enumerate} 

Previous works have presented ideas that overlap those in this survey, but from different perspectives. There are surveys on topics relating to social robotics \citep{fong2003survey}$^{R1,R2}$; and to numerous related navigation topics such as: path planning \citep{cai2020mobile}$^{R1}$, vision for navigation \citep{desouza2002vision, bonin2008visual}$^{R1}$, perception and semantics \citep{garg2020semantics} and localization and mapping \citep{gaber2017localization, crespo2020semantic, shit2020precise}$^{R1,R3}$. There are also many surveys on social navigation that focus on elements such: as joint or group navigation \citep{moussaid2010walking, yao2019following, prassler2002key, karunarathne2018model}, giving navigational instructions \citep{thrun2000probabilistic, yedidsion2019optimal}$^{R1}$, detecting dynamic objects \citep{ess2009moving, kit2012change}$^{R1}$, social contexts such as waiting in line \citep{nakauchi2002social}$^{R1,R2}$ or distributing flyers \citep{shi2018robot}$^{R1,R2}$, and other factors which are not discussed in this survey \citep{pirk2022protocol}. None of these surveys, however, focus specifically on assisting in detecting or avoiding conflicts. Here we provide details on the major related surveys, both to provide a reference for readers who are interested in those different points of view and to define the scope of this survey.

\citet{kruse2013human}$^{R1,R2}$ highlight a rising interest in the topic of social navigation since 2000, and identify specific tasks and challenges that social navigation encompasses. Interest is still on the rise, meaning that there are many new works on this toplic; requiring this survey to narrow its focus somewhat as we update their coverage of the topic. Our focus is on the narrower topic of conflicts that arise between robots and pedestrians. \citet{hoogendoorn2003simulation}$^{R2,R3}$ introduced a three-tiered model of navigation utility, decomposing it into strategic (high-level decision making), tactical (global navigation), and operational (local navigation and event handling) levels. This survey focuses mostly on the operational level: setting local goals and re-planning as needed. Recently, \citet{gao2021evaluation} provided a review of scenarios, datasets, and methods used in social navigation. They described the main use-cases as: passing, crossing, overtaking, approaching, following, leading, accompanying, and combinations thereof. Our survey's perspective is different in that it does not categorize papers according to the aim of the navigating parties, but rather according to situations in which these parties are (or will be) in conflict. In this sense, \citet{gao2021evaluation} review a wider set of social navigation tasks, though they do not propose a taxonomy of conflicts as introduced in this survey.

\citet{charalampous2017recent} present a survey in which they aim \emph{``to systemize the recent literature by describing the required levels of robot perception, focusing on methods related to a robot’s social awareness, the availability of datasets these methods can be compared with, as well as issues that remain open and need to be confronted when robots operate in close proximity with humans.''}
This survey extends their initial discussion on robot design for operation in close proximity to humans; or as we refer to it, robots in conflict situations. Specifically, we aim to provide basic definitions to be used to standardize future works on the problem of robots that navigate in close proximity to people. 
\citet{lopez2019walking}$^{R1,R3}$ provide a survey on turn prediction and how upper body kinematics can signal upcoming turns. In their survey, they identified that Gaze Yaw is the earliest predictor of walking turns; but that existing data do not support 
quantifying how much --- or how reliably --- timing and distance can be anticipated. They found, however, that Head Yaw was the most reliable kinematic variable for predicting walking turns about 200ms from commencing to turn. Their survey can inform the design of conflict resolution by enabling the robot to predict upcoming turns using these signals. Another recent survey focuses on algorithmic requirements and methodologies for robot navigation \citep{moller2021survey}. Their survey revolves mostly around perception and trajectory modeling rather than actuation. While the authors mention collision avoidance as an important robot navigation task; they do not focus their survey around collision avoidance, as presented here.

The survey by \citet{xiao2020motion} reviews methods that use machine learning techniques for the general problem of mobile robot navigation. Their survey focuses on the comparison between machine learning and classical approaches in terms of their scope and performance on real-world navigation problems. In contrast, this survey is on social navigation, with focusing specifically on conflict avoidance, and the papers may use any (learning or non-learning) method in approaching the problem.
For a more general perspective on the current state of social navigation, \citet{mavrogiannis2021core} identified three broad themes that are being investigated: planning, behavior design, and evaluation. These themes impact all social navigation tasks rather than being specific to conflict avoidance, and thus their discussion does not focus on this aspect. This survey is more specific to the context of collision avoidance in social navigation, and it drills down to provide an elaborate taxonomy of models and algorithms for such scenarios.

The remainder of this survey is organized as follows: Section \ref{sec::taxonomy} proposes a taxonomy for social navigation, identifying important factors of the social navigation problem. Sections \ref{sec::models} and \ref{sec::algorithms} present a selection of relevant works that have contributed models and algorithms, respectively. Section \ref{sec::evaluating} focuses on the evaluation metrics used in social navigation and refers to some existing benchmarks. Finally, Section~\ref{sec::discussions} highlights open problems in social navigation with respect to the proposed taxonomy and provides a checklist for researchers to consult when investigating a new social navigation problem.



\section{Taxonomy}
\label{sec::taxonomy}

\begin{figure}
    \centering
    \includegraphics[width=0.9\textwidth]{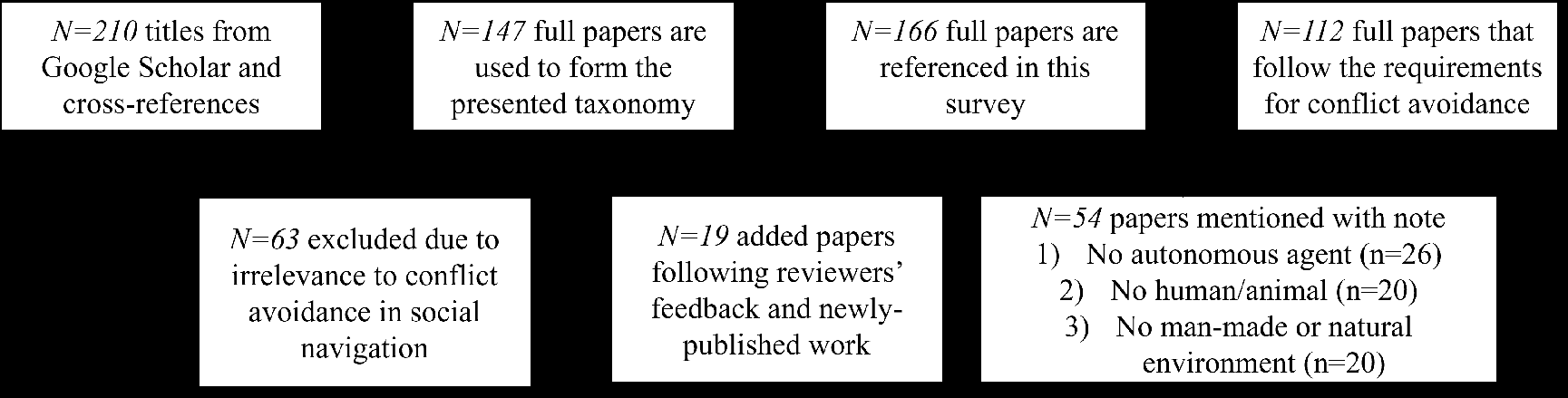}
    \caption{Study flow diagram showing the inclusion methodology used in this survey.}
    \label{fig:methodology}
\end{figure}

This section systematically describes a taxonomy used in this survey to categorize social navigation models (Section \ref{sec::models}) and algorithms (Section \ref{sec::algorithms})
.  
Here we describe the process used to collect the papers used in this survey. We started with existing surveys on social navigation \citep{charalampous2014social,kruse2013human} and we collected all of their references, as well as papers that cite these works using Google Scholar. In selecting which papers to include, we used the criteria specified in Section \ref{sec:intro} to guide the process. Overall, this survey contains 54 (out of 166) citations that do not meet all three criteria outlined in Section \ref{sec:intro}, but which nonetheless provide fundamental contributions to out understanding of the social navigation problem; or which are surveys on topics relevant to social navigation. We iterated through the process of collecting papers that cite, and are cited by, our current bibliography, until doing so yielded no new papers meeting all of the outlined requirements. The only exception to this process is when several papers have been published by the same group. Research groups often publish multiple papers on the same project. In these cases we include more than one paper if they are categorized differently by our taxonomy. Otherwise, we include only the most recent paper. Figure \ref{fig:methodology} summarizes the paper selection process for this survey.

For each of the resulting 112 papers on conflict avoidance in social navigation, we identify seven attributes, listed in Table~\ref{tab:taxonomy}. 
Below we discuss this list of attributes (in bold) and the values (in italics) they can take.  (Abbreviations for many values are used in tables in Sections \ref{sec::models} and \ref{sec::algorithms}. These abbreviations appear in parentheses next to their corresponding value.). We acknowledge that not all papers can be situated precisely within this taxonomy. 
In these cases, or if the value is not stated in the relevant paper, we label the corresponding attribute with the value ``None'' or ``Neither'' (e.g. some of the papers do not provide any empirical analysis, and thus the experiment type attribute is ``None''). This taxonomy is constructed with the goal of encompassing as much work as possible, such that any new contribution can be easily placed in a clear context.

\begin{table}[h]
\small
\centering
\caption{The Social Navigation Taxonomy}
\begin{tabular}{cc}
\toprule
Attributes                         & Values                                                                                                                                  \\ \midrule \midrule
\textbf{Robot Role}                & \textit{Reactor (R) / Initiator (I) / Both (B) / Neither (N)}                                                                                           \\ \hline
\textbf{Number of Agents}          & \textit{Absolute Number ($Abs=\#$ of agents)  / Density ($D=\#/m^2$)}                                                                                                                 \\ \hline
\textbf{Observability}             & \textit{Full / Partial / Depth / RGB}                                                                                                   \\ \hline
\textbf{Motion Control}            & \textit{SFM / ORCA / ROS / Human / Other}                                                                                               \\ \hline
\textbf{Communication}             & \textit{None (N) / Indirect (I) / Direct (D)}                                                                                           \\ \hline
\textbf{Experiment Type}           & \textit{Simulation (Sim) / In the Lab (Lab) / In the Wild (ItW) / Survey (Sur)}                                                                              \\ \hline
\textbf{Agent Type}                & \textit{\begin{tabular}[c]{@{}c@{}}Human-Robot (H-R) / Human-Agent (H-A) / Human-Human (H-H) / \\ Robot-Robot (R-R) / Homogeneous Agents (Hom) / Heterogeneous Agents (Het) \end{tabular}} \\ \bottomrule
\end{tabular}
\normalsize
\label{tab:taxonomy}
\end{table}

\subsection{Taxonomy Attributes and Values}
Some of the attributes and the values presented here are not intuitive. Here, we explain their rationale. 

\begin{figure}
    \centering
    \includegraphics{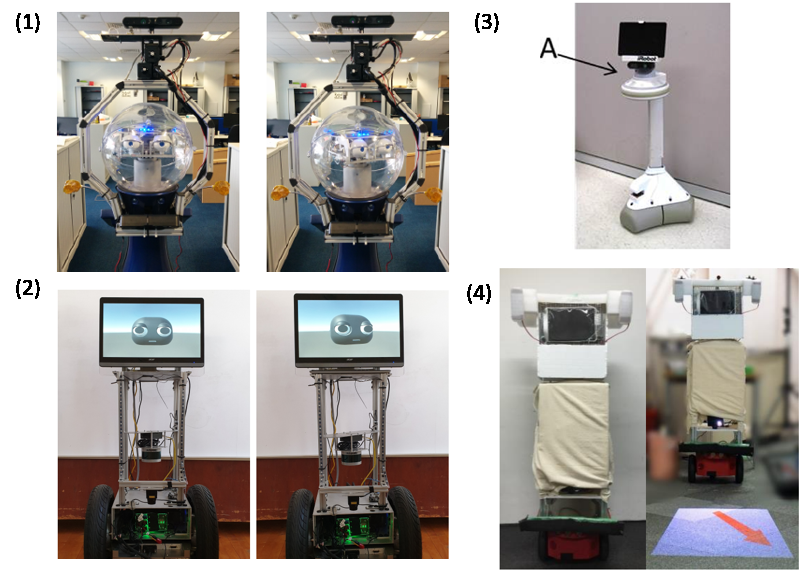}
    \caption{Various direct communication behaviors: (1) mechanical gaze \citep{may2015show}, (2) virtual gaze \citep{hart2020hallway}, (3) sensor rotation \citep{fiore2013toward}, (4) arrow signaling \citep{shrestha2018communicating}. }
    \label{fig:direct_com}
\end{figure}

\begin{description}
    
    \item [Number of Agents] \emph{Absolute Number (Abs) / Density (D)}. Some papers deal with a one-on-one interaction whereas others deal with multiple agents in a shared space. We mention, when known, how crowded the environment is. Most works report either an \emph{Absolute Number} of participants or a \emph{Density} (measured as $\# people / m^2$). 
    When presenting an absolute number of pedestrians, we include the navigating robot in the count. This allows comparison with multi-robot research where the number of agents includes multiple robots that are running the same algorithm. 

    \item [Observability] \emph{Full / Partial / Depth / RGB}. If the work is set up in simulation, the robot can have either full or partial observability. Work that involves experiments or evaluations with real robots usually reports specific type(s) of sensors that were used, such as depth sensors (e.g. LIDAR), or cameras (e.g. RGB, or RGBD). If more than one type of sensor is used, we mention all of them.
    
    \item [Motion Control] \emph{SFM / ORCA / ROS / Human / Other}. Most robots in these papers rely on an existing motion controller, and the robot is augmented with a new component for social navigation. This survey classifies the main types of motion control used in these papers: the Social Force Model (\emph{SFM}), Optimal Reciprocal Collision Avoidance (\emph{ORCA}), the ROS \texttt{move\_base} navigation stack (\emph{ROS})\footnote{\url{https://www.ros.org/}}, evaluation of human behavior without any existing robot (\emph{Human}), and \emph{Other}. The ``other'' category includes both papers in which the motion control is not significant (such as research projects that use cellular automata, point-based navigation, Dijkstra's algorithm, or other types of search for motion planning), and in which the motion control is novel and is a major part of the paper's contribution (such as Social Momentum \citep{mavrogiannis2018social} or LM-SARL \citep{chen2019crowd}).  We mention the specific motion control that is used when possible.

    \item [Communication] \emph{None (N) / Indirect (I) / Direct (D)}. This attribute refers to communication that is conveyed by the robot, and not to communication that is conveyed by the other agents. 
    \emph{None} means that the robot is not doing anything specifically to convey its navigational goal. \emph{Indirect} communication refers to situations where the robot uses whatever mechanisms it already possesses to signal its intentions, such as legibility~\citep{dragan2013legibility}$^{R1}$ and stigmergy~\citep{bonabeau1999swarm}$^{R2}$. \emph{Direct} communication means that there is some mechanism that is added to the robot to allow communication. See Figure \ref{fig:direct_com} for examples.
    
    \item [Experiment Type] \emph{Simulation (Sim) / Laboratory (Lab) / In the Wild (ItW) / Survey (Sur)}. Many researchers run experiments in \emph{Simulation} as part of their evaluation, either as the only type of evaluation or in addition to real-world experiments. \emph{Laboratory} experiments are defined as experiments in the real world in a controlled environment such as a laboratory or using a scripted scenario. \emph{In the Wild} are real world experiments in an unstructured environment or with no predefined script for the pedestrians. All of these types of experiments can be accompanied by post-interaction \emph{Surveys}. When a paper reports on more than one type of experiment, we include the details of one experiment, ordered in this prioritized order: In the Wild, Laboratory, Simulation, Survey (when more than one methodology used). There are two exceptions to this policy: the first exception regards surveys, which are often used as an additional metric for an experiment in the wild or in the laboratory. Thus, if an experiment is accompanied by a survey, the survey is also mentioned.
    The second exception regards papers that report two or more experiment types, where one of them is a small-scale in the wild experiment that does not report significant results. In such cases, we report the paper according to the experiment with reported results, but add a superscript $+$ symbol next to it to indicate that the paper also includes an in the wild experiment (e.g. Lab$^+$).   
        
    \item [Agent Type] \emph{Human-Robot (H-R), Human-Agent (H-A), Human-Human (H-H), Robot-Robot (R-R), Homogeneous Agents (Hom), Heterogeneous Agents (Het)}. This survey focuses on social navigation involving a person and a robot (\emph{Human-Robot}). Due to the difficulty of evaluating such interactions, many models and algorithms are evaluated on a different set of agents. The most common approaches are running a simulation in which the human counterparts are controlled by a real human (\emph{Human-Agent}) or by some other set of predefined or learned behaviors (either \emph{Homogeneous Agents} or \emph{Heterogeneous Agents}). Several included papers provide a fundamental understanding of human navigation and present evaluations that do not involve robots at all (\emph{Human-Human}) or that do not involve humans (\emph{Robot-Robot}). These papers are cited using the notation presented earlier (e.g. citation$^{R1}$), highlighting that they do not satisfy one or more of the inclusion criteria.
    
\end{description}

\textbf{Observability} is important to consider, especially when discussing simulations. Simulations explicitly model the observations that can be made by agents acting in a scene. Many simulations assume that a robot (or pedestrians around it) has full (ground truth) observability. Other simulations restrict observability in artificial ways, attempting to emulate realistic sensing capabilities (partial observability). In the discussion of these papers, it is important to note that some observation modalities may be unrealistic to implement on real robots. The conclusions of such papers may not translate to the context of real-world embodied social navigation. 

With respect to the \textbf{Communication} attribute, we make a distinction between communication that is \textit{indirect} or \textit{direct} and communication that is implicit or explicit. Implicit communication is often used to describe any non-verbal communication that is conveyed by people (e.g. the interpretation of eye gaze is implicit), and explicit communication is performed specifically with the intention of communicating with others (e.g. speech is explicit) \citep{cui2020empathic}$^{R2,R3}$. Robots do not generally naturally communicate implicitly (for example, not all robots have ``eyes'' and those that do do not necessarily need to turn them to ``look'' at something, or reflexively turn them to where they are about to navigate). As such, we make the distinction between \textit{direct} and \textit{indirect} communication as defined above, and keep the implicit/explicit distinction as one reflective of mimicking human behavior. Using these definitions, the possible combinations for robot communication are: implicit-indirect (e.g. velocity change \citep{van2011reciprocal}$^{R1}$), implicit-direct (e.g. gaze change on a virtual head \citep{admoni2011robot}), and explicit-direct (e.g. arrow projections on the floor \citep{watanabe2015communicating}).

Some papers present more than one set of experiments. An example would be presenting both a \textit{Human-Robot} laboratory study and a simulation of \textit{Heterogeneous agents}. We choose to highlight \textit{Human-Robot} experiments; a particularly relevant format for studies in social navigation. In general, for papers that present more than one set of experiments we categorize them by the values most relevant to \textit{Human-Robot} interaction on the attributes of \textbf{Experiment Type} and \textbf{Agent Type}: $\emph{In the Wild} > \emph{In the Lab} > \emph{Simulation} > \emph{Survey}$. 

\begin{figure}
    \centering
    \includegraphics[width=0.8\textwidth]{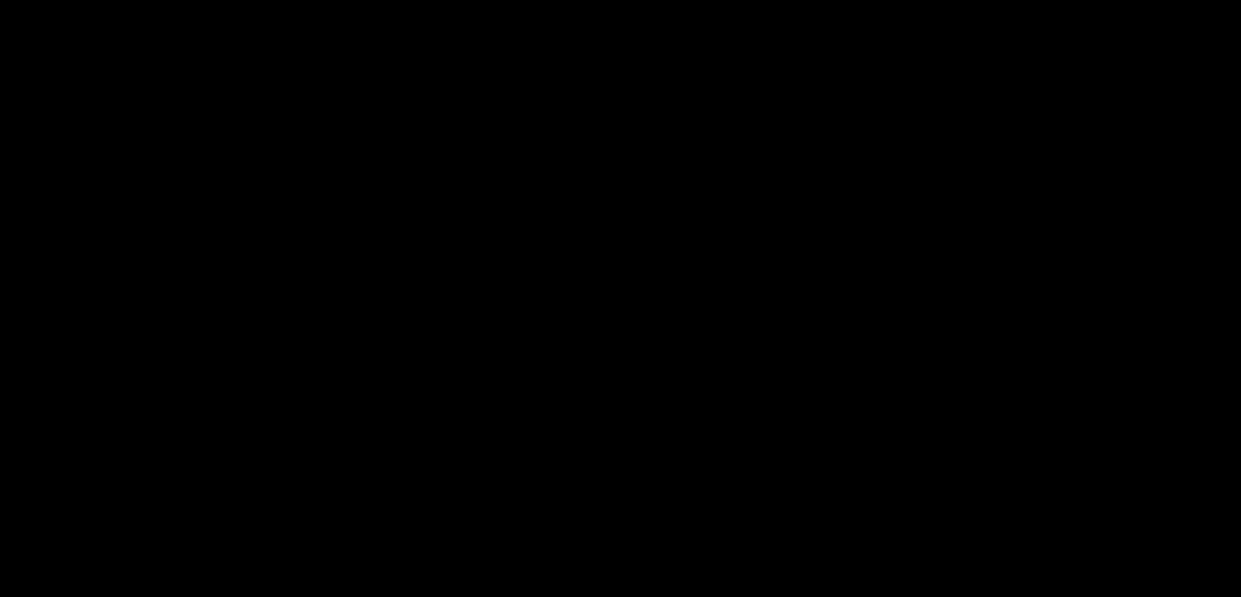}
    \caption{The hierarchical structure of the taxonomy's attributes}
    \label{fig:hierarchy}
\end{figure}

We also highlight that some of the taxonomy attributes are very concrete and define low-level components used in the interaction (e.g. the motion control used), while other attributes are more abstract (e.g. robot role). Usually, the abstract attributes and their values depend on the concrete attributes. Figure~\ref{fig:hierarchy} presents the hierarchical structure of these attributes, in work that is consistent with the three requirements outlined in Section \ref{sec:intro}. The bottom part represents the attributes that are independent of other attributes. The values assigned to the attribute at the end of an edge affect the values that can be assigned at its origin. For example, the values of the \textbf{Communication} attribute will be directly affected by the \textbf{Number of Agents} in the environment and the robot's \textbf{Observability}. In turn, the choice of value for the \textbf{Communication} attribute directly affects 
the \textbf{Agent Type}s that can perceive the chosen communication channel.

\subsection{Additional Concepts}
\label{sec::taxonomy:extraConcepts}
There are some additional concepts that are worth mentioning, but which we decided to exclude from our taxonomy. Here we list these concepts and explain why they are not included in the taxonomy. As research and discussion on social navigation progresses, this taxonomy could be extended to include these attributes.

One seemingly-important factor to consider in the taxonomy is \textbf{collision type}. When referring to collisions, most papers describe head-on collisions or side-on collisions, with rear-end collisions as the least commonly investigated type. Among the papers in this survey, none explicitly discuss only one type of collision. There are several papers that propose ways to categorize collisions according to the required response from pedestrians and / or the robot. \citet{reynolds1999steering} defines two types of collisions: unaligned collision avoidance and separation. Unaligned collision avoidance is a behavior that \emph{``tries to keep characters which are moving in arbitrary directions from running into each other \citep{reynolds1999steering}."} Separation is similar to a rear-end collision and refers to a simpler form of movement: \emph{``Separation steering behavior gives a character the ability to maintain a certain separation distance from others nearby\citep{reynolds1999steering}.''} \citet{mavrogiannis2018social} discuss the point in space and time where agents collide, calling this point ``entanglement.'' This concept raises an additional question about the concrete implementation of this collision point --- what is considered close enough to be an entanglement in a social context? For example, \citet{mavrogiannis2019effects} utilized a minimum distance of $d \leq 1$ meter between the robot and the human. While it is simple to classify the direction of a collision, it is more challenging to define properly the minimal requirements of an encounter to be considered a collision. Is entering a person's personal space a collision? Is brushing against their leg? Overall, the definition for collision varies between researchers and may be a parameter that can be adjusted.

Another common discussion point is \textbf{context awareness and semantic mapping}. Many papers discuss the need for mobile social robots to be aware of their context \citep{charalampous2017recent}. A leading approach to enable this is semantic mapping, where the robot constructs maps that represent not only a metric occupancy grid but also other properties of the environment \citep{kostavelis2015semantic}$^{R1,R3}$. This survey does not focus on the mental model of the navigating robot (or of the other agents) in the environment, so this is not included in the taxonomy. It is, however, an important factor to consider when designing a robot for social navigation, as context awareness could greatly influence a robot's behavior.
    
Another thing to consider when designing interactions between mobile robots and pedestrians is how people \textbf{react to humans vs. robotic counterparts}. Will humans interacting with another people produce a similar or different responses when interacting with robots? The assumption that people will behave in the same way when encountering a robot as they would another human is common in HRI and other research communities, though it is not unanimously agreed upon. In their survey on proxemics for social navigation, \citet{rios2015proxemics} stated that \emph{``This article starts from the idea that people will keep the same conventions of social space management when they interact with robots than when they interact with humans. Researchers in social robotics that believe in that hypothesis can rely on the rich sociological literature to propose innovative models of social robots.''} As a counter opinion, \citet{butler2001psychological} indicate that people are most comfortable when a robot moves at speeds that are between $0.254m/s$ and $0.381m/s$, while the normal walking speed for young human is about $1m/s$. This difference suggests that people prefer a robot that moves more slowly than people do. Until there is a clear theory regarding the reactions of people to other people vs. robots in social navigation --- and until that theory is tested --- it is reasonable to exclude assumptions regarding whether people react to robots similarly or differently from how they react to other people from this taxonomy.

The distinction between \textbf{social cues and social signals} \citep{vinciarelli2009social}$^{R2,R3}$ is used in this survey, but they are not included as attributes in the taxonomy. Cues are the low-level inputs that the robot can receive or send, such as gaze, position, language, etc. Signals, on the other hand, are emotions, personality, and other traits that are more high-level. Signals discussed in the context of social navigation usually serve a purpose in conflict avoidance, and the way to implement them in a robot (or detect them in a human) is through social cues. How a robot can best communicate with humans is a rich and versatile research area; and is taken into consideration through the attributes \textbf{observability} and \textbf{communication} in the taxonomy.
    
One attribute that is relevant in a broader context than social navigation is \textbf{focused vs. unfocused interaction}. \citet{goffman2008behavior}$^{R2,R3}$ defines these terms to categorize scenarios in which the robot and the human share their focus (shared attention) vs. scenarios in which the robot and the human share an environment, but not attention. \citet{rios2015proxemics} use this attribute to identify different types of navigational behaviors in robots: minimizing probability of encounter, avoiding collisions, passing people, staying in line, approaching humans, following people, and walking side-by-side. Because the papers in this survey revolve around conflicts, the robot and the human do not share focus, and hence all included papers involve strictly unfocused interactions. Focused vs. unfocused interaction are not considered as part of the taxonomy.
    
Additionally, the topic of differences in navigation between \textbf{independent pedestrians, groups}, and \textbf{crowds} has enjoyed recent popularity \citep{yao2019following,murakami2021mutual,gupta2022intention}. Most social navigation papers either consider interactions with a single individual or with a crowd of individuals (as defined as \textbf{Number of Agents} in our taxonomy). An early sociological study showed that people tend to move in small groups rather than alone, but that group size distribution depends greatly on context (a casual Saturday afternoon stroll vs. a workday morning commute) \citep{coleman1961equilibrium}$^{R2,R3}$. Recent research has demonstrated that in many contexts, more than $50\%$ of pedestrians are traveling in groups \citep{moussaid2010walking}$^{R2,R3}$.  Thus, the context in which navigation takes place determines whether it is necessary to consider the surrounding crowd. 


Lastly, we address a distinction that is relatively straightforward to understand intuitively but is challenging to formalize: Conflict \textbf{Prevention vs. Resolution}. Consider a person walking in a crowded environment who is looking at their phone. Without watching the surrounding crowd, two people might collide --- which means they have reached a conflict. If the person looks up early enough, they might side-step abruptly without a change of speed --- which means that a conflict was resolved. If the person decides to step away to a less crowded area, this behavior is prevention.
On one hand, it is clear that prevention and resolution are different tasks that can direct the robot's behavior: prevention is the task of designing the robot's motion to steer away from potential conflicts, while resolution is the task of altering the robot's motion and behavior when a conflict is already imminent. On the other hand, formalizing this distinction is challenging, as it is non-trivial to define what is an ``imminent'' conflict. Whether a robot is designed to prevent or resolve a conflict, the premise of all of the covered work in this paper is that the robot is always attempting to avoid conflicts in social navigation. This requirement provides a crisp way to identify relevant papers that fit into this survey, without the need to explicitly cluster the interaction into prevention or resolution.   


\section{Models}
\label{sec::models}
This section details various models used for social navigation. The discussion is grouped according to three main underlying models: Multiagent systems, human-inspired models, and physics-based models (specifically, the social force model and other force models). Each of these categories represents a different set of assumptions --- as well as a different research community --- that each model stems from. Navigation in multiagent systems is usually designed with the premise that agents navigating in an environment are homogeneous. These papers include multi-robot navigation models and crowd modeling. A multiagent social navigation model generally reasons about agents with different --- sometimes unknown --- behaviors. Other models are inspired by insights about human navigation. These papers provide measurements and rules that explain how people navigate among each other. Such a social navigation model translates these rules into robot motion and perception. We taxonomize papers using the social force model in their own category; inspired by physical force modeling. Many models have been proposed which build upon the seminal work by \citet{helbing1995social}$^{R2,R3}$, using additional forces. Finally, some papers sit at an intersection between two categories. In such cases, the paper is grouped with work that uses similar motion control.

\subsection{Multiagent Systems}
\begin{table}[]
\caption{An overview of the different multiagent based models used in social navigation. \textbf{Role} refers to the robot role, \textbf{Obs.} is observability, \textbf{Com.} refers to communication, and \textbf{Exp. type} is the experiment type.}
\label{tab:mod_MA}
\footnotesize
\begin{tabular}{|l|l|l|l|l|l|l|l|l|}
\hline
\textbf{Year} & \textbf{Paper}                     & \textbf{Role} & \textbf{\# Agents} & \textbf{Obs.}                                          & \textbf{Motion Control}                                      & \textbf{Com.} & \textbf{\begin{tabular}[c]{@{}l@{}}Exp.\\ Type\end{tabular}} & \textbf{\begin{tabular}[c]{@{}l@{}}Agent\\ Type\end{tabular}} \\ \hline
1997          & \citet{musse1997model}                             & R             & Abs=10             & Full                                                   & \begin{tabular}[c]{@{}l@{}}Other\\ (Hand Coded)\end{tabular} & N             & Sim                                                          & Hom                                                           \\ \hline
2005          & \citet{strassner2005virtual}                       & N             & Abs=2              & Partial                                                & \begin{tabular}[c]{@{}l@{}}Other\\ (Hand Coded)\end{tabular} & N             & Sim                                                          & Hom                                                           \\ \hline
2010          & \citet{foka2010probabilistic}                      & R             & Abs=6              & Depth                                                  & \begin{tabular}[c]{@{}l@{}}Other\\ (POMDP)\end{tabular}      & I             & ItW                                                           & H-R                                                           \\ \hline
2011          & \citet{van2011reciprocal}                          & R             & Abs=1000           & Full                                                   & ORCA                                                         & I             & Sim                                                          & Hom                                                           \\ \hline
2013          & \citet{bandyopadhyay2013intention}                 & R             & Abs=4              & \begin{tabular}[c]{@{}l@{}}RGB + \\ Depth\end{tabular} & SFM                                                          & N             & Lab                                                     & H-R                                                           \\ \hline
2014          & \citet{bonneaud2014empirically}                    & R             & Abs=20             & Full                                                   & Other                                                        & N             & Sim                                                          & Hom                                                           \\ \hline
2014          & \citet{okal2014towards}                            & R             & Abs=176            & Partial                                                & SFM                                                          & N             & Sim                                                          & Hom                                                           \\ \hline
2016          & \citet{godoy2016moving}                            & B             & Abs=100            & Full                                                   & Other                                                        & N             & Lab                                                     & R-R                                                           \\ \hline
2019          & \citet{chen2019crowd}                              & R             & Abs=6              & Full                                                   & \begin{tabular}[c]{@{}l@{}}Other\\ (LM-SARL)\end{tabular}    & N             & Sim$^+$                                                         & Hom                                                           \\ \hline
2022          & \citet{gupta2022intention}                              & R             & Abs=401              & Partial                                                   & \begin{tabular}[c]{@{}l@{}}Other\\ (POMDP)\end{tabular}    & N             & Sim                                                         & Hom                                                           \\ \hline

\end{tabular}
\normalsize
\end{table}

Two communities that have contributed significantly to the study of social navigation are the multi-robot navigation and graphics communities. Both of these communities have proposed different approaches to model the behavior of a crowd. The multi-robot community focuses more on safety and feasibility in the real-world, while the graphics community focuses on robustness. Multi-robot work usually is based on a few interactions under realistic constraints. On the other hand, the challenge of crowd modeling taken on by the graphics community is to model interactions between hundreds and thousands of agents simultaneously. However, because the graphics community does not need to implement these systems on real robots, the perception and movement restrictions on those agents tends not to be grounded in the physical constraints that both robots and real people must contend with. 

Many researchers have approached the challenge of multi-robot navigation \citep{yan2013survey}$^{R1,R3}$.
This is a fertile and active research area that deserves its own survey. We discuss only a few selected publications that have had significant influence on social navigation.
\citet{van2011reciprocal} present the principle of optimal reciprocal collision avoidance (ORCA), which provides a sufficient condition for multiple robots to avoid collisions among one another, and guarantees collision-free navigation. 
\citet{chen2019crowd} model human-robot and human-human interactions, then infer the relative importance of learned features through a pooling module via a self-attention mechanism, finally planning motions.

Another branch of multi-robot research focuses on planning under uncertainty, and leverages Markov Decision Processes (MDPs). \citet{foka2010probabilistic} model a probabilistic prediction of people's destinations. They use a Partially Observable MDP (POMDP) solved online at each time step to determine which actions the robot actually performs. \citet{gupta2022intention} recently presented an additional POMDP model for intention-aware navigation in crowds, where the model can address decisions related both to the robot's speed and its heading.
\citet{bandyopadhyay2013intention} model human intention with a Mixed Observability MDP (MOMDP), then plan the motion of a robot leveraging this model. 

The graphics community has contributed several important models to social navigation, as well as simulation environments that can be utilized to evaluate other models and algorithms (see more about these simulation environments in Section \ref{sec::evaluating}).
\citet{musse1997model} propose a model of crowd behavior, where agent behavior is determined using a predefined set of rules.
\citet{strassner2005virtual} use behavioral rules for modeling each person’s behavior in a crowd. Such behaviors include perceiving, storing, and forgetting knowledge.
\citet{bonneaud2014empirically} model pedestrian behavior using an empirically-grounded emergent approach, where the local control laws for locomotor behavior are derived experimentally, and the global crowd behavior is emergent.
\citet{okal2014towards} present a model for crowd behavior in which groups are formed. Their representation gives each individual an internal state, where under a set of predefined conditions pedestrians can choose to walk together.

Table \ref{tab:mod_MA} summarizes the taxonomy values for models inspired by multiagent systems research.

\subsection{Psychology and Human-Inspired Models}

\begin{table}[]
\caption{An overview of the different human-inspired and psychology-based models used in social navigation. \textbf{Role} refers to the robot role, \textbf{Obs.} is observability, \textbf{Com.} refers to communication, and \textbf{Exp. type} is the experiment type.}
\label{tab:mod_hum}
\footnotesize
\begin{tabular}{|l|l|l|l|l|l|l|l|l|}
\hline
\textbf{Year} & \textbf{Paper}                   & \textbf{Role} & \textbf{\# Agents} & \textbf{Obs.} & \textbf{Motion Control}                                      & \textbf{Com.} & \textbf{\begin{tabular}[c]{@{}l@{}}Exp.\\ Type\end{tabular}} & \textbf{\begin{tabular}[c]{@{}l@{}}Agent\\ Type\end{tabular}} \\ \hline
1995          & \citet{cutting1995we}                            & R             & Abs=2             & Partial       & \begin{tabular}[c]{@{}l@{}}Other\\ (Hand Coded)\end{tabular} & I             & Sim + Sur                                                    & H-H                                                           \\ \hline
1998          & \citet{jeffrey1998constructing}                  & R             & Abs=4+            & Partial       & Human                                                        & N             & Sim                                                          & H-H                                                           \\ \hline
1999          & \citet{Patla1999}                                & R             & Abs=2             & Partial       & Human                                                        & N             & Lab                                                           & H-H                                                           \\ \hline
1999          & \citet{reynolds1999steering}                     & R             & Abs=2             & Full          & None                                                         & N             & None                                                         & Hom                                                           \\ \hline
2002          & \citet{bennewitz2002learning}                    & R             & Abs=2             & Depth         & \begin{tabular}[c]{@{}l@{}}Other\\ (Learned)\end{tabular}    & N             & Lab                                                            & H-R                                                           \\ \hline
2008          & \citet{gerin2008characteristics}                 & R             & Abs=2             & None          & Human                                                        & N             & Lab                                                           & H-H                                                           \\ \hline
2010          & \citet{henry2010learning}                        & R             & None              & Depth         & Other (A*)                                                   & N             & Sim                                                          & Hom                                                           \\ \hline
2010          & \citet{kitazawa2010pedestrian}                   & R             & Abs=4             & Partial       & Human                                                        & N             & Lab                                                           & H-H                                                           \\ \hline
2011          & \citet{moussaid2011simple}                       & R             & Abs=96            & Partial       & \begin{tabular}[c]{@{}l@{}}Other\\ (Hand Coded)\end{tabular} & N             & Lab 
& Hom                                                           \\ \hline
2011          & \citet{o2011learning}                            & R             & Abs=2             & Depth         & \begin{tabular}[c]{@{}l@{}}Other\\ (Planner)\end{tabular}    & N             & ItW                                                           & H-R                                                           \\ \hline
2012          & \citet{rios2012navigating}                       & R             & Abs = 6           & RGB+Depth     & ROS                                                          & N             & Sim                                                          & Hom                                                           \\ \hline
2013          & \citet{lu2013tuning}                             & R             & Abs = 2           & Partial       & ROS                                                          & N             & Sim                                                          & Het                                                           \\ \hline
2013          & \citet{park2013collision}                        & R             & D=0.1-1           & Partial       & \begin{tabular}[c]{@{}l@{}}Other\\ (Hand Coded)\end{tabular} & I             & Sim                                                          & Hom                                                           \\ \hline
2014          & \citet{charalampous2014social}                   & R             & Abs=2             & RGB+Depth     & Other                                                        & N             & ItW                                                           & H-R                                                           \\ \hline
2014          & \citet{papadakis2014adaptive}                    & I             & Abs=2             & RGB+Depth     & None                                                         & I             & Lab                                                           & H-R                                                           \\ \hline
2014          & \citet{vasquez2014inverse}                       & R             & Abs=6+            & Full          & \begin{tabular}[c]{@{}l@{}}Other\\ (Dijkstra)\end{tabular}   & I             & Sim                                                          & H-A                                                           \\ \hline
2015          & \citet{unhelkar2015human}                        & R             & Abs=2             & Full          & \begin{tabular}[c]{@{}l@{}}Other\\ (SIPP)\end{tabular}       & N             & Lab 
& H-A                                                           \\ \hline
2016          & \citet{mead2016perceptual}                       & I             & Abs=2             & RGB           & None                                                         & D             & Lab                                                           & H-R                                                           \\ \hline
2016          & \citet{truong2016dynamic}                        & R             & Abs = 4           & RGB+Depth     & Other (D*)                                                   & N             & Lab 
& H-R                                                           \\ \hline
2020          & \citet{senft2020would}                           & R             & Abs = 2           & Depth         & \begin{tabular}[c]{@{}l@{}}Other\\ (Hand Coded)\end{tabular} & I             & Lab                                                           & H-R                                                           \\ \hline
2022          & \citet{karnan2022socially}                          & I             & Abs = 2           & RGB+Depth         & \begin{tabular}[c]{@{}l@{}}Other\\ (Teleoperation)\end{tabular} & I             & ItW                                                           & H-R                                                           \\ \hline
\end{tabular}
\normalsize
\end{table}

\begin{figure}
    \centering
    \includegraphics[width=\textwidth]{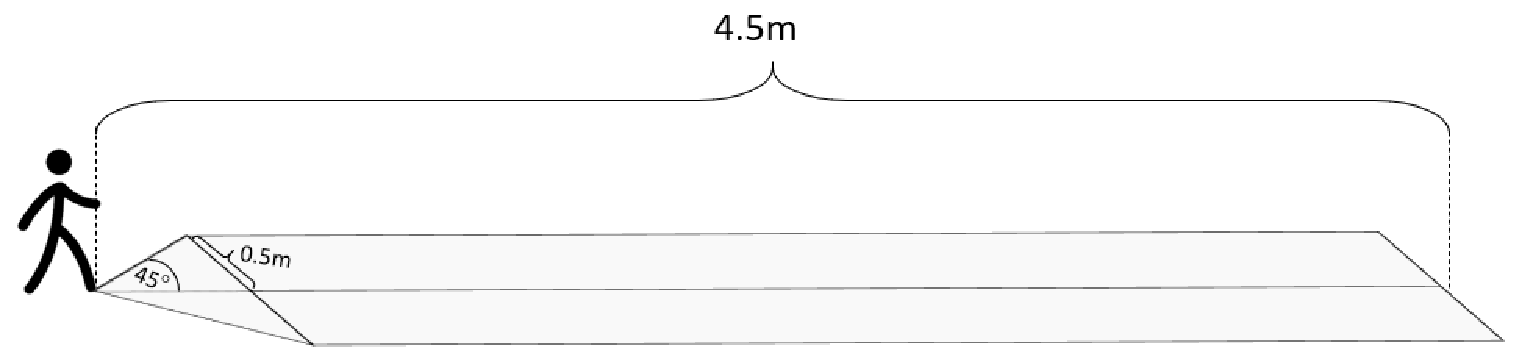}
    \caption{Information Process Space - the visual processing coverage of pedestrians, as measured by \citep{kitazawa2010pedestrian} and depicted by \citet{rios2015proxemics}.}
    \label{fig:IPS}
\end{figure}

The contributions discussed so far have focused on multiagent or multi-robot navigation systems that have been adapted to accommodate human pedestrians. A different approach starts with the modeling of human behavior, which then leverages these models for improving robot navigation. \citet{cutting1995we}$^{R2,R3}$ empirically evaluates human behavior in situations of obstacle avoidance. Their work investigates the relationship between object avoidance and finding one's aimpoint in a series of human studies. Their results are summarized as a decision-tree to facilitate reasoning about collision detection with other objects (static or moving) and Gaze-Movement Angle (GMA); the angle between one's gaze and one's direction of movement. Their model can be used to estimate where a collision might occur. As a different way to estimate an expected collision point, \citet{carel1961visual} defined $\tau$ to be the time to bypass a dynamic obstacle (human or not). \citet{moussaid2011simple} use $\tau$ to heuristically plan how to navigate in a way that avoids collisions. \citet{park2013collision} claim that GMA-based collision prediction has several advantages over the time-to-contact ($\tau$) approach. It is more robust to variations in the speed and the path of the other pedestrian. It also does not assume either constant speed or a linear path, so the accuracy of the prediction is not affected by these variations.
\citet{kitazawa2010pedestrian} investigate the Information Process Space (IPS) of a navigating person when walking in a hallway in the presence of static objects and other pedestrians. In this work, they identify the area that the observing pedestrian considers as the one in which a collision with another pedestrian could occur in a short time (see Figure~\ref{fig:IPS}). 
In an extension of this work, \citet{park2013collision} propose a collision avoidance behavior model that is based on their empirical results about IPS to generate more human-like collision avoidance behaviors. 

Another concept from psychology that has had a significant impact on social navigation is that of personal space \citep{hall1966hidden, jeffrey1998constructing, gerin2008characteristics}. While the original formulation of personal space is depicted by \citet{hall1966hidden}$^{R2,R3}$ as a concentric circle, later work extends that to an egg \citep{hayduk1981shape}$^{R2,R3}$, ellipse \citep{helbing1995social}, or as asymmetrical (smaller on the dominant side) \citep{gerin2008characteristics}$^{R2,R3}$ shape. Closely related to personal space is the concept of density in crowds. The average density of people in a non-crowded environment has been evaluated to be $0.03$ pedestrians per $m^2$, whereas in a moderately crowded environment, there are $0.25$ pedestrians per $m^2$ \citep{moussaid2010walking}.
\citet{rios2012navigating} incorporate both personal space and IPS-based constraints into an adaptive optimization algorithm to enable more human-like navigation. \citet{truong2016dynamic} propose a comprehensive framework that reasons about pedestrians' extended personal space and social interaction space to identify a Dynamic Social Zone (DSZ); a concept which is incorporated into their motion planner.

Others have analyzed how gait and posture are affected by a sudden trajectory changes, as one might expect to see in conflict avoidance. \citet{Patla1999}$^{R2,R3}$ analyzed head yaw, trunk yaw and foot position when turning due to an expected obstacle vs. turning abruptly due to an unexpected obstacle. To analyze the relationship between head pose and predicted walking trajectory, \citet{unhelkar2015human}$^{R2,R3}$ discretized walking trajectories as a decision problem regarding which target a person would walk toward. They incorporated this information into an anytime path planner ~\citep{narayanan2012anytime}$^{R1}$ and evaluated this enhanced planner in simulation. \citet{holman2021watch}$^{R2,3}$ extend this predictive model to incorporate gaze. \citet{senft2020would} identify and implement a navigational pattern for making space in a hallway. Their model involves controlling the robot's rotation and sliding motion, and consists of three steps: step, slide, and rotate.

All of the contributions above leverage insights from empirical studies on humans and robots to manually construct models for social navigation. However, together with the improving abilities of machine learning, different learning techniques have been used to learn models of navigation in social contexts.
\citet{lu2013tuning} propose a planning model that can be tuned to match different social navigation contexts.
\citet{bennewitz2002learning} learn motion patterns of people that can be used for trajectory prediction in social robots. \citet{henry2010learning} extend this approach by modeling partial trajectories. More recently, \citet{vasquez2014inverse} used Inverse Reinforcement Learning (IRL) to infer a reward function for social navigation. They introduce a new software framework to systematically investigate the effect of features and learning algorithms used in the literature. They investigate the task of socially-compliant robot navigation in crowds, evaluating two different IRL approaches and several feature sets in large-scale simulations. \citet{karnan2022socially} collected a large scale human demonstration dataset, containing socially compliant data of navigation behaviors in natural indoor and outdoor spaces on a university campus. They used behavior cloning to learn a global and local planner to mimic human navigation behaviors.  

Table \ref{tab:mod_hum} summarizes the taxonomy values for models inspired by human behavior, physiology, and psychology research.


\subsection{Physics-based Models}

\begin{table}[]
\caption{An overview of the different physics-inspired models used in social navigation. \textbf{Role} refers to the robot role, \textbf{Obs.} is observability, \textbf{Com.} refers to communication, and \textbf{Exp. type} is the experiment type.}
\label{tab:mod_phy}
\footnotesize
\begin{tabular}{|l|l|l|l|l|l|l|l|l|}
\hline
\textbf{Year} & \textbf{Paper}                   &  \textbf{Role} & \textbf{\# Agents} & \textbf{Obs.} & \textbf{Motion Control}                                          & \textbf{Com.} & \textbf{\begin{tabular}[c]{@{}l@{}}Exp.\\ Type\end{tabular}} & \textbf{\begin{tabular}[c]{@{}l@{}}Agent\\ Type\end{tabular}} \\ \hline
1995          & \citet{helbing1995social}                        & R             & D=0.3             & Full          & SFM                                                              & N             & Sim                                                          & H-A                                                           \\ \hline
2003          & \citet{loscos2003intuitive}                      & R             & Abs=6000          & Partial       & \begin{tabular}[c]{@{}l@{}}Other\\ (Hand Coded)\end{tabular}     & N             & Sim                                                          & Hom                                                           \\ \hline
2009          & \citet{karamouzas2009predictive}                 & R             & Abs=1000          & Full          & SFM                                                              & N             & Sim                                                          & Hom                                                           \\ \hline
2010          & \citet{moussaid2010walking}                      & N             & D=0.03-0.25       & Full          & SFM                                                              & N             & ItW 
& H-H                                                           \\ \hline
2010          & \citet{svenstrup2010trajectory}                  & R             & Abs=40            & Full          & \begin{tabular}[c]{@{}l@{}}Other\\ (Modified\\ RRT)\end{tabular} & I             & Sim                                                          & Hom                                                           \\ \hline
2020          & \citet{swofford2020improving}                    & I             & Abs=18            & RGB           & ROS                                                              & N             & Lab                                                           & H-R                                                           \\ \hline
\end{tabular}
\normalsize
\end{table}

Researchers have also used models inspired by physics to represent dynamics and interactions between different moving agents. \citet{helbing1995social} were the first to propose the Social Force Model (SFM), a model inspired by fluid dynamics that describes an agent's motion using a set of repelling and attracting forces. They evaluate this model in a simulation of homogeneous SFM-based agents.
Many contributions extend SFM models to handle additional forces: \citet{karamouzas2009predictive} add an evasive force that uses collision prediction and avoidance, which makes agents more proactive and anticipatory than the classical SFM. \citet{moussaid2010walking} propose several group-related forces that help model pedestrians that walk in a group. \citet{swofford2020improving}$^{R2,R3}$ use a Deep Affinity Network (DANTE) to predict the likelihood that two individuals in a scene are part of the same conversational group. They take into consideration the social context in which these interactions take place. A different type of force inspired work uses potential fields attached to moving pedestrians \citep{svenstrup2010trajectory}. This model has been leveraged in a modified Rapidly-exploring Random Tree (RRT) for navigation in human environments, though it assumes access to full state information.

Table \ref{tab:mod_phy} summarizes the taxonomy values of models inspired by physics and mechanical engineering research.











\section{Algorithms}
\label{sec::algorithms}

This section discusses contributions in the form of algorithms and hardware augmentations that enhance social navigation. Most of the work presented here fits our basic definition of social navigation, however several papers are included which have not been evaluated in the context of navigating around people. These papers are included if their contribution can be applied in the context of social navigation.
Broadly speaking, this section is divided into three main approaches: Approaches that \textbf{infer} the human's trajectory and adapt to it; Approaches that \textbf{convey} the goal or trajectory of the robot to the person it is interacting with before reaching a conflict; and mixed approaches which \textbf{mediate} between the inferred trajectory of the human and the desired goal of the robot. 

\subsection{Inferring Human Trajectories}

\begin{table}[ht]
\caption{An overview of the different inference algorithms used in social navigation. \textbf{Role} refers to the robot role, \textbf{Obs.} is observability, \textbf{Com.} refers to communication, and \textbf{Exp. type} is the experiment type.}
\label{tab:alg_inf}
\footnotesize
\begin{tabular}{|l|l|l|l|l|l|l|l|l|}
\hline
\textbf{Year} & \textbf{Paper}               & \textbf{Role} & \textbf{\# Agents} & \textbf{Obs.}                                          & \textbf{Motion Control}                                          & \textbf{Com.} & \textbf{\begin{tabular}[c]{@{}l@{}}Exp.\\ Type\end{tabular}} & \textbf{\begin{tabular}[c]{@{}l@{}}Agent\\ Type\end{tabular}} \\ \hline
2006          & \citet{pacchierotti2006design}                 & R             & Abs=3             & Depth                                                  & \begin{tabular}[c]{@{}l@{}}Other\\ (Hand Coded)\end{tabular}     & N             & Sim                                                          & Hom                                                           \\ \hline
2007          & \citet{gockley2007natural}                      & R             & Abs=2             & Depth                                                  & \begin{tabular}[c]{@{}l@{}}Other\\ (CVM)\end{tabular}            & D             & Lab                                                           & H-R                                                           \\ \hline
2007          & \citet{sisbot2007human}                        & R             & Abs=2             & \begin{tabular}[c]{@{}l@{}}RGB + \\ Depth\end{tabular} & \begin{tabular}[c]{@{}l@{}}Other\\ (HAMP)\end{tabular}           & I             & Lab                                                           & H-R                                                           \\ \hline
2009          & \citet{kirby2009companion}                    & R             & Abs=4             & Depth                                                  & Other (A*)                                                       & N             & Sim                                                          & Hom                                                           \\ \hline
2010          & \citet{ohki2010collision}                      & R             & Abs=5             & Full                                                   & \begin{tabular}[c]{@{}l@{}}Other\\ (Hand Coded)\end{tabular}     & N             & Sim                                                            & Hom                                                           \\ \hline
2010          & \citet{pandey2010framework}                    & R             & Abs=2             & Full                                                   & \begin{tabular}[c]{@{}l@{}}Other\\ (Hand Coded)\end{tabular}     & N             & Lab                                                           & H-R                                                           \\ \hline
2010          & \citet{tamura2010smooth}                       & R             & Abs=2             & Depth                                                  & SFM                                                              & N             & Lab                                                           & H-R                                                           \\ \hline
2011          & \citet{diego2011please}                         & R             & Abs=5             & None                                                   & \begin{tabular}[c]{@{}l@{}}Other\\ (Modified TSP)\end{tabular}   & N             & Sim                                                           & Het                                                           \\ \hline
2012          & \citet{kuderer2012feature}                    & R             & Abs=3             & Full                                                   & Other (learned)                                                  & N             & Lab                                                           & H-R                                                           \\ \hline
2012          & \citet{luber2012socially}                      & R             & Abs=2             & Full                                                   & Other (RMP)                                                      & N             & Sim                                                          & H-A                                                           \\ \hline
2013          & \citet{ratsamee2013human}                    & R             & Abs=2             & \begin{tabular}[c]{@{}l@{}}RGB +\\ Depth\end{tabular}  & SFM                                                              & N             & Lab                                                           & H-R                                                           \\ \hline
2014          & \citet{gomez2014fast}                           & R             & Abs=5             & Full                                                   & \begin{tabular}[c]{@{}l@{}}Other\\ (Planning)\end{tabular}       & N             & Sim                                                          & R-R                                                           \\ \hline
2016          & \citet{kim2016socially}                       & R             & Crowd             & \begin{tabular}[c]{@{}l@{}}RGB + \\ Depth\end{tabular} & \begin{tabular}[c]{@{}l@{}}Other\\ (Costmap Search)\end{tabular} & I             & ItW                                                           & H-R                                                           \\ \hline
2016          & \citet{kretzschmar2016socially}                 & R             & Abs=3             & Depth                                                  & \begin{tabular}[c]{@{}l@{}}Other\\ (RPROP)\end{tabular}          & I             & Lab                                                           & H-R                                                           \\ \hline
2016          & \citet{okal2016learning}                       & R             & Abs=4             & Depth                                                  & ROS                                                              & I             & Sim$^+$                                                           & Hom                                                           \\ \hline
2016          & \citet{pfeiffer2016predicting}                  & R             & Abs=891           & RGB                                                    & \begin{tabular}[c]{@{}l@{}}Other\\ (Max Entropy)\end{tabular}    & I             & ItW                                                           & Hom                                                           \\ \hline
2017          & \citet{bera2017sociosense}                      & R             & D<=2              & Full                                                   & \begin{tabular}[c]{@{}l@{}}Other\\ (SocioSense)\end{tabular}     & N             & Sim                                                          & Het                                                           \\ \hline
2017          & \citet{chen2017socially}                        & R             & Abs=10+           & RGB                                                    & \begin{tabular}[c]{@{}l@{}}Other\\ (Learned)\end{tabular}        & N             & ItW                                                           & H-R                                                           \\ \hline
2017          & \citet{chen2017decentralized}                 & R             & Abs=6             & Full                                                   & \begin{tabular}[c]{@{}l@{}}Other\\ (Learned)\end{tabular}        & N             & Sim                                                          & R-R                                                           \\ \hline
2018          & \citet{ding2018hierarchical}                   & R             & Abs=20            & Depth                                                  & None                                                             & N             & Sim                                                          & R-R                                                           \\ \hline
2018          & \citet{everett2018motion}                      & R             & Abs=10+           & \begin{tabular}[c]{@{}l@{}}RGB +\\ Depth\end{tabular}  & Crowd                                                            & N             & Sim$^+$                                                           & H-A                                                           \\ \hline
2018          & \citet{ROMAN18-Jiang}                           & B             & Abs=2             & RGB                                                    & \begin{tabular}[c]{@{}l@{}}Other\\ (Hand Coded)\end{tabular}     & N             & Sim                                                          & H-A                                                           \\ \hline
2018          & \citet{li2018role}                             & R             & Abs=3+            & Depth                                                  & \begin{tabular}[c]{@{}l@{}}Other\\ (Learned)\end{tabular}        & N             & Lab                                                           & H-R                                                           \\ \hline
2018          & \citet{long2018towards}                        & R             & Abs=100           & Depth                                                  & \begin{tabular}[c]{@{}l@{}}Other\\ (Learned)\end{tabular}        & N             & Sim                                                          & R-R                                                           \\ \hline
2018          & \citet{tai2018socially}                        & R             & Abs=3             & Depth                                                  & \begin{tabular}[c]{@{}l@{}}Other\\ (Learned)\end{tabular}        & N             & Sim$^+$                                                           & H-A                                                           \\ \hline
2019          & \citet{jin2019mapless}                        & R             & Abs=4             & Depth                                                  & \begin{tabular}[c]{@{}l@{}}Other\\ (Learned)\end{tabular}        & N             & Lab                                                           & H-R                                                           \\ \hline
2019          & \citet{meng2019scaling}                        & N             & Abs=1             & RGB                                                    & None                                                             & N             & Sim                                                            & Hom                                                           \\ \hline
2019          & \citet{nardi2019long}                          & N             & Abs=1             & Full                                                   & \begin{tabular}[c]{@{}l@{}}Other\\ (Hand Coded)\end{tabular}     & N             & Sim                                                          & R-R                                                           \\ \hline
2020          & \citet{liang2020crowdsteer}                    & R             & Abs=10+           & \begin{tabular}[c]{@{}l@{}}RGB +\\ Depth\end{tabular}  & \begin{tabular}[c]{@{}l@{}}Other\\ (Learned)\end{tabular}        & N             & Lab                                                           & H-R                                                           \\ \hline

2022          & \citet{lu2022socially}                   & R             & Abs=5           & \begin{tabular}[c]{@{}l@{}} Depth\end{tabular}  & \begin{tabular}[c]{@{}l@{}}Other\\ (Learned)\end{tabular}        & N             & Sim                                                           & Hom                                                           \\ \hline

\end{tabular}
\normalsize
\end{table}

Many social navigation contributions have been inspired by the way humans navigate in social contexts. The majority of these papers can be split into two categories: online and offline inference. Online inference means that a robot observes the behavior of a person during deployment and incorporates its inference about the person's planned trajectory into its execution. Offline inference happens prior to the execution stage, usually on more than a single trajectory. The robot learns to predict human trajectories or imitate them from a set of observed trajectories. 


\subsubsection{Online Inference}

\citet{cutting1995we} offer an early attempt to evaluate the trajectory of a passerby by calculating their GMA and reacting to it.
The robot designed by \citet{tamura2010smooth} detects pedestrians by using a laser range finder and tracks using a Kalman filter. They apply a social force model to the observed trajectory to determine whether the pedestrian intends to avoid a collision with the robot or not, and select an appropriate behavior based on the estimation result.
\citet{gockley2007natural} discuss how to avoid rear-end collisions in the context of person following. They propose a laser-based person-tracking method and evaluate two different approaches to person-following: direction-following, where the robot follows the current location of the person; and path-following, where the robot tries to follow the exact path that the person took. They show that while no significant difference was found between the two approaches in terms of the distance or time between tracking errors, participants rated the robot’s behavior as significantly more natural and human-like in the direction-following condition. In addition, participants
felt that the direction-following robot's behavior is more similar to the participants' expectations.

Others have leveraged human gaze to infer the trajectory of pedestrians. Gaze is a very strong communicative cue used by humans, in the context of collaborative settings in general \citep{castiello2003understanding}$^{R2,R3}$ and for navigation in particular \citep{admoni2017social}. It has been shown that humans are not the only species that can partially understand gaze cues from a very young age, but also chimpanzees and dogs \citep{povinelli1999comprehension, soproni2001comprehension}$^{R2,R3}$. Gaze and head pose have both been shown to be significant indicators of a person's attention, which can be used to infer navigational goals. \citet{stiefelhagen1999gaze}$^{R2,R3}$ show that the visual focus of a person's attention can be deduced from head pose when the visual resolution is insufficient to determine eye gaze. \citet{4359373}$^{R2,R3}$ extend their work to a varying number of moving pedestrians.
Of course, this gaze behavior extends beyond walking and bicycling. Recent work has studied the use of gaze as a modality for plan recognition in games \citep{singh2020combining} and as a cue for interacting with copilot systems in cars \citep{ROMAN18-Jiang,ICMI18-Jiang}, also with the aim of inferring the driver's intended trajectory. Gaze is also often fixated on objects being manipulated, which can be leveraged to improve algorithms which learn from human demonstrations \citep{saran2020understanding}$^{R1,R2}$. Though the use of instrumentation such as head-mounted gaze trackers or static gaze tracking cameras is limiting for mobile robots, recent work in the development of gaze trackers which work without such equipment \citep{saran2018human}$^{R1,R2}$ may soon allow us to perform the inverse of the robot experiments presented here, with the robot reacting to human gaze.
\citet{ratsamee2013human} propose to avoid collisions with humans by considering a social model that takes into consideration body pose and face orientation.

\subsubsection{Offline Inference and Learning }
While the previous subsection focused on the recognition of human's trajectories during execution, some leverage these trajectories to learn and infer how a human would react in a social navigation interaction.
\citet{pacchierotti2006design} design a rule-based strategy for people passing that was inspired by spatial behavior studies. This strategy intends to mimic the way people avoid collisions once inside a person's personal space.

One such successful approach uses \textbf{Inverse Reinforcement Learning (IRL)} to elicit the explicit cost representation to imitate human's social navigation behavior. Instead of hand-crafted functions, these papers use IRL to leverage data-driven approaches. IRL was extensively used to infer reward (cost) functions from human demonstrations. The most straightforward application of IRL is by \citet{kim2016socially}, to learn a cost function that respects social variables over features extracted from a RGB-D sensor. This work used IRL to infer cost functions in a social navigation context: navigational features were firstly extracted from an RGB-D sensor, then represented as a local cost function learned from a set of demonstration trajectories by an expert using IRL. The system still operated under the classical navigation pipeline, with a global path planned using a shortest-path algorithm, and local path using the learned cost function to respect social variables. Obstacle avoidance was still handled by a low-level controller. 
\citet{okal2016learning} tackle cost function representation at a global level in social context: they developed a graph structure and used Bayesian IRL to learn the cost for this representation. With the learned global representation, traditional global planner (A*) planned a global path over this graph, and POSQ steer function for differential-drive mobile robots served as a local planner. 
\citet{henry2010learning} use Inverse Reinforcement Learning to learn motion patterns of humans in simulation that can later be used for planning in social navigation.

An alternative approach to IRL with a similar objective is to model social navigation trajectories using a \textbf{Maximum Entropy Probability Distribution}, where cost is also implicitly defined by identifying an underlying model from demonstrated data. Maximum entropy probability distribution has been used by \citet{pfeiffer2016predicting} to model agents’ trajectories for planning and by \citet{kretzschmar2016socially} to infer the parameters of the navigation model that matches the observed behavior in expectation. 
\citet{kuderer2012feature} also use human demonstrations, but instead of using a Markov Decision Process, they elicit features from the human trajectories, and then use entropy maximization to determine the robot's behavior. \citet{luber2012socially} use unsupervised learning from surveillance data to learn motion patterns and augment a motion planner with this knowledge.

\citet{sisbot2007human} create a human aware motion planner (HAMP) that is explicitly given a cost model for safety and for legibility, and the robot reasons about the joint cost of these two properties in its planning process. Costs were also implicitly defined by identifying an underlying model from demonstrated data. 
\citet{kirby2009companion} model human social conventions at the global planning stage. This enables it to mediate between different, sometimes conflicting objectives. For example, consider a goal that is down an intersecting hallway to the robot's left. While the social norm in many places is to pass a pedestrian from the right side, the robot may choose to walk across the hallway in front of an oncoming person, effectively passing them on the left of the corridor. This behavior is the result of mediating between two objectives: complying with the right-alignment social norm, and minimizing the time to the goal.

Many algorithms use hand-crafted behaviors to avoid conflicts, i.e. to realize collision avoidance. As a continuation of previous Collision Avoidance Deep Reinforcement Learning (CADRL) work \citep{chen2017decentralized}, \citet{chen2017socially} further propose a hand-crafted reward function to incorporate the social norm of left or right-handed passing in a DRL approach and enabled a physical robot to move at human walking speed in an environment with many pedestrians, called Socially Aware CADRL (SA-CADRL). 
Along the same line of research, but to relax the assumption of other agents' dynamics, \citet{everett2018motion} propose GA3C-CADRL, using an LSTM to allow reasoning about an arbitrary number of nearby agents and GPU to maximize the number of training experiences.
Similarly, the reward function by \citet{jin2019mapless} accounts for ego-safety, to measure collision from the robot’s perspective, and social-safety, to measure the impact of the robot’s actions on surrounding pedestrians. 
Other options that utilize DRL include using a Hidden Markov Model (HMM) in a higher hierarchy to learn to choose between target pursuing and collision avoidance using RL \citep{ding2018hierarchical}. 
%
%
\citet{tai2018socially} use Generative Adversarial Imitation Learning (GAIL) to learn continuous actions and desired force toward the target. This improved safety and efficiency over pure BC.
\citet{li2018role} propose a new problem: socially concomitant navigation (SCN). In addition to collision avoidance in traditional social navigation, in SCN the robot also needs to consider the motion of its companion so as to maintain a sense of affinity when they are traveling together towards a certain goal. Taking features extracted from a LiDAR sensor along with the goal as input, a navigation policy is trained by Trust Region Policy Optimization (TRPO) to output continuous velocity commands for navigation. 
\citet{bera2017sociosense} create SocioSense, a social navigation algorithm that categorizes pedestrians according to psychological traits (e.g. shy, tense) and adjusts the robot's velocity according to the pedestrians around it.
\citet{lu2022socially} incorporated a dynamic measure into their reward to reason about the density of the crowd when deciding on the distance from other pedestrians. They then extended the deep neural network architecture from SARL \citep{chen2019crowd} to choose the optimal action with the shaped reward that reasons about the ``uncomfortable distance" between the robot and a pedestrian.

To observe social rules when navigating in densely populated environments, \citet{yao2019following} propose to utilize information about social groups to address the ``naturalness'' aspect from the perspective of collective formation behaviors in the complex real world. They used a deep neural network, called Group-Navi GAN, to track social groups and navigate the robot to join the flow of a social group through providing a local goal to the local planner. Other components of the existing navigation pipeline, e.g. state estimation, collision avoidance, etc., remained the same. The classical navigation pipeline, with the assistance of a learned local goal, was capable of navigating safely in a densely populated area following crowd flows to reach the goal. 
 \citet{liang2020crowdsteer} develop CrowdSteer, a RL-based collision-avoidance algorithm that navigates in dense and crowded environments. The algorithm is trained using PPO in simulation with simulated human agents, and was deployed in the real-world. 
 \citet{martins2019clusternav} propose ClusterNav, an algorithm that gets human demonstrations using teleoperation, then uses Expectation Maximization to learn how to navigate in an unsupervised manner. Their approach cannot reason about dynamic obstacles, hence it is unable to reason about interactions with people during navigation, so it does not appear in our tables.

Table \ref{tab:alg_inf} summarizes the taxonomy values for the inference algorithms for social navigation discussed in this subsection.

\subsection{Conveying the Robot's Goal to the Human}
\begin{table}[ht]
\caption{An overview of the different intention-conveying algorithms used in social navigation. \textbf{Role} refers to the robot role, \textbf{Obs.} is observability, \textbf{Com.} refers to communication, and \textbf{Exp. type} is the experiment type.}
\label{tab:alg_conv}
\footnotesize
\begin{tabular}{|r|l|l|l|l|l|l|l|l|}
\hline
\multicolumn{1}{|l|}{\textbf{Year}} & \textbf{Paper}                          &\textbf{Role} & \textbf{\# Agents} & \textbf{Obs.}                                          & \textbf{Motion Control}                                      & \textbf{Com.} & \textbf{\begin{tabular}[c]{@{}l@{}}Exp.\\ Type\end{tabular}} & \textbf{\begin{tabular}[c]{@{}l@{}}Agent\\ Type\end{tabular}} \\ \hline
2009                                & \citet{nummenmaa2009ll}                                 & I             & Abs=2             & Partial                                                & \begin{tabular}[c]{@{}l@{}}Other\\ (Hand Coded)\end{tabular} & D             & Sim                                                          & H-A                                                           \\ \hline
2013                                & \citet{fiore2013toward}                               & I             & Abs=2             & Depth                                                  & \begin{tabular}[c]{@{}l@{}}Other\\ (Hand Coded)\end{tabular} & I + D         & Sim                                                           & H-A                                                           \\ \hline
2015                                & \citet{may2015show}                                   & I             & Abs=2             & \begin{tabular}[c]{@{}l@{}}RGB +\\ Depth\end{tabular}  & \begin{tabular}[c]{@{}l@{}}Other\\ (A*)\end{tabular}         & D             & Lab                                                           & H-R                                                           \\ \hline
2015                                & \citet{Szafir:2015:CDF:2696454.2696475}                 & I             & Abs=2             & \begin{tabular}[c]{@{}l@{}}RGB +\\ Depth\end{tabular}  & \begin{tabular}[c]{@{}l@{}}Other\\ (Hand Coded)\end{tabular} & D             & Lab + Sur                                                     & H-R                                                           \\ \hline
2015                                & \citet{unhelkar2015human}                               & N             & Abs=1             & Full                                                   & \begin{tabular}[c]{@{}l@{}}Other\\ (SIPP)\end{tabular}       & N             & Sim                                                          & Hom                                                           \\ \hline
2015                                & \citet{watanabe2015communicating}                       & I             & Abs=2             & Depth                                                  & ROS                                                          & D             & Lab                                                           & H-R                                                           \\ \hline
2016                                & \citet{khambhaita2016head}                             & I             & Abs=2             & \begin{tabular}[c]{@{}l@{}}RGB + \\ Depth\end{tabular} & ROS                                                          & D             & Lab + Sur                                                     & H-R                                                           \\ \hline
2018                                & \citet{Baraka2018}                                  & I             & Abs=2             & \begin{tabular}[c]{@{}l@{}}RGB +\\ Depth\end{tabular}  & ROS                                                          & D             & Lab + Sur                                                     & H-R                                                           \\ \hline
2018                                & \citet{fernandez2018passive}                           & I             & Abs=2             & Depth                                                  & \begin{tabular}[c]{@{}l@{}}Other\\ (Hand Coded)\end{tabular} & D             & Lab                                                           & H-R                                                           \\ \hline
2018                                & \citet{lynch2018effect}                              & I             & Abs=2             & Full                                                   & \begin{tabular}[c]{@{}l@{}}Other\\ (Hand Coded)\end{tabular} & D             & Sim                                                          & H-A                                                           \\ \hline
2018                                & \citet{shrestha2018communicating}                     & I             & Abs=2             & Full                                                   & \begin{tabular}[c]{@{}l@{}}Other\\ (Hand Coded)\end{tabular} & D             & Lab + Sur                                                     & H-R                                                           \\ \hline
2020                                & \citet{hart2020hallway}                                 & I             & Abs=2             & Depth                                                  & \begin{tabular}[c]{@{}l@{}}Other\\ (Hand Coded)\end{tabular} & D             & Lab                                                           & H-R                                                           \\ \hline
\end{tabular}
\normalsize
\end{table}

\citet{dragan2013legibility} formally define the concepts of legibility (motion that allows the observer to confidently infer the correct goal) and predictability (motion that conforms with the observer's expectations) in robot navigation. They show that human-robot collaboration is affected by the way the robot plans its motion, and to perform better, the robot design should switch from a focus on predictability to a focus on legibility. This section presents several approaches to increase the robot's legibility and explicability, with an emphasis on interaction points where there is a conflict between the human pedestrian and the robot.
More details about the specific mechanisms that are activated in humans when interacting with a robot can be found by the work by \citet{sciutti2012measuring}, who survey the concept of ``motor resonance'' between an acting robot and an observing human. \citet{Ryo2021Omni} recently presented a motion planning algorithm for omni-directional robots to resemble human movements in a time-efficient manner.

Many contributions use verbal signals for guidance \citep{thrun2000probabilistic}. \citet{jeffrey1998constructing} investigate human navigational behavior in the context of two simulated environments. In these simulations, people could communicate using either text messages or audio. \citet{yedidsion2019optimal} investigate how verbal instructions given by more than one robot can assist humans in navigation in a new environment. However, for the social navigation task, verbal communication is considered less useful, as the navigation is expected to take place seamlessly without demanding the high awareness level that verbal communication requires
 \citep{cha2018survey}. 
To deal with this challenge, many contributions take inspiration from the theory of proxemics \citep{hall1966hidden} as a non-verbal way to convey intent or restriction. \citet{rios2015proxemics} investigate the comfort zone of people when a robot approaches them and  \citet{torta2013design} identify specific values for this comfort zone (182 cm from a sitting person and 173 cm from a standing person) or imitate them from a set of observed trajectories, and uses the learned model for online planning. 
 
\textbf{LED and Artificial Signals}
 \citet{Baraka2018} use an LED configuration on their CoBot to indicate a number of robot states --- including turning --- focusing on the design of LED animations to address legibility. Their study shows that the use of these signals increases participants' willingness to aid the robot. 
\citet{shrestha2018communicating} augment their robot with projection indicators to signal the robot's intended path.
 \citet{Szafir:2015:CDF:2696454.2696475} equip quad-rotor drones with LEDs mounted in a ring at the base, providing four different signal designs along this strip. They found that their LEDs improve participants' ability to quickly infer the intended motion of the drone.  \citet{8525528} perform a study in which a robot crosses a humans' path, indicating its intended path with an arrow projected onto the floor. They demonstrate their method to be effective in expressing the robot's intended trajectory.  \citet{fernandez2018passive} introduce the concept of a ``passive demonstration'', in order to disambiguate the intention of a robot's LED turn signal. \citet{watanabe2015communicating} evaluate a robotic wheelchair that autonomously navigates the environment with and without intention communication. They show that passengers and pedestrians found intention communication intuitive and helpful for passing by actions.

\textbf{Robot Gaze as Signal}
 Several contributions build on the fact that humans infer other people’s movement trajectories from their gaze direction \citep{nummenmaa2009ll}$^{R1,R2}$, and from the relationship between head pose and gaze direction \citep{kar2017review}$^{R2,R3}$.  \citet{norman2009design}$^{R2,R3}$ speculates that bicycle riders know how to avoid collisions with pedestrians since pedestrian motion can be predicted by gaze. Similarly, \citet{unhelkar2015human} found that head pose is a significant predictor of the direction that a person intends to walk.

Following a similar line of thought, \citet{khambhaita2016head} propose a motion planner which coordinates head motion to the path a robot will take $4$ seconds in the future. In a video survey in which their robot approaches a T-intersection in a hallway, they found that study participants are significantly more able to determine the intended path of the robot in terms of the left or right branch of the intersection when the robot uses the gaze cue as opposed to when it does not. 
Using a different gaze cue,  \citet{lynch2018effect}$^{R1,R2}$ perform a study in a virtual environment in which virtual agents establish mutual gaze with participants during path-crossing events in a virtual hallway, finding no significant effect in helping participants to disambiguate their paths from those of the virtual agents.

\citet{fiore2013toward} propose an analysis of human interpretation of social cues in hallway navigation. Their study design included different proxemic and gaze cues that were implemented by rotating the sensors of the robot. Their results show that  cues associated with the robot’s proxemic behavior were found to significantly affect participant perceptions of the robot’s social presence while cues associated with the robot’s gaze behavior were not found to be significant. However, \citet{fernandez2018passive} show that people are able to adapt to LED-based cues after watching a demonstration of its use, and \citet{may2015show} present a robot that was able to convey its intention using a mechanical signal but not using a gaze cue. \citet{hart2020hallway} challenge these previous results by providing a different naturalistic gaze cue using a virtual agent head which is added to a mobile robot platform, and compared its performance against a similar robot with an LED turn signal. The results of this work suggest that people are able to perceive the naturalistic gaze cue and react to it. These conflicting results can be attributed to the vast differences in signal implementation between the different experiments.

Table \ref{tab:alg_conv} summarizes the taxonomy values for algorithms that focus on conveying the robot's intention to a human.


\subsection{Mediating Conflicts in Navigational Intentions}
\begin{table}[ht]
\caption{An overview of the different mediation algorithms used in social navigation.  \textbf{Role} refers to the robot role, \textbf{Obs.} is observability, \textbf{Com.} refers to communication, and \textbf{Exp. type} is the experiment type.}
\label{tab:alg_med}
\footnotesize
\begin{tabular}{|r|l|l|l|l|l|l|l|l|}
\hline
\multicolumn{1}{|l|}{\textbf{Year}} & \textbf{Paper}                   & \textbf{Role} & \textbf{\# Agents} & \textbf{Obs.}                                         & \textbf{\begin{tabular}[c]{@{}l@{}}Motion\\ Control\end{tabular}}             & \textbf{Com.} & \textbf{\begin{tabular}[c]{@{}l@{}}Exp.\\ Type\end{tabular}} & \textbf{\begin{tabular}[c]{@{}l@{}}Agent\\ Type\end{tabular}} \\ \hline
2002                                & \citet{murakami2002collision}                  & B             & Abs=2              & \begin{tabular}[c]{@{}l@{}}RGB +\\ Depth\end{tabular} & \begin{tabular}[c]{@{}l@{}}Other\\ (Hand Coded)\end{tabular}                  & I             & Lab                                                           & H-R                                                           \\ \hline
2005                                & \citet{topp2005tracking}                       & R             & Abs=4              & Depth                                                 & \begin{tabular}[c]{@{}l@{}}Other\\ (Person \\ Tracking)\end{tabular}          & N             & Lab                                                           & H-R                                                           \\ \hline
2008                                & \citet{muller2008socially}                     & R             & Abs=7              & Depth                                                 & \begin{tabular}[c]{@{}l@{}}Other \\ (A* + \\ Person \\ Tracking)\end{tabular} & N             & Lab                                                           & H-R                                                           \\ \hline
2010                                & \citet{svenstrup2010trajectory}                 & R             & Abs=39             & Full                                                  & \begin{tabular}[c]{@{}l@{}}Other \\ (Modified\\ RRT)\end{tabular}             & N             & Sim                                                          & H-R                                                           \\ \hline
2013                                & \citet{ferrer2013social}                        & R             & Abs=10            & Depth                                                 & SFM                                                                           & N             & ItW                                                           & H-R                                                           \\ \hline
2013                                & \citet{guzzi2013human}                          & B             & Abs=6              & RGB                                                   & \begin{tabular}[c]{@{}l@{}}Other\\ (Hand Coded)\end{tabular}                  & N             & R                                                            & R-R                                                           \\ \hline
2014                                & \citet{karamouzas2014universal}                & R             & D=0.27-2.5         & Full                                                  & \begin{tabular}[c]{@{}l@{}}Other\\ (Hand Coded)\end{tabular}                  & N             & Sim                                                          & Hom                                                           \\ \hline
2014                                & \citet{kruse2014evaluating}                     & B             & Abs=2              & Full                                                  & \begin{tabular}[c]{@{}l@{}}Other\\ (Hand Coded)\end{tabular}                  & I             & Lab + Sur                                                     & H-R                                                           \\ \hline
2017                                & \citet{silva2017human}                         & B             & Abs=2              & Full                                                  & ROS                                                                           & N             & Sim                                                          & Hom                                                           \\ \hline
2019                                & \citet{yao2019following}                      & R             & Abs=6              & \begin{tabular}[c]{@{}l@{}}RGB +\\ Depth\end{tabular} & \begin{tabular}[c]{@{}l@{}}Other\\ (Geometry\\ based)\end{tabular}            & N             & Lab                                                           & H-R                                                           \\ \hline
2022                                & \citet{truc2022khaos}                       & R             & Abs=2              & \begin{tabular}[c]{@{}l@{}}Full\end{tabular} & \begin{tabular}[c]{@{}l@{}}Other\\ (Hand Coded)\end{tabular}            & N             & Sim                                                           & Het                                                           \\ \hline
\end{tabular}
\normalsize
\end{table}

\citet{karamouzas2014universal} identify a power-law interaction that is based not on the physical separation between pedestrians but on their projected time to a potential future collision, and is therefore fundamentally anticipatory in nature. This finding highlights that there is a value in understanding and mediating between the human's navigational goal and the robot's.

\citet{murakami2002collision} propose to smooth a wheelchair's trajectory to avoid colliding with pedestrians. 
\citet{kruse2012legible, kruse2014evaluating} investigate classic navigation algorithms that create erratic trajectories near obstacles that make a robot look confused. To address this challenge, they use context-dependent cost functions and directional cost functions that help a robot to solve spatial conflicts. One result, for example, is adjusting the robot's velocity instead of its path. 
\citet{silva2017human} tackle the mediation problem using the notion of motion effort and how it should be shared between the robot and the person in order to avoid collisions. To that end, their approach learns a robot behavior using Reinforcement Learning that enables it to mutually solve the collision avoidance problem during simulated trials.
\citet{svenstrup2010trajectory} propose a modified RRT for navigation in human environments assuming access to full state information. The proposed RRT planner plans with a potential field representation of the world,  with a potential model designed for moving humans. 
Alternatively, recent work by \citet{truc2022khaos} focused on drone navigation around people. This work introduced a human-aware 3D reactive planner for drone navigation. This planner is based on stochastic optimization of two criteria: discomfort due to the proximity of the drone to pedestrians, and visibility of the drone. 

A different line of research combines social navigation and person following. This combination can work in several directions: both \citet{topp2005tracking, muller2008socially} present collision avoidance algorithms that are utilized in the context of following one particular person through a populated environment. 
Alternatively, in \citet{yao2019following}, the robot leverages the planning of other pedestrians and follows them instead of searching for a solution on its own.

Table \ref{tab:alg_med} summarizes the taxonomy values for mediation algorithms for social navigation discussed in this subsection.

\section{Evaluating an Interaction}
\label{sec::evaluating}
The numerous different metrics and evaluation methods used in social navigation make apparent the need to standardize them. This section is meant to provide tools and metrics to evaluate new research in social navigation with respect to the existing literature and with our proposed taxonomy to provide context for evaluation. 
As we are surveying an interdisciplinary area, many of the metrics used so far for evaluation were adapted from other research areas (e.g. Human-Computer Interfaces, psychology, physics, mechanical engineering, and more). To pinpoint the most common and useful metrics, we discuss only the metrics that were used in the papers that were presented in the tables in Sections \ref{sec::models} and \ref{sec::algorithms}. For each metric we present, we mention the taxonomy attributes that are the most relevant and can directly affect the values of the metric. For example, measuring group formation directly depends on the \textbf{Number of Agents} in the environment, since if there is only one pedestrian it cannot form a group. Table \ref{tab:eval} summarizes this evaluation according to the different aspects of the interaction: properties of the interaction itself, actions taken by the human or the robot, emergent behaviors, algorithmic properties, and others. This last aspect includes both qualitative evaluation and prediction accuracy, which is a very common metric to estimate the proficiency of \emph{obstacle detection}, a preliminary step before the actual interaction.

\subsection{Interaction Properties}
This subsection discusses measurements that are related to the nature of the interaction itself, and are meant to evaluate how successful and efficient an interaction is. These metrics are objective, and external to the robot and the human.
\noindent\textit{Conflicts Count} is one of the most common approaches to estimate the success of an interaction. This measurement is quantified in several ways: by counting desirable outcomes vs. undesirable outcomes, by counting accidents, or by counting interactions that ended without the robot reaching its goal. In this category, we also consider experiments that counted how many times the robot was required to replan \citep{muller2008socially} and how many targets it was able to reach in total \citep{guzzi2013human}. This measure is affected by the \textbf{Number of Agents}, the \textbf{Experiment Type}, and the evaluated \textbf{Agent Type}.

\noindent\textit{Speed} is another very common metric used to evaluate an interaction. In general, faster velocities imply that the robot was able to navigate confidently without slowing down. Many researchers used this metric to complement conflict count, to account for cases where a robot may reach its goal quickly but frequently collides with walls. As a reference point, the robot's speed is usually compared to the average pedestrian speed ($1.3 \pm 0.2$ m/s), but this value depends on whether they walk alone or in a group, as group size affects speed more than density level \citep{moussaid2010walking}. \citet{gerin2008characteristics} measured similar results for natural walking around dynamic obstacles ($1.44 \pm 0.17$ m/s). Accordingly, this measurement is greatly affected by the \textbf{Robot's Role} in the interaction, the \textbf{Number of Agents}, the \textbf{Experiment Type}, and the \textbf{Agent Type}.

\noindent\textit{Path Time} is a way to measure the velocity of the robot throughout a full interaction. As the robot might accelerate or decelerate, recording the total time that it took the robot to reach its goal is a simple way to measure its performance. One unique metric that is also relevant to throughput is ``social work'', defined by \citet{ferrer2013social}. This metric measures the total work done by the robot, and the summation of the work done by each person in the scene. \citet{kanazawa2019adaptive} examined the total waiting time that the robot had experienced during the interaction.
This measure depends on the \textbf{Robot's Role}, the \textbf{Number of Agents} in the environment, and the \textbf{Experiment Type}.

\noindent\textit{Path Length} provides another perspective about the interaction, and is correlated with speed and path time: by counting any two of these three metrics (Speed, Path Time, and Path Length) one can get a reasonable estimation of the third. As such, this metric is also affected by the same attributes as the other two metrics: the \textbf{Robot's Role}, the \textbf{Number of Agents}, and the \textbf{Experiment Type}.

\noindent\textit{Acceleration} is a way to measure the changes in the robot's behavior throughout the interaction. A robot that accelerates or decelerates several times in an interaction is an indication that it had to replan or adjust to avoid a conflict. This metric is highly affected by 
the \textbf{Robot's Role}, and the \textbf{Number of Agents}.

\noindent\textit{Smoothness} is a generalization for several metrics that measure the total energy that was put into the interaction by the robot or the human. Successful interactions are expected to require less energy than unsuccessful interactions, which force the robot to replan. Smoothness can be evaluated in several ways, including acceleration/deceleration over time, total kinetic energy used \citep{park2013collision}, path irregularity (how many unnecessary turns were taken) \citep{guzzi2013human}, cumulative heading change \citep{okal2016learning}, and the integral of the square of the curvature to measure the smoothness of a pedestrian’s path \citep{karamouzas2009predictive}. This measure is influenced by 
the \textbf{Robot's Role}, the \textbf{Observability} that can enable the robot to plan better ahead, and the \textbf{Motion Control} used.

\noindent\textit{Avoidance Distance} is a way to measure how close the robot came to a conflict or a full collision with a human. Usually, a robot that is able to avoid pedestrians from afar is considered more successful than a robot that almost reaches collision \citep{svenstrup2010trajectory}. However, this success sometimes creates a tradeoff with the total length of the path the robot needs to take and the smoothness of the path. This metric is affected by 
 the \textbf{Robot's Role}, the \textbf{Number of Agents}, and the \textbf{Motion Control} used that might have its own predefined distance-keeping restrictions.

\subsection{Robot/Human Actions}
While the previous subsection considered measurements of the interaction as a whole, in this subsection we discuss measures that evaluate the actions taken by the robot or the human.

\noindent\textit{Degrees Turned} As part of an interaction, either the robot or the human (or both) turn to avoid collision. Evaluation which consists of this measurement usually tracks the degrees of the lane change of either party. This measure will be highly affected by the \textbf{Robot's Role} which will determine who will turn, the \textbf{Number of Agents} in the environment, and the \textbf{Motion Control} used.

\noindent\textit{Gaze} is a general measurement, in which several different aspects can be evaluated, including fixation count and length \citep{nummenmaa2009ll}, and the Gaze-Movement Angle (GMA) \citep{cutting1995we}. \citet{kitazawa2010pedestrian} investigated gaze patterns in a collision avoidance scenario with multiple pedestrians moving in a wide hallway shape area. They show that pedestrians pay much more attention to ground surface to detect potential immediate environmental hazards than fixating on obstacles. Therefore, most of their fixations fall within a cone-shape area rather than semicircle, and the attention paid to approaching pedestrians is not as high as that to static obstacles. Metrics that involve gaze are affected by the \textbf{Robot's Role}, \textbf{Observability}, \textbf{Communication} protocols that the human should be aware of, the \textbf{Experiment Type}, and \textbf{Agent Type} which can all have great effects on gaze patterns.

\noindent\textit{Head Orientation and Body Positions} are ways to capture some intermediate value between the degrees turned in practice, and the changes in GMA. Recently, \citet{Ryo2021Omni} leveraged people's reliance on such cues and incorporated similar body rotations into an omni-directional robot to improve the way pedestrians perceive its performance. These metrics are highly affected by the \textbf{Robot's Role} in the interaction, the \textbf{Communication} channel used, and the \textbf{Agent Type}.

\subsection{Emergent Behaviors} 
Several experiments have been designed to identify specific movement patterns and flow patterns that emerge during execution of social navigation algorithms, or to mimic human movement patterns that emerge in these contexts \citep{bennewitz2002learning,loscos2003intuitive}. In many cases, these patterns are in the form of lanes \citep{helbing1995social} or group clusters.

\begin{wrapfigure}{l}{0.5\textwidth}
    \centering
    \includegraphics[width=0.4\textwidth]{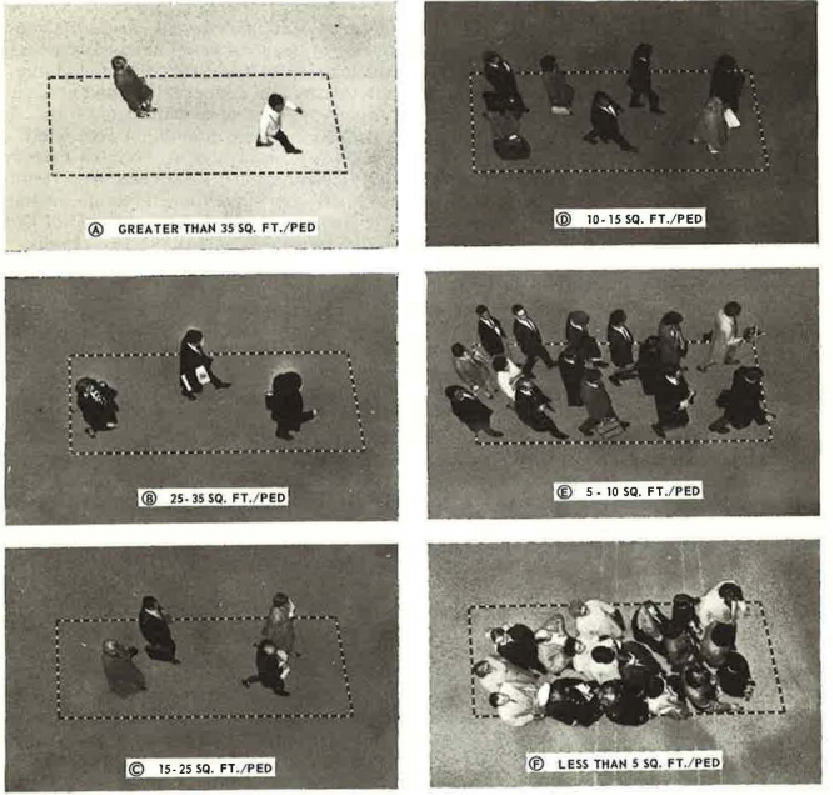}
    \caption{Levels of Service from A to F: How crowded is the environment (taken from \citet{fruin1971designing})}.
    \label{fig:LoS}
\end{wrapfigure}

\noindent\textit{Lane Emergence} is a phenomenon that exists in human crowds --- whenever an environment becomes crowded enough, it is likely that people will follow the path of others who are going in the same direction \citep{yao2019following, gockley2007natural}. For several algorithms deployed in crowded environments, the researchers were able to detect the emergence of lanes in robotic navigation context, and considered this behavior as a sign of success, since lanes are usually an efficient way to navigate in crowds. This measure is affected by the \textbf{Number of Agents}, the \textbf{Experiment Type}, and the evaluated \textbf{Agent Type}.

\noindent\textit{Group Formation} is another phenomenon whose appearance implies the success of the interaction. However, unlike lane emergence, group formation is usually an explicit objective of a work that discusses these types of interactions: such work focuses on understanding how groups of pedestrians move together \citep{moussaid2010walking}, and are investigating whether a robot can seamlessly join such a group \citep{musse1997model}, bypass it \citep{swofford2020improving}, or disperse it~\citep{chen9autonomous}. This measure is affected by the \textbf{Number of Agents} and the \textbf{Agent Type}.

\noindent\textit{Maximal Density} is a metric frequently used in simulations to stress-test an agent's ability to navigate in an environment with multiple other agents. When shifting to the real world, \citet{fruin1971designing}$^{R2,R3}$ proposed 6 levels of crowdness, which is referred to as \emph{Level of Service}, as depicted in Figure \ref{fig:LoS}. When comparing to human-only navigation, the average density of people in a non-crowded environment was evaluated to $0.03$ pedestrians per $m^2$, and in a moderately crowded environment, there are $0.25$ pedestrians per $m^2$ \citep{moussaid2010walking}.
Notice that density, or the \textbf{Number of Agents} is an attribute in this survey's taxonomy --- in this specific section, we only refer to evaluation that uses density as a metric, rather than as a controlled variable.

\subsection{Algorithmic Properties}
The previous subsections focused on measuring physical quantities, either about the interaction as a whole or about one of the parties. In this subsection, we focus on more algorithmic aspects of the interaction. The metrics presented here can often be measured internally by the robot.

\noindent\textit{Computation Time} in social navigation refers to the robot's processing time. As the robot should perform in real-time, there is a need to evaluate whether the robot can process the required information, plan, and execute its plan on time. 
Two different components that are measured by computation time are: interaction processing, which is usually measured in milliseconds \citep{van2011reciprocal}, and learning (in data-driven approaches), which is usually measured in learning episodes for achieving a desired behavior \citep{ding2018hierarchical}. Computation time is influenced by 
the \textbf{Number of Agents}, \textbf{Experiment Type} and \textbf{Agent Type}.

\noindent\textit{Model Prediction} is a crucial part of every social navigation interaction: in order to properly act, the robot should first be able to accurately predict the behavior of other agents in the environment. Some contributions focus solely on improving the part of the interaction that involves understanding the environment given sensor information, and accurately predicting trajectories \citep{kuderer2012feature, meng2019scaling}$^{R2,R3}$, while others evaluate the prediction of pedestrian trajectories interleaved with robot execution \citep{bera2017sociosense, nardi2019long}. This metric is influenced by the \textbf{Robot's Role} in the interaction, \textbf{Observability}, and \textbf{Agent Type}.

\subsection{Other Evaluations}
So far, all evaluation metrics were objective and could usually be quantitatively evaluated. Some contributions focus on analyzing an interaction and identifying theoretical concepts, thus have no empirical evaluation, while others test subjective quantities (e.g. comfort level) or provide a qualitative evaluation of an interaction.

\noindent\textit{Survey Questions} are the most common approach to elicit information from users about how they perceive an interaction with an agent or a robot. These metrics consist of comfort levels during the interaction \citep{jeffrey1998constructing,may2015show, vasquez2014inverse}, social presence \citep{pacchierotti2006design, khambhaita2016head}, expectation matching \citep{gockley2007natural, kretzschmar2016socially}, and more. With respect to comfort, \citet{torta2013design} identify specific values for this comfort zone (182 cm from a sitting person and 173 cm from a standing person).
\citet{syrdal2008video} present an empirical evaluation of the role of video prototyping and evocation as a good way to evaluate non-functional aspects of HRI. Another type of subjective evaluation is of proxemics \citep{svenstrup2010trajectory, okal2016learning}, which is related to avoidance distance that was discussed earlier, but can encompass additional information about the interaction. For example, \citet{hall1966hidden} identifies different interaction ranges: intimate space (up to $0.45$m), personal space ($1.2$m), social space ($3.6$m), and public space ($7.6$m). When mapping these distances to human-robot interactions, the comfortable distance from a robot is $0.2$m, and arrival tolerance $0.5$m \citep{chen2019crowd,kruse2013human}.  
A survey is also referred to in this survey as an \textbf{Experiment Type}, hence this is the most related attribute.

\noindent\textit{No Evaluation} is a category designated for papers that make only a theoretical contribution, such as classifying different abstract types of interactions \citep{reynolds1999steering} or ones that provide only a qualitative analysis of an interaction \citep{topp2005tracking}. Accordingly, research with no empirical evaluation might be affected by all attributes of the taxonomy, depending on the subject of the analysis.
  
 \begin{table}[]
\scriptsize
\caption{An overview of the different metrics used in the surveyed papers to evaluate a social interaction.}
\begin{tabular}{|l|l|l|}
\hline
\textbf{Evaluation Type}                & \textbf{Metric Evaluated}                                           & \textbf{Relevant Works}                                                                                                                                                                                                                                                                                                                                                                                                                                                                                                                                                                                                                                                                                                                                                                                                                                                    \\ \hline
\multirow{7}{*}{Interaction Properties} & Conflicts Count                                                     & \begin{tabular}[c]{@{}l@{}}\citet{murakami2002collision}, \citet{pacchierotti2006design}, \citet{muller2008socially}, \\ \citet{kirby2009companion},\citet{svenstrup2010trajectory}, \citet{tamura2010smooth},\\  \citet{diego2011please},\citet{bandyopadhyay2013intention}, \citet{park2013collision},\\  \citet{ma2013modeling},\citet{guzzi2013human}, \citet{unhelkar2015human}, \\ \citet{godoy2016moving},\citet{okal2016learning}, \citet{kretzschmar2016socially},\\  \citet{khambhaita2016head}, \citet{fernandez2018passive}, \citet{li2018role},\\  \citet{everett2018motion},\citet{ding2018hierarchical}, \citet{long2018towards}, \\ \citet{lynch2018effect}, \citet{ROMAN18-Jiang}, \citet{yao2019following},\\ \citet{jin2019mapless}, \citet{meng2019scaling}, \citet{chen2019crowd},\\ \citet{hart2020hallway}, \citet{liang2020crowdsteer}, \citet{lu2022socially}, \citet{gupta2022intention} \end{tabular} \\ \cline{2-3} 
                                        & Speed                                                               & \begin{tabular}[c]{@{}l@{}}\citet{helbing1995social}, \citet{gerin2008characteristics}, \citet{karamouzas2009predictive},\\  \citet{moussaid2011simple}, \citet{kruse2014evaluating}, \citet{unhelkar2015human},\\  \citet{kretzschmar2016socially}, \citet{long2018towards}, \citet{liang2020crowdsteer}\end{tabular}                                                                                                                                                                                                                                                                                                                                                                                                                                                                                                                                                     \\ \cline{2-3} 
                                        & Path Time                                                           & \begin{tabular}[c]{@{}l@{}}\citet{helbing1995social}, \citet{pacchierotti2006design}, \citet{karamouzas2009predictive}, \\ \citet{foka2010probabilistic}, \citet{bandyopadhyay2013intention}, \citet{ferrer2013social}, \\ \citet{godoy2016moving}, \citet{chen2017socially}, \citet{chen2017decentralized},\\  \citet{tai2018socially}, \citet{everett2018motion}, \citet{ding2018hierarchical}, \\ \citet{long2018towards}, \citet{ROMAN18-Jiang}, \citet{jin2019mapless}, \citet{kanazawa2019adaptive},\\  \citet{chen2019crowd}, \citet{liang2020crowdsteer}, \citet{lu2022socially}, \citet{gupta2022intention}  \end{tabular}                                                                                                                                                                                                                                                                                              \\ \cline{2-3} 
                                        & Path Length                                                         & \begin{tabular}[c]{@{}l@{}}\citet{helbing1995social}, \citet{pacchierotti2006design}, \citet{karamouzas2009predictive}\\ , \citet{henry2010learning}, \citet{luber2012socially}, \citet{rios2012navigating},\\  \citet{lu2013tuning}, \citet{vasquez2014inverse}, \citet{okal2016learning}, \\ \citet{ding2018hierarchical}, \citet{ROMAN18-Jiang}, \citet{long2018towards},\\  \citet{nardi2019long}, \citet{liang2020crowdsteer}\end{tabular}                                                                                                                                                                                                                                                                                                                                                                                                                            \\ \cline{2-3} 
                                        & Acceleration                                                        & \citet{helbing1995social}, \citet{bonneaud2014empirically}                                                                                                                                                                                                                                                                                                                                                                                                                                                                                                                                                                                                                                                                                                                                                                                                                 \\ \cline{2-3} 
                                        & Avoidance Distance                                                  & \begin{tabular}[c]{@{}l@{}}\citet{luber2012socially}, \citet{kruse2014evaluating}, \citet{may2015show}, \\ \citet{kim2016socially}, \citet{kretzschmar2016socially}, \citet{chen2017socially}, \\ \citet{tai2018socially}, \citet{lynch2018effect}, \citet{jin2019mapless},\\  \citet{kanazawa2019adaptive}, \citet{lu2022socially} \end{tabular}                                                                                                                                                                                                                                                                                                                                                                                                                                                                                                                                                  \\ \cline{2-3} 
                                        & Smoothness                                                          & \begin{tabular}[c]{@{}l@{}}\citet{helbing1995social}, \citet{gockley2007natural}, \citet{karamouzas2009predictive}, \\ \citet{park2013collision}, \citet{guzzi2013human}, \citet{vasquez2014inverse},\\  \citet{karamouzas2014universal}, \citet{okal2016learning}, \citet{truc2022khaos} \end{tabular}                                                                                                                                                                                                                                                                                                                                                                                                                                                                                                                                                                                            \\ \hline
\multirow{5}{*}{Robot / Human Actions}  & Degrees Turned                                                      & \begin{tabular}[c]{@{}l@{}}\citet{helbing1995social}, \citet{karamouzas2009predictive}, \citet{bonneaud2014empirically},\\  \citet{truong2016dynamic}\end{tabular}                                                                                                                                                                                                                                                                                                                                                                                                                                                                                                                                                                                                                                                                                                         \\ \cline{2-3} 
                                        & Gaze Fixations                                                      & \citet{nummenmaa2009ll}, \citet{kitazawa2010pedestrian}                                                                                                                                                                                                                                                                                                                                                                                                                                                                                                                                                                                                                                                                                                                                                                                                                    \\ \cline{2-3} 
                                        & Gaze-Movement Angle                                                 & \begin{tabular}[c]{@{}l@{}}\citet{murakami2002collision}, \citet{pacchierotti2006design}, \citet{muller2008socially},\\  \citet{kirby2009companion},\citet{svenstrup2010trajectory}, \citet{diego2011please},\\ \citet{bandyopadhyay2013intention}, \citet{ratsamee2013human}, \citet{park2013collision},\\  \citet{ma2013modeling},\citet{guzzi2013human}, \citet{unhelkar2015human}, \\ \citet{godoy2016moving},\citet{okal2016learning}, \citet{kretzschmar2016socially}, \\ \citet{khambhaita2016head},\citet{fernandez2018passive}, \citet{li2018role}, \\ \citet{everett2018motion},\citet{ding2018hierarchical}, \citet{long2018towards}, \\ \citet{lynch2018effect}, \citet{yao2019following},\citet{jin2019mapless},\\  \citet{meng2019scaling}, \citet{chen2019crowd},\citet{hart2020hallway}, \\ \citet{liang2020crowdsteer}\end{tabular}                       \\ \cline{2-3} 
                                        & Head Orientation                                                    & \citet{Patla1999}, \citet{ratsamee2013human}, \citet{unhelkar2015human}                                                                                                                                                                                                                                                                                                                                                                                                                                                                                                                                                                                                                                                                                                                                                                                                    \\ \cline{2-3} 
                                        & Body Position                                                       & \citet{Patla1999}, \citet{unhelkar2015human}                                                                                                                                                                                                                                                                                                                                                                                                                                                                                                                                                                                                                                                                                                                                                                                                                               \\ \hline
\multirow{3}{*}{Emergent Behaviors}     & Lane Emergence                                                      & \begin{tabular}[c]{@{}l@{}}\citet{helbing1995social}, \citet{bennewitz2002learning}, \citet{loscos2003intuitive},\\  \citet{van2011reciprocal}, \citet{karamouzas2014universal}\end{tabular}                                                                                                                                                                                                                                                                                                                                                                                                                                                                                                                                                                                                                                                                               \\ \cline{2-3} 
                                        & Group Formation                                                     & \citet{musse1997model}, \citet{moussaid2010walking}, \citet{swofford2020improving}                                                                                                                                                                                                                                                                                                                                                                                                                                                                                                                                                                                                                                                                                                                                                                                         \\ \cline{2-3} 
                                        & Maximal density                                                     & \citet{bandyopadhyay2013intention}, \citet{ma2013modeling}, \citet{mead2016perceptual}                                                                                                                                                                                                                                                                                                                                                                                                                                                                                                                                                                                                                                                                                                                                                                                     \\ \hline
\multirow{2}{*}{Algorithmic Properties} & Computation Time                                                    & \begin{tabular}[c]{@{}l@{}}\citet{sisbot2007human}, \citet{moussaid2010walking}, \citet{van2011reciprocal}, \\ \citet{silva2017human}, \citet{ding2018hierarchical}\end{tabular}                                                                                                                                                                                                                                                                                                                                                                                                                                                                                                                                                                                                                                                                                           \\ \cline{2-3} 
                                        & Model Prediction                                                    & \begin{tabular}[c]{@{}l@{}}\citet{kuderer2012feature}, \citet{okal2014towards}, \citet{kim2016socially}, \\ \citet{kretzschmar2016socially}, \citet{bera2017sociosense}, \citet{silva2017human},\\  \citet{yao2019following}, \citet{nardi2019long}, \citet{meng2019scaling}\end{tabular}                                                                                                                                                                                                                                                                                                                                                                                                                                                                                                                                                                                  \\ \hline
\multirow{2}{*}{Other}                  & Survey Questions                                                    & \begin{tabular}[c]{@{}l@{}}\citet{jeffrey1998constructing}, \citet{murakami2002collision}, \citet{pacchierotti2006design}, \\ \citet{gockley2007natural}, \citet{svenstrup2010trajectory}, \citet{vasquez2014inverse}, \\ \citet{kruse2014evaluating}, \citet{may2015show}, \citet{watanabe2015communicating},\\  \citet{Szafir:2015:CDF:2696454.2696475}, \citet{okal2016learning}, \citet{kretzschmar2016socially}, \\ \citet{khambhaita2016head}, \citet{chen2017socially}, \citet{Baraka2018}, \\ \citet{shrestha2018communicating}, \citet{senft2020would}\end{tabular}                                                                                                                                                                                                                                                                                               \\ \cline{2-3} 
                                        & \begin{tabular}[c]{@{}l@{}}No Interaction\\ Evaluation\end{tabular} & \begin{tabular}[c]{@{}l@{}}\citet{reynolds1999steering}, \citet{strassner2005virtual}, \citet{topp2005tracking},\\  \citet{ohki2010collision}, \citet{pandey2010framework}, \citet{o2011learning},\\  \citet{gomez2014fast}, \citet{papadakis2014adaptive}, \citet{charalampous2014social}\end{tabular}                                                                                                                                                                                                                                                                                                                                                                                                                                                                                                                                                                    \\ \hline
\end{tabular}
\normalsize
\label{tab:eval}
\end{table}
 
\subsection{Simulations and Resources}
So far, this section discussed specific metrics and evaluation methods that have been used in social navigation. One of the goals of this discussion is to promote better comparisons between different contributions in the field. Another way to promote this goal is by using existing simulations or resources that can have a similar baseline. In this subsection, we identify some of the recent efforts to create social navigation benchmarks and evaluation frameworks.

\citet{carton2016measuring} propose a framework for the analysis of human trajectories, and show that humans plan their navigation trajectories in a similar fashion when walking past a robot or a human.

Simulations are commonly used to evaluate of a social navigation algorithm or model ($39$ of the $75$ surveyed papers used simulations), either as a preliminary step to physical navigation or as a completely independent task. Next we point out several available simulation tools that can be used to evaluate new contributions.
 \citet{loscos2003intuitive} created a rule-based simulation that can model up to $10,000$ pedestrians in an urban environment.
\citet{treuille2006continuum} offered a real-time crowd model based on continuum dynamics, which can facilitate large-scale simulations for navigation. \citet{heigeas2010physically} presented a simulation platform where pedestrians act according to a physics-based particle  force interaction model. Recently, \citet{khambhaita2016head} created a simulated benchmark for social navigation tasks instead of physical experiments. This simulation is implemented with OpenAI Gym. \citet{tsoi2020sean} presents a testing platform that combines ROS and Unity into a social navigation testbed. In this platform's current version, it can measure whether or not the robot reaches its goal, time to goal, collisions with static objects, final distance to goal, collisions with pedestrians, and closest distance to pedestrians.

For various reasons, there are not many contributions that can generalize to real-world interactions: First, robots can only be tested under similar conditions, meaning that an evaluation platform for large mobile robots will be different from one for smaller robots. Explicitly identifying how accurate a robotic design is (e.g. 2D vs. 3D representation, joint movement, 3rd person vs. 1st person evaluation, etc.) is a key component in the design of any real-world robot experiment \citep{syrdal2008video}.
 In addition, real human-robot interactions require human presence, which introduces a lot of variability and cannot be just compiled into an algorithm that can be used repeatedly.
 
\citet{mavrogiannis2019effects} recently published a case study where people and robots navigated in a shared space. The robots used three distinct navigation strategies, executed by a telepresence robot (two autonomous, one teleoperated). The first is Optimal Reciprocal Collision Avoidance (ORCA), a local collision-free motion planner for a large number of robots as proposed by \citet{van2011reciprocal} and the second is the social momentum (SM) planning framework, which estimates the most likely intended avoidance protocols of others based on their past behaviors, superimposes them, and generates an expressive and socially compliant robot action that reinforces the expectations of others regarding these avoidance protocols \citep{mavrogiannis2018social}.  These two chosen navigational strategies are agnostic to the fact that the other agent is a human. This assumption leaves an opportunity for further investigation.  

\section{Discussion}
\label{sec::discussions}

In this survey, we identified specific components that comprise a social navigation interaction, and introduced a detailed taxonomy to provide researchers with a framework and a language for comparing and contrasting research in social navigation (Section \ref{sec::taxonomy}). We then compiled a comprehensive list of papers that contribute to social navigation and discussed them according to their values given our taxonomy (Sections \ref{sec::models} and \ref{sec::algorithms}). Next, we surveyed the different measurements used to evaluate an interaction in this context, and highlighted the relations between these measurements and the taxonomy attributes (Section \ref{sec::evaluating}).

Social navigation is a growing research area. While we expect that the attributes we chose for the taxonomy will remain relevant in the years to come, additional attributes will be added and the focus of specific work might shift to deal with new settings. However, any progress in the field must be rooted in the fundamental components of social navigation as they are presented in this survey. In addition, the proposed taxonomy can serve as a framework that enables researchers to properly place their contributions with respect to other work and to provide better benchmarks, which we hope will lead to an additional growth in this research area.

To conclude this survey and to consolidate its contributions into a coherent guide, we offer the readers the following checklist to assist with the design of social navigation interaction between a human and a robot. When introducing a new contribution to social navigation, potential aspects to consider include the following. 
\begin{enumerate}
    \item \textbf{Taxonomy;} Identify the values your work has with respect to the taxonomy's attributes in this survey: Robot's Role, Number of Agents, Observability, Motion Control, Communication, Experiment Type, and Agent Type. As shown in this survey, the values of these attributes differ greatly among different papers; thus using this taxonomy is expected to help  place new contributions within useful contexts and scopes.
    \item \textbf{Reliability;} Provide as many details as possible about the choices made in the design of the robot, and about the implementation details. For example, when reporting an absolute number of pedestrians, also report the size of the area in which the experiment was conducted.
    \item \textbf{Human Presence;} If your work consists of an interaction with pedestrians, what is their level of familiarity with the robot prior to the interaction? As presented in this survey, often experiments with human subjects are conducted in the lab rather than in the wild, where the subjects are often the roboticists who designed the robot.
    \item \textbf{Context;} Identify what is exactly the context in which the interaction takes place. As with other decisions, the context in which the chosen design is utilized can affect the behavior of pedestrians.
    \item \textbf{Success;} If your work consists of empirical evaluation, identify in advance what is considered a success in an interaction. For example, if your work introduces a new indirect communication method, the success of the evaluation should properly isolate the effect of that method.
    \item \textbf{Evaluation;} Detail which metrics will be used to evaluate this success, and what values are these metrics expected to have.  Based on the presented taxonomy and surveyed papers, evaluation can be placed in comparison to other existing work.
\end{enumerate}

While the presented taxonomy and the above checklist can be useful resources, in Section \ref{sec::taxonomy} we mentioned some additional concepts that are not yet mature enough to be included in the taxonomy, but might become more significant as the field grows. These concepts include: an analysis of different collision types, context awareness and semantic mapping, reactions to a robot vs. to a human, social cues and social signals, focused interaction, and navigating with groups of pedestrians. We see a surge of work that breaks traditional assumptions about pedestrian behavior in the context of social navigation ~\citep{chen9autonomous, murakami2021mutual, reig2020not}, and these new settings may not be reflected using the existing attributes of the social navigation taxonomy. These papers are part of a fast evolving field, in which we predict an immense growth in the next decade. It is hence a good time to gather and map the knowledge that was already acquired, so it will also be easier to identify the differences when attacking new problem domains. 

There are numerous open problems related to social navigation, given our current understanding and technological abilities: standardization of evaluation metrics and domains, context-aware navigation (workday vs. weekend), group understanding (avoid collision with a group participant), and adaptive navigation via machine learning (lifelong learning). Each of these problems offers many opportunities to leverage recent advances in machine learning, robotics, and human-robot interactions and  implement them in a social navigation context. For those interested in contributing to this research area, the above problems ought to serve as a promising starting point. More information about these problems can be found in Subsection \ref{sec::taxonomy:extraConcepts}.

To conclude, we expect the field of social navigation to gain increased popularity and lead to more real-world applications during the next decade. This survey aims to help lay the groundwork for these exciting developments by mapping existing approaches onto a novel taxonomy, and providing a context for new contributions to social navigation.


  \bibliographystyle{ACM-Reference-Format}
  \bibliography{references}


\begin{thebibliography}{166}


\ifx \showCODEN    \undefined \def \showCODEN     #1{\unskip}     \fi
\ifx \showDOI      \undefined \def \showDOI       #1{#1}\fi
\ifx \showISBNx    \undefined \def \showISBNx     #1{\unskip}     \fi
\ifx \showISBNxiii \undefined \def \showISBNxiii  #1{\unskip}     \fi
\ifx \showISSN     \undefined \def \showISSN      #1{\unskip}     \fi
\ifx \showLCCN     \undefined \def \showLCCN      #1{\unskip}     \fi
\ifx \shownote     \undefined \def \shownote      #1{#1}          \fi
\ifx \showarticletitle \undefined \def \showarticletitle #1{#1}   \fi
\ifx \showURL      \undefined \def \showURL       {\relax}        \fi
\providecommand\bibfield[2]{#2}
\providecommand\bibinfo[2]{#2}
\providecommand\natexlab[1]{#1}
\providecommand\showeprint[2][]{arXiv:#2}

\bibitem[\protect\citeauthoryear{Admoni, Bank, Tan, Toneva, and
  Scassellati}{Admoni et~al\mbox{.}}{2011}]%
        {admoni2011robot}
\bibfield{author}{\bibinfo{person}{Henny Admoni}, \bibinfo{person}{Caroline
  Bank}, \bibinfo{person}{Joshua Tan}, \bibinfo{person}{Mariya Toneva}, {and}
  \bibinfo{person}{Brian Scassellati}.} \bibinfo{year}{2011}\natexlab{}.
\newblock \showarticletitle{Robot gaze does not reflexively cue human
  attention}. In \bibinfo{booktitle}{\emph{Proceedings of the Annual Meeting of
  the Cognitive Science Society}}, Vol.~\bibinfo{volume}{33}.
\newblock


\bibitem[\protect\citeauthoryear{Admoni and Scassellati}{Admoni and
  Scassellati}{2017}]%
        {admoni2017social}
\bibfield{author}{\bibinfo{person}{Henny Admoni} {and} \bibinfo{person}{Brian
  Scassellati}.} \bibinfo{year}{2017}\natexlab{}.
\newblock \showarticletitle{Social eye gaze in human-robot interaction: a
  review}.
\newblock \bibinfo{journal}{\emph{Journal of Human-Robot Interaction}}
  \bibinfo{volume}{6}, \bibinfo{number}{1} (\bibinfo{year}{2017}),
  \bibinfo{pages}{25--63}.
\newblock


\bibitem[\protect\citeauthoryear{Bandyopadhyay, Won, Frazzoli, Hsu, Lee, and
  Rus}{Bandyopadhyay et~al\mbox{.}}{2013}]%
        {bandyopadhyay2013intention}
\bibfield{author}{\bibinfo{person}{Tirthankar Bandyopadhyay},
  \bibinfo{person}{Kok~Sung Won}, \bibinfo{person}{Emilio Frazzoli},
  \bibinfo{person}{David Hsu}, \bibinfo{person}{Wee~Sun Lee}, {and}
  \bibinfo{person}{Daniela Rus}.} \bibinfo{year}{2013}\natexlab{}.
\newblock \showarticletitle{Intention-aware motion planning}.
\newblock In \bibinfo{booktitle}{\emph{Algorithmic foundations of robotics X}}.
  \bibinfo{publisher}{Springer}, \bibinfo{pages}{475--491}.
\newblock


\bibitem[\protect\citeauthoryear{Baraka and Veloso}{Baraka and Veloso}{2018}]%
        {Baraka2018}
\bibfield{author}{\bibinfo{person}{Kim Baraka} {and}
  \bibinfo{person}{Manuela~M. Veloso}.} \bibinfo{year}{2018}\natexlab{}.
\newblock \showarticletitle{Mobile Service Robot State Revealing Through
  Expressive Lights: Formalism, Design, and Evaluation}.
\newblock \bibinfo{journal}{\emph{International Journal of Social Robotics}}
  \bibinfo{volume}{10}, \bibinfo{number}{1} (\bibinfo{date}{01 Jan}
  \bibinfo{year}{2018}), \bibinfo{pages}{65--92}.
\newblock
\showISSN{1875-4805}
\urldef\tempurl%
\url{https://doi.org/10.1007/s12369-017-0431-x}
\showDOI{\tempurl}


\bibitem[\protect\citeauthoryear{Bennewitz, Burgard, and Thrun}{Bennewitz
  et~al\mbox{.}}{2002}]%
        {bennewitz2002learning}
\bibfield{author}{\bibinfo{person}{Maren Bennewitz}, \bibinfo{person}{Wolfram
  Burgard}, {and} \bibinfo{person}{Sebastian Thrun}.}
  \bibinfo{year}{2002}\natexlab{}.
\newblock \showarticletitle{Learning motion patterns of persons for mobile
  service robots}. In \bibinfo{booktitle}{\emph{Proceedings 2002 IEEE
  International Conference on Robotics and Automation (Cat. No. 02CH37292)}},
  Vol.~\bibinfo{volume}{4}. IEEE, \bibinfo{pages}{3601--3606}.
\newblock


\bibitem[\protect\citeauthoryear{Bera, Randhavane, Prinja, and Manocha}{Bera
  et~al\mbox{.}}{2017}]%
        {bera2017sociosense}
\bibfield{author}{\bibinfo{person}{Aniket Bera}, \bibinfo{person}{Tanmay
  Randhavane}, \bibinfo{person}{Rohan Prinja}, {and} \bibinfo{person}{Dinesh
  Manocha}.} \bibinfo{year}{2017}\natexlab{}.
\newblock \showarticletitle{Sociosense: Robot navigation amongst pedestrians
  with social and psychological constraints}. In \bibinfo{booktitle}{\emph{2017
  IEEE/RSJ International Conference on Intelligent Robots and Systems (IROS)}}.
  IEEE, \bibinfo{pages}{7018--7025}.
\newblock


\bibitem[\protect\citeauthoryear{Bonabeau, Dorigo, Marco, Theraulaz,
  Th{\'e}raulaz, et~al\mbox{.}}{Bonabeau et~al\mbox{.}}{1999}]%
        {bonabeau1999swarm}
\bibfield{author}{\bibinfo{person}{Eric Bonabeau}, \bibinfo{person}{Marco
  Dorigo}, \bibinfo{person}{Directeur de Recherches Du~Fnrs Marco},
  \bibinfo{person}{Guy Theraulaz}, \bibinfo{person}{Guy Th{\'e}raulaz},
  {et~al\mbox{.}}} \bibinfo{year}{1999}\natexlab{}.
\newblock \bibinfo{booktitle}{\emph{Swarm intelligence: from natural to
  artificial systems}}.
\newblock Number~1. \bibinfo{publisher}{Oxford university press}.
\newblock


\bibitem[\protect\citeauthoryear{Bonin-Font, Ortiz, and Oliver}{Bonin-Font
  et~al\mbox{.}}{2008}]%
        {bonin2008visual}
\bibfield{author}{\bibinfo{person}{Francisco Bonin-Font},
  \bibinfo{person}{Alberto Ortiz}, {and} \bibinfo{person}{Gabriel Oliver}.}
  \bibinfo{year}{2008}\natexlab{}.
\newblock \showarticletitle{Visual navigation for mobile robots: A survey}.
\newblock \bibinfo{journal}{\emph{Journal of intelligent and robotic systems}}
  \bibinfo{volume}{53}, \bibinfo{number}{3} (\bibinfo{year}{2008}),
  \bibinfo{pages}{263--296}.
\newblock


\bibitem[\protect\citeauthoryear{Bonneaud and Warren}{Bonneaud and
  Warren}{2014}]%
        {bonneaud2014empirically}
\bibfield{author}{\bibinfo{person}{Stephane Bonneaud} {and}
  \bibinfo{person}{William~H Warren}.} \bibinfo{year}{2014}\natexlab{}.
\newblock \showarticletitle{An empirically-grounded emergent approach to
  modeling pedestrian behavior}.
\newblock In \bibinfo{booktitle}{\emph{Pedestrian and evacuation dynamics
  2012}}. \bibinfo{publisher}{Springer}, \bibinfo{pages}{625--638}.
\newblock


\bibitem[\protect\citeauthoryear{Butler and Agah}{Butler and Agah}{2001}]%
        {butler2001psychological}
\bibfield{author}{\bibinfo{person}{John~Travis Butler} {and}
  \bibinfo{person}{Arvin Agah}.} \bibinfo{year}{2001}\natexlab{}.
\newblock \showarticletitle{Psychological effects of behavior patterns of a
  mobile personal robot}.
\newblock \bibinfo{journal}{\emph{Autonomous Robots}} \bibinfo{volume}{10},
  \bibinfo{number}{2} (\bibinfo{year}{2001}), \bibinfo{pages}{185--202}.
\newblock


\bibitem[\protect\citeauthoryear{Cai, Wang, Cheng, De~Silva, and Meng}{Cai
  et~al\mbox{.}}{2020}]%
        {cai2020mobile}
\bibfield{author}{\bibinfo{person}{Kuanqi Cai}, \bibinfo{person}{Chaoqun Wang},
  \bibinfo{person}{Jiyu Cheng}, \bibinfo{person}{Clarence~W De~Silva}, {and}
  \bibinfo{person}{Max Q-H Meng}.} \bibinfo{year}{2020}\natexlab{}.
\newblock \showarticletitle{Mobile robot path planning in dynamic environments:
  a survey}.
\newblock \bibinfo{journal}{\emph{arXiv preprint arXiv:2006.14195}}
  (\bibinfo{year}{2020}).
\newblock


\bibitem[\protect\citeauthoryear{Carel}{Carel}{1961}]%
        {carel1961visual}
\bibfield{author}{\bibinfo{person}{WL Carel}.} \bibinfo{year}{1961}\natexlab{}.
\newblock \showarticletitle{Visual factors in the contact analog (Publication
  R61ELC60)}.
\newblock \bibinfo{journal}{\emph{Ithaca, NY: General Electric Company Advanced
  Electronics Center}} (\bibinfo{year}{1961}).
\newblock


\bibitem[\protect\citeauthoryear{Carton, Olszowy, and Wollherr}{Carton
  et~al\mbox{.}}{2016}]%
        {carton2016measuring}
\bibfield{author}{\bibinfo{person}{Daniel Carton}, \bibinfo{person}{Wiktor
  Olszowy}, {and} \bibinfo{person}{Dirk Wollherr}.}
  \bibinfo{year}{2016}\natexlab{}.
\newblock \showarticletitle{Measuring the effectiveness of readability for
  mobile robot locomotion}.
\newblock \bibinfo{journal}{\emph{International Journal of Social Robotics}}
  \bibinfo{volume}{8}, \bibinfo{number}{5} (\bibinfo{year}{2016}),
  \bibinfo{pages}{721--741}.
\newblock


\bibitem[\protect\citeauthoryear{Cassandra, Kaelbling, and Kurien}{Cassandra
  et~al\mbox{.}}{1996}]%
        {cassandra1996acting}
\bibfield{author}{\bibinfo{person}{Anthony~R Cassandra},
  \bibinfo{person}{Leslie~Pack Kaelbling}, {and} \bibinfo{person}{James~A
  Kurien}.} \bibinfo{year}{1996}\natexlab{}.
\newblock \showarticletitle{Acting under uncertainty: Discrete Bayesian models
  for mobile-robot navigation}. In \bibinfo{booktitle}{\emph{Proceedings of
  IEEE/RSJ International Conference on Intelligent Robots and Systems.
  IROS'96}}, Vol.~\bibinfo{volume}{2}. IEEE, \bibinfo{pages}{963--972}.
\newblock


\bibitem[\protect\citeauthoryear{Castiello}{Castiello}{2003}]%
        {castiello2003understanding}
\bibfield{author}{\bibinfo{person}{Umberto Castiello}.}
  \bibinfo{year}{2003}\natexlab{}.
\newblock \showarticletitle{Understanding other people's actions: Intention and
  attention.}
\newblock \bibinfo{journal}{\emph{Journal of Experimental Psychology: Human
  Perception and Performance}} \bibinfo{volume}{29}, \bibinfo{number}{2}
  (\bibinfo{year}{2003}), \bibinfo{pages}{416}.
\newblock


\bibitem[\protect\citeauthoryear{Cha, Kim, Fong, Mataric, et~al\mbox{.}}{Cha
  et~al\mbox{.}}{2018}]%
        {cha2018survey}
\bibfield{author}{\bibinfo{person}{Elizabeth Cha}, \bibinfo{person}{Yunkyung
  Kim}, \bibinfo{person}{Terrence Fong}, \bibinfo{person}{Maja~J Mataric},
  {et~al\mbox{.}}} \bibinfo{year}{2018}\natexlab{}.
\newblock \showarticletitle{A survey of nonverbal signaling methods for
  non-humanoid robots}.
\newblock \bibinfo{journal}{\emph{Foundations and Trends{\textregistered} in
  Robotics}} \bibinfo{volume}{6}, \bibinfo{number}{4} (\bibinfo{year}{2018}),
  \bibinfo{pages}{211--323}.
\newblock


\bibitem[\protect\citeauthoryear{Charalampous, Emmanouilidis, and
  Gasteratos}{Charalampous et~al\mbox{.}}{2014}]%
        {charalampous2014social}
\bibfield{author}{\bibinfo{person}{Konstantinos Charalampous},
  \bibinfo{person}{Christos Emmanouilidis}, {and} \bibinfo{person}{Antonios
  Gasteratos}.} \bibinfo{year}{2014}\natexlab{}.
\newblock \showarticletitle{Social mapping on RGB-D scenes}. In
  \bibinfo{booktitle}{\emph{2014 IEEE International Conference on Imaging
  Systems and Techniques (IST) Proceedings}}. IEEE, \bibinfo{pages}{398--403}.
\newblock


\bibitem[\protect\citeauthoryear{Charalampous, Kostavelis, and
  Gasteratos}{Charalampous et~al\mbox{.}}{2017}]%
        {charalampous2017recent}
\bibfield{author}{\bibinfo{person}{Konstantinos Charalampous},
  \bibinfo{person}{Ioannis Kostavelis}, {and} \bibinfo{person}{Antonios
  Gasteratos}.} \bibinfo{year}{2017}\natexlab{}.
\newblock \showarticletitle{Recent trends in social aware robot navigation: A
  survey}.
\newblock \bibinfo{journal}{\emph{Robotics and Autonomous Systems}}
  \bibinfo{volume}{93} (\bibinfo{year}{2017}), \bibinfo{pages}{85--104}.
\newblock


\bibitem[\protect\citeauthoryear{Chen, Liu, Kreiss, and Alahi}{Chen
  et~al\mbox{.}}{2019}]%
        {chen2019crowd}
\bibfield{author}{\bibinfo{person}{Changan Chen}, \bibinfo{person}{Yuejiang
  Liu}, \bibinfo{person}{Sven Kreiss}, {and} \bibinfo{person}{Alexandre
  Alahi}.} \bibinfo{year}{2019}\natexlab{}.
\newblock \showarticletitle{Crowd-robot interaction: Crowd-aware robot
  navigation with attention-based deep reinforcement learning}. In
  \bibinfo{booktitle}{\emph{2019 International Conference on Robotics and
  Automation (ICRA)}}. IEEE, \bibinfo{pages}{6015--6022}.
\newblock


\bibitem[\protect\citeauthoryear{Chen, Everett, Liu, and How}{Chen
  et~al\mbox{.}}{2017a}]%
        {chen2017socially}
\bibfield{author}{\bibinfo{person}{Yu~Fan Chen}, \bibinfo{person}{Michael
  Everett}, \bibinfo{person}{Miao Liu}, {and} \bibinfo{person}{Jonathan~P
  How}.} \bibinfo{year}{2017}\natexlab{a}.
\newblock \showarticletitle{Socially aware motion planning with deep
  reinforcement learning}. In \bibinfo{booktitle}{\emph{2017 IEEE/RSJ
  International Conference on Intelligent Robots and Systems (IROS)}}. IEEE,
  \bibinfo{pages}{1343--1350}.
\newblock


\bibitem[\protect\citeauthoryear{Chen, Liu, Everett, and How}{Chen
  et~al\mbox{.}}{2017b}]%
        {chen2017decentralized}
\bibfield{author}{\bibinfo{person}{Yu~Fan Chen}, \bibinfo{person}{Miao Liu},
  \bibinfo{person}{Michael Everett}, {and} \bibinfo{person}{Jonathan~P How}.}
  \bibinfo{year}{2017}\natexlab{b}.
\newblock \showarticletitle{Decentralized non-communicating multiagent
  collision avoidance with deep reinforcement learning}. In
  \bibinfo{booktitle}{\emph{2017 IEEE international conference on robotics and
  automation (ICRA)}}. IEEE, \bibinfo{pages}{285--292}.
\newblock


\bibitem[\protect\citeauthoryear{Chen, Fan, Zhao, Liang, Shen, Chen, Manocha,
  Pan, and Zhang}{Chen et~al\mbox{.}}{2021}]%
        {chen9autonomous}
\bibfield{author}{\bibinfo{person}{Zhiming Chen}, \bibinfo{person}{Tingxiang
  Fan}, \bibinfo{person}{Xuan Zhao}, \bibinfo{person}{Jing Liang},
  \bibinfo{person}{Cong Shen}, \bibinfo{person}{Hua Chen},
  \bibinfo{person}{Dinesh Manocha}, \bibinfo{person}{Jia Pan}, {and}
  \bibinfo{person}{Wei Zhang}.} \bibinfo{year}{2021}\natexlab{}.
\newblock \showarticletitle{Autonomous Social Distancing in Urban Environments
  Using a Quadruped Robot}.
\newblock \bibinfo{journal}{\emph{IEEE Access}}  \bibinfo{volume}{9}
  (\bibinfo{year}{2021}), \bibinfo{pages}{8392--8403}.
\newblock


\bibitem[\protect\citeauthoryear{Coleman and James}{Coleman and James}{1961}]%
        {coleman1961equilibrium}
\bibfield{author}{\bibinfo{person}{James~S Coleman} {and} \bibinfo{person}{John
  James}.} \bibinfo{year}{1961}\natexlab{}.
\newblock \showarticletitle{The equilibrium size distribution of freely-forming
  groups}.
\newblock \bibinfo{journal}{\emph{Sociometry}} \bibinfo{volume}{24},
  \bibinfo{number}{1} (\bibinfo{year}{1961}), \bibinfo{pages}{36--45}.
\newblock


\bibitem[\protect\citeauthoryear{Crespo, Castillo, Mozos, and Barber}{Crespo
  et~al\mbox{.}}{2020}]%
        {crespo2020semantic}
\bibfield{author}{\bibinfo{person}{Jonathan Crespo},
  \bibinfo{person}{Jose~Carlos Castillo}, \bibinfo{person}{Oscar~Martinez
  Mozos}, {and} \bibinfo{person}{Ramon Barber}.}
  \bibinfo{year}{2020}\natexlab{}.
\newblock \showarticletitle{Semantic information for robot navigation: A
  survey}.
\newblock \bibinfo{journal}{\emph{Applied Sciences}} \bibinfo{volume}{10},
  \bibinfo{number}{2} (\bibinfo{year}{2020}), \bibinfo{pages}{497}.
\newblock


\bibitem[\protect\citeauthoryear{Cui, Zhang, Allievi, Stone, Niekum, and
  Knox}{Cui et~al\mbox{.}}{2020}]%
        {cui2020empathic}
\bibfield{author}{\bibinfo{person}{Yuchen Cui}, \bibinfo{person}{Qiping Zhang},
  \bibinfo{person}{Alessandro Allievi}, \bibinfo{person}{Peter Stone},
  \bibinfo{person}{Scott Niekum}, {and} \bibinfo{person}{W~Bradley Knox}.}
  \bibinfo{year}{2020}\natexlab{}.
\newblock \showarticletitle{The EMPATHIC Framework for Task Learning from
  Implicit Human Feedback}.
\newblock \bibinfo{journal}{\emph{arXiv preprint arXiv:2009.13649}}
  (\bibinfo{year}{2020}).
\newblock


\bibitem[\protect\citeauthoryear{Cutting, Vishton, and Braren}{Cutting
  et~al\mbox{.}}{1995}]%
        {cutting1995we}
\bibfield{author}{\bibinfo{person}{James~E Cutting}, \bibinfo{person}{Peter~M
  Vishton}, {and} \bibinfo{person}{Paul~A Braren}.}
  \bibinfo{year}{1995}\natexlab{}.
\newblock \showarticletitle{How we avoid collisions with stationary and moving
  objects.}
\newblock \bibinfo{journal}{\emph{Psychological review}} \bibinfo{volume}{102},
  \bibinfo{number}{4} (\bibinfo{year}{1995}), \bibinfo{pages}{627}.
\newblock


\bibitem[\protect\citeauthoryear{DeSouza and Kak}{DeSouza and Kak}{2002}]%
        {desouza2002vision}
\bibfield{author}{\bibinfo{person}{Guilherme~N DeSouza} {and}
  \bibinfo{person}{Avinash~C Kak}.} \bibinfo{year}{2002}\natexlab{}.
\newblock \showarticletitle{Vision for mobile robot navigation: A survey}.
\newblock \bibinfo{journal}{\emph{IEEE transactions on pattern analysis and
  machine intelligence}} \bibinfo{volume}{24}, \bibinfo{number}{2}
  (\bibinfo{year}{2002}), \bibinfo{pages}{237--267}.
\newblock


\bibitem[\protect\citeauthoryear{Diego and Arras}{Diego and Arras}{2011}]%
        {diego2011please}
\bibfield{author}{\bibinfo{person}{Gian Diego} {and} \bibinfo{person}{Tipaldi
  Kai~O Arras}.} \bibinfo{year}{2011}\natexlab{}.
\newblock \showarticletitle{Please do not disturb! minimum interference
  coverage for social robots}. In \bibinfo{booktitle}{\emph{2011 IEEE/RSJ
  International Conference on Intelligent Robots and Systems}}. IEEE,
  \bibinfo{pages}{1968--1973}.
\newblock


\bibitem[\protect\citeauthoryear{Ding, Li, Qian, and Chen}{Ding
  et~al\mbox{.}}{2018}]%
        {ding2018hierarchical}
\bibfield{author}{\bibinfo{person}{Wenhao Ding}, \bibinfo{person}{Shuaijun Li},
  \bibinfo{person}{Huihuan Qian}, {and} \bibinfo{person}{Yongquan Chen}.}
  \bibinfo{year}{2018}\natexlab{}.
\newblock \showarticletitle{Hierarchical reinforcement learning framework
  towards multi-agent navigation}. In \bibinfo{booktitle}{\emph{2018 IEEE
  International Conference on Robotics and Biomimetics (ROBIO)}}. IEEE,
  \bibinfo{pages}{237--242}.
\newblock


\bibitem[\protect\citeauthoryear{Dragan, Lee, and Srinivasa}{Dragan
  et~al\mbox{.}}{2013}]%
        {dragan2013legibility}
\bibfield{author}{\bibinfo{person}{Anca~D Dragan}, \bibinfo{person}{Kenton~CT
  Lee}, {and} \bibinfo{person}{Siddhartha~S Srinivasa}.}
  \bibinfo{year}{2013}\natexlab{}.
\newblock \showarticletitle{Legibility and predictability of robot motion}. In
  \bibinfo{booktitle}{\emph{2013 8th ACM/IEEE International Conference on
  Human-Robot Interaction (HRI)}}. IEEE, \bibinfo{pages}{301--308}.
\newblock


\bibitem[\protect\citeauthoryear{Ess, Leibe, Schindler, and Van~Gool}{Ess
  et~al\mbox{.}}{2009}]%
        {ess2009moving}
\bibfield{author}{\bibinfo{person}{Andreas Ess}, \bibinfo{person}{Bastian
  Leibe}, \bibinfo{person}{Konrad Schindler}, {and} \bibinfo{person}{Luc
  Van~Gool}.} \bibinfo{year}{2009}\natexlab{}.
\newblock \showarticletitle{Moving obstacle detection in highly dynamic
  scenes}. In \bibinfo{booktitle}{\emph{2009 IEEE International Conference on
  Robotics and Automation}}. IEEE, \bibinfo{pages}{56--63}.
\newblock


\bibitem[\protect\citeauthoryear{Everett, Chen, and How}{Everett
  et~al\mbox{.}}{2018}]%
        {everett2018motion}
\bibfield{author}{\bibinfo{person}{Michael Everett}, \bibinfo{person}{Yu~Fan
  Chen}, {and} \bibinfo{person}{Jonathan~P How}.}
  \bibinfo{year}{2018}\natexlab{}.
\newblock \showarticletitle{Motion planning among dynamic, decision-making
  agents with deep reinforcement learning}. In \bibinfo{booktitle}{\emph{2018
  IEEE/RSJ International Conference on Intelligent Robots and Systems (IROS)}}.
  IEEE, \bibinfo{pages}{3052--3059}.
\newblock


\bibitem[\protect\citeauthoryear{Fernandez, John, Kirmani, Hart, Sinapov, and
  Stone}{Fernandez et~al\mbox{.}}{2018}]%
        {fernandez2018passive}
\bibfield{author}{\bibinfo{person}{Rolando Fernandez}, \bibinfo{person}{Nathan
  John}, \bibinfo{person}{Sean Kirmani}, \bibinfo{person}{Justin Hart},
  \bibinfo{person}{Jivko Sinapov}, {and} \bibinfo{person}{Peter Stone}.}
  \bibinfo{year}{2018}\natexlab{}.
\newblock \showarticletitle{Passive Demonstrations of Light-Based Robot Signals
  for Improved Human Interpretability}. In \bibinfo{booktitle}{\emph{2018 27th
  IEEE International Symposium on Robot and Human Interactive Communication
  (RO-MAN)}}. IEEE, \bibinfo{pages}{234--239}.
\newblock


\bibitem[\protect\citeauthoryear{Ferrer, Garrell, and Sanfeliu}{Ferrer
  et~al\mbox{.}}{2013}]%
        {ferrer2013social}
\bibfield{author}{\bibinfo{person}{Gonzalo Ferrer}, \bibinfo{person}{Anais
  Garrell}, {and} \bibinfo{person}{Alberto Sanfeliu}.}
  \bibinfo{year}{2013}\natexlab{}.
\newblock \showarticletitle{Social-aware robot navigation in urban
  environments}. In \bibinfo{booktitle}{\emph{2013 European Conference on
  Mobile Robots}}. IEEE, \bibinfo{pages}{331--336}.
\newblock


\bibitem[\protect\citeauthoryear{Fiore, Wiltshire, Lobato, Jentsch, Huang, and
  Axelrod}{Fiore et~al\mbox{.}}{2013}]%
        {fiore2013toward}
\bibfield{author}{\bibinfo{person}{Stephen~M Fiore}, \bibinfo{person}{Travis~J
  Wiltshire}, \bibinfo{person}{Emilio~JC Lobato}, \bibinfo{person}{Florian~G
  Jentsch}, \bibinfo{person}{Wesley~H Huang}, {and} \bibinfo{person}{Benjamin
  Axelrod}.} \bibinfo{year}{2013}\natexlab{}.
\newblock \showarticletitle{Toward understanding social cues and signals in
  human--robot interaction: effects of robot gaze and proxemic behavior}.
\newblock \bibinfo{journal}{\emph{Frontiers in psychology}}
  \bibinfo{volume}{4} (\bibinfo{year}{2013}), \bibinfo{pages}{859}.
\newblock


\bibitem[\protect\citeauthoryear{Foka and Trahanias}{Foka and
  Trahanias}{2010}]%
        {foka2010probabilistic}
\bibfield{author}{\bibinfo{person}{Amalia~F Foka} {and}
  \bibinfo{person}{Panos~E Trahanias}.} \bibinfo{year}{2010}\natexlab{}.
\newblock \showarticletitle{Probabilistic autonomous robot navigation in
  dynamic environments with human motion prediction}.
\newblock \bibinfo{journal}{\emph{International Journal of Social Robotics}}
  \bibinfo{volume}{2}, \bibinfo{number}{1} (\bibinfo{year}{2010}),
  \bibinfo{pages}{79--94}.
\newblock


\bibitem[\protect\citeauthoryear{Fong, Nourbakhsh, and Dautenhahn}{Fong
  et~al\mbox{.}}{2003}]%
        {fong2003survey}
\bibfield{author}{\bibinfo{person}{Terrence Fong}, \bibinfo{person}{Illah
  Nourbakhsh}, {and} \bibinfo{person}{Kerstin Dautenhahn}.}
  \bibinfo{year}{2003}\natexlab{}.
\newblock \showarticletitle{A survey of socially interactive robots}.
\newblock \bibinfo{journal}{\emph{Robotics and autonomous systems}}
  \bibinfo{volume}{42}, \bibinfo{number}{3-4} (\bibinfo{year}{2003}),
  \bibinfo{pages}{143--166}.
\newblock


\bibitem[\protect\citeauthoryear{Fruin}{Fruin}{1971}]%
        {fruin1971designing}
\bibfield{author}{\bibinfo{person}{John~J Fruin}.}
  \bibinfo{year}{1971}\natexlab{}.
\newblock \bibinfo{booktitle}{\emph{Designing for pedestrians: A
  level-of-service concept}}.
\newblock Number HS-011 999.
\newblock


\bibitem[\protect\citeauthoryear{Gaber, Marey, Amin, and Tolba}{Gaber
  et~al\mbox{.}}{2017}]%
        {gaber2017localization}
\bibfield{author}{\bibinfo{person}{Heba Gaber}, \bibinfo{person}{Mohamed
  Marey}, \bibinfo{person}{Safaa Amin}, {and} \bibinfo{person}{Mohamed~F
  Tolba}.} \bibinfo{year}{2017}\natexlab{}.
\newblock \showarticletitle{Localization and Mapping for Indoor Navigation:
  Survey}.
\newblock \bibinfo{journal}{\emph{Handbook of Research on Machine Learning
  Innovations and Trends}} (\bibinfo{year}{2017}), \bibinfo{pages}{136--160}.
\newblock


\bibitem[\protect\citeauthoryear{Gao and Huang}{Gao and Huang}{2021}]%
        {gao2021evaluation}
\bibfield{author}{\bibinfo{person}{Yuxiang Gao} {and}
  \bibinfo{person}{Chien-Ming Huang}.} \bibinfo{year}{2021}\natexlab{}.
\newblock \showarticletitle{Evaluation of socially-aware robot navigation}.
\newblock \bibinfo{journal}{\emph{Frontiers in Robotics and AI}}
  (\bibinfo{year}{2021}), \bibinfo{pages}{420}.
\newblock


\bibitem[\protect\citeauthoryear{Garg, S{\"u}nderhauf, Dayoub, Morrison,
  Cosgun, Carneiro, Wu, Chin, Reid, Gould, et~al\mbox{.}}{Garg
  et~al\mbox{.}}{2020}]%
        {garg2020semantics}
\bibfield{author}{\bibinfo{person}{Sourav Garg}, \bibinfo{person}{Niko
  S{\"u}nderhauf}, \bibinfo{person}{Feras Dayoub}, \bibinfo{person}{Douglas
  Morrison}, \bibinfo{person}{Akansel Cosgun}, \bibinfo{person}{Gustavo
  Carneiro}, \bibinfo{person}{Qi Wu}, \bibinfo{person}{Tat-Jun Chin},
  \bibinfo{person}{Ian Reid}, \bibinfo{person}{Stephen Gould}, {et~al\mbox{.}}}
  \bibinfo{year}{2020}\natexlab{}.
\newblock \showarticletitle{Semantics for robotic mapping, perception and
  interaction: A survey}.
\newblock \bibinfo{journal}{\emph{Foundations and Trends{\textregistered} in
  Robotics}} \bibinfo{volume}{8}, \bibinfo{number}{1--2}
  (\bibinfo{year}{2020}), \bibinfo{pages}{1--224}.
\newblock


\bibitem[\protect\citeauthoryear{G{\'e}rin-Lajoie, Richards, Fung, and
  McFadyen}{G{\'e}rin-Lajoie et~al\mbox{.}}{2008}]%
        {gerin2008characteristics}
\bibfield{author}{\bibinfo{person}{Martin G{\'e}rin-Lajoie},
  \bibinfo{person}{Carol~L Richards}, \bibinfo{person}{Joyce Fung}, {and}
  \bibinfo{person}{Bradford~J McFadyen}.} \bibinfo{year}{2008}\natexlab{}.
\newblock \showarticletitle{Characteristics of personal space during obstacle
  circumvention in physical and virtual environments}.
\newblock \bibinfo{journal}{\emph{Gait \& posture}} \bibinfo{volume}{27},
  \bibinfo{number}{2} (\bibinfo{year}{2008}), \bibinfo{pages}{239--247}.
\newblock


\bibitem[\protect\citeauthoryear{Gockley, Forlizzi, and Simmons}{Gockley
  et~al\mbox{.}}{2007}]%
        {gockley2007natural}
\bibfield{author}{\bibinfo{person}{Rachel Gockley}, \bibinfo{person}{Jodi
  Forlizzi}, {and} \bibinfo{person}{Reid Simmons}.}
  \bibinfo{year}{2007}\natexlab{}.
\newblock \showarticletitle{Natural person-following behavior for social
  robots}. In \bibinfo{booktitle}{\emph{Proceedings of the ACM/IEEE
  international conference on Human-robot interaction}}.
  \bibinfo{pages}{17--24}.
\newblock


\bibitem[\protect\citeauthoryear{Godoy, Karamouzas, Guy, and Gini}{Godoy
  et~al\mbox{.}}{2016}]%
        {godoy2016moving}
\bibfield{author}{\bibinfo{person}{Julio Godoy}, \bibinfo{person}{Ioannis
  Karamouzas}, \bibinfo{person}{Stephen~J Guy}, {and} \bibinfo{person}{Maria~L
  Gini}.} \bibinfo{year}{2016}\natexlab{}.
\newblock \showarticletitle{Moving in a Crowd: Safe and Efficient Navigation
  among Heterogeneous Agents.}. In \bibinfo{booktitle}{\emph{IJCAI}}.
  \bibinfo{pages}{294--300}.
\newblock


\bibitem[\protect\citeauthoryear{Goffman}{Goffman}{2008}]%
        {goffman2008behavior}
\bibfield{author}{\bibinfo{person}{Erving Goffman}.}
  \bibinfo{year}{2008}\natexlab{}.
\newblock \bibinfo{booktitle}{\emph{Behavior in public places}}.
\newblock \bibinfo{publisher}{Simon and Schuster}.
\newblock


\bibitem[\protect\citeauthoryear{G{\'o}mez, Mavridis, and Garrido}{G{\'o}mez
  et~al\mbox{.}}{2014}]%
        {gomez2014fast}
\bibfield{author}{\bibinfo{person}{Javier~V G{\'o}mez},
  \bibinfo{person}{Nikolaos Mavridis}, {and} \bibinfo{person}{Santiago
  Garrido}.} \bibinfo{year}{2014}\natexlab{}.
\newblock \showarticletitle{Fast marching solution for the social path planning
  problem}. In \bibinfo{booktitle}{\emph{2014 IEEE International Conference on
  Robotics and Automation (ICRA)}}. IEEE, \bibinfo{pages}{1871--1876}.
\newblock


\bibitem[\protect\citeauthoryear{Gupta, Hayes, and Sunberg}{Gupta
  et~al\mbox{.}}{2022}]%
        {gupta2022intention}
\bibfield{author}{\bibinfo{person}{Himanshu Gupta}, \bibinfo{person}{Bradley
  Hayes}, {and} \bibinfo{person}{Zachary Sunberg}.}
  \bibinfo{year}{2022}\natexlab{}.
\newblock \showarticletitle{Intention-Aware Navigation in Crowds with
  Extended-Space POMDP Planning}. In \bibinfo{booktitle}{\emph{Proceedings of
  the 21st International Conference on Autonomous Agents and Multiagent
  Systems}}. \bibinfo{pages}{562--570}.
\newblock


\bibitem[\protect\citeauthoryear{Guzzi, Giusti, Gambardella, Theraulaz, and
  Di~Caro}{Guzzi et~al\mbox{.}}{2013}]%
        {guzzi2013human}
\bibfield{author}{\bibinfo{person}{J{\'e}r{\^o}me Guzzi},
  \bibinfo{person}{Alessandro Giusti}, \bibinfo{person}{Luca~M Gambardella},
  \bibinfo{person}{Guy Theraulaz}, {and} \bibinfo{person}{Gianni~A Di~Caro}.}
  \bibinfo{year}{2013}\natexlab{}.
\newblock \showarticletitle{Human-friendly robot navigation in dynamic
  environments}. In \bibinfo{booktitle}{\emph{2013 IEEE International
  Conference on Robotics and Automation}}. IEEE, \bibinfo{pages}{423--430}.
\newblock


\bibitem[\protect\citeauthoryear{Hall}{Hall}{1966}]%
        {hall1966hidden}
\bibfield{author}{\bibinfo{person}{Edward~Twitchell Hall}.}
  \bibinfo{year}{1966}\natexlab{}.
\newblock \bibinfo{booktitle}{\emph{The hidden dimension}}.
  Vol.~\bibinfo{volume}{609}.
\newblock \bibinfo{publisher}{Garden City, NY: Doubleday}.
\newblock


\bibitem[\protect\citeauthoryear{Hart, Mirsky, Xiao, Tejeda, Mahajan, Goo,
  Baldauf, Owen, and Stone}{Hart et~al\mbox{.}}{2020}]%
        {hart2020hallway}
\bibfield{author}{\bibinfo{person}{Justin Hart}, \bibinfo{person}{Reuth
  Mirsky}, \bibinfo{person}{Xuesu Xiao}, \bibinfo{person}{Stone Tejeda},
  \bibinfo{person}{Bonny Mahajan}, \bibinfo{person}{Jamin Goo},
  \bibinfo{person}{Kathryn Baldauf}, \bibinfo{person}{Sydney Owen}, {and}
  \bibinfo{person}{Peter Stone}.} \bibinfo{year}{2020}\natexlab{}.
\newblock \showarticletitle{Using Human-Inspired Signals to Disambiguate
  Navigational Intentions}. In \bibinfo{booktitle}{\emph{International
  Conference on Social Robotics}}. Springer, \bibinfo{pages}{320--331}.
\newblock


\bibitem[\protect\citeauthoryear{Hayduk}{Hayduk}{1981}]%
        {hayduk1981shape}
\bibfield{author}{\bibinfo{person}{Leslie~A Hayduk}.}
  \bibinfo{year}{1981}\natexlab{}.
\newblock \showarticletitle{The shape of personal space: An experimental
  investigation.}
\newblock \bibinfo{journal}{\emph{Canadian Journal of Behavioural Science/Revue
  canadienne des sciences du comportement}} \bibinfo{volume}{13},
  \bibinfo{number}{1} (\bibinfo{year}{1981}), \bibinfo{pages}{87}.
\newblock


\bibitem[\protect\citeauthoryear{He{\"\i}geas, Luciani, Thollot, and
  Castagn{\'e}}{He{\"\i}geas et~al\mbox{.}}{2010}]%
        {heigeas2010physically}
\bibfield{author}{\bibinfo{person}{Laure He{\"\i}geas}, \bibinfo{person}{Annie
  Luciani}, \bibinfo{person}{Joelle Thollot}, {and} \bibinfo{person}{Nicolas
  Castagn{\'e}}.} \bibinfo{year}{2010}\natexlab{}.
\newblock \showarticletitle{A physically-based particle model of emergent crowd
  behaviors}.
\newblock \bibinfo{journal}{\emph{arXiv preprint arXiv:1005.4405}}
  (\bibinfo{year}{2010}).
\newblock


\bibitem[\protect\citeauthoryear{Helbing and Molnar}{Helbing and
  Molnar}{1995}]%
        {helbing1995social}
\bibfield{author}{\bibinfo{person}{Dirk Helbing} {and} \bibinfo{person}{Peter
  Molnar}.} \bibinfo{year}{1995}\natexlab{}.
\newblock \showarticletitle{Social force model for pedestrian dynamics}.
\newblock \bibinfo{journal}{\emph{Physical review E}} \bibinfo{volume}{51},
  \bibinfo{number}{5} (\bibinfo{year}{1995}), \bibinfo{pages}{4282}.
\newblock


\bibitem[\protect\citeauthoryear{Henry, Vollmer, Ferris, and Fox}{Henry
  et~al\mbox{.}}{2010}]%
        {henry2010learning}
\bibfield{author}{\bibinfo{person}{Peter Henry}, \bibinfo{person}{Christian
  Vollmer}, \bibinfo{person}{Brian Ferris}, {and} \bibinfo{person}{Dieter
  Fox}.} \bibinfo{year}{2010}\natexlab{}.
\newblock \showarticletitle{Learning to navigate through crowded environments}.
  In \bibinfo{booktitle}{\emph{2010 IEEE International Conference on Robotics
  and Automation}}. IEEE, \bibinfo{pages}{981--986}.
\newblock


\bibitem[\protect\citeauthoryear{Holman, Anwar, Singh, Tec, Hart, and
  Stone}{Holman et~al\mbox{.}}{2021}]%
        {holman2021watch}
\bibfield{author}{\bibinfo{person}{Blake Holman}, \bibinfo{person}{Abrar
  Anwar}, \bibinfo{person}{Akash Singh}, \bibinfo{person}{Mauricio Tec},
  \bibinfo{person}{Justin Hart}, {and} \bibinfo{person}{Peter Stone}.}
  \bibinfo{year}{2021}\natexlab{}.
\newblock \showarticletitle{Watch Where You’re Going! Gaze and Head
  Orientation as Predictors for Social Robot Navigation}. In
  \bibinfo{booktitle}{\emph{Proceedings of the IEEE International Conference on
  Robotics and Automation (ICRA)}}. IEEE, \bibinfo{pages}{6183--6190}.
\newblock


\bibitem[\protect\citeauthoryear{Hoogendoorn and Bovy}{Hoogendoorn and
  Bovy}{2003}]%
        {hoogendoorn2003simulation}
\bibfield{author}{\bibinfo{person}{Serge Hoogendoorn} {and}
  \bibinfo{person}{Piet Bovy}.} \bibinfo{year}{2003}\natexlab{}.
\newblock \showarticletitle{Simulation of pedestrian flows by optimal control
  and differential games}.
\newblock \bibinfo{journal}{\emph{Optimal control applications and methods}}
  \bibinfo{volume}{24}, \bibinfo{number}{3} (\bibinfo{year}{2003}),
  \bibinfo{pages}{153--172}.
\newblock


\bibitem[\protect\citeauthoryear{Jeffrey and Mark}{Jeffrey and Mark}{1998}]%
        {jeffrey1998constructing}
\bibfield{author}{\bibinfo{person}{Phillip Jeffrey} {and}
  \bibinfo{person}{Gloria Mark}.} \bibinfo{year}{1998}\natexlab{}.
\newblock \showarticletitle{Constructing social spaces in virtual environments:
  A study of navigation and interaction}. In \bibinfo{booktitle}{\emph{Workshop
  on personalised and social navigation in information space}}. Stockholm:
  Swedish Institute of Computer Science, \bibinfo{pages}{24--38}.
\newblock


\bibitem[\protect\citeauthoryear{Jiang, Warnell, Munera, and Stone}{Jiang
  et~al\mbox{.}}{2018b}]%
        {ROMAN18-Jiang}
\bibfield{author}{\bibinfo{person}{Yu-Sian Jiang}, \bibinfo{person}{Garrett
  Warnell}, \bibinfo{person}{Eduardo Munera}, {and} \bibinfo{person}{Peter
  Stone}.} \bibinfo{year}{2018}\natexlab{b}.
\newblock \showarticletitle{A Study of Human-Robot Copilot Systems for En-Route
  Destination Changing}. In \bibinfo{booktitle}{\emph{Proceedings of the 27th
  IEEE International Conference on Robot and Human Interactive Communication
  (RO-MAN2018)}}. \bibinfo{address}{Nanjing, China}.
\newblock
\urldef\tempurl%
\url{http://www.cs.utexas.edu/users/ai-lab/?ROMAN18-Jiang}
\showURL{%
\tempurl}


\bibitem[\protect\citeauthoryear{Jiang, Warnell, and Stone}{Jiang
  et~al\mbox{.}}{2018a}]%
        {ICMI18-Jiang}
\bibfield{author}{\bibinfo{person}{Yu-Sian Jiang}, \bibinfo{person}{Garrett
  Warnell}, {and} \bibinfo{person}{Peter Stone}.}
  \bibinfo{year}{2018}\natexlab{a}.
\newblock \showarticletitle{Inferring User Intention using Gaze in Vehicles}.
  In \bibinfo{booktitle}{\emph{The 20th ACM International Conference on
  Multimodal Interaction (ICMI)}}. \bibinfo{address}{Boulder, Colorado}.
\newblock
\urldef\tempurl%
\url{http://www.cs.utexas.edu/users/ai-lab/?ICMI18-Jiang}
\showURL{%
\tempurl}


\bibitem[\protect\citeauthoryear{Jin, Nguyen, Sakib, Graves, Yao, and
  Jagersand}{Jin et~al\mbox{.}}{2019}]%
        {jin2019mapless}
\bibfield{author}{\bibinfo{person}{Jun Jin}, \bibinfo{person}{Nhat~M Nguyen},
  \bibinfo{person}{Nazmus Sakib}, \bibinfo{person}{Daniel Graves},
  \bibinfo{person}{Hengshuai Yao}, {and} \bibinfo{person}{Martin Jagersand}.}
  \bibinfo{year}{2019}\natexlab{}.
\newblock \showarticletitle{Mapless Navigation among Dynamics with
  Social-safety-awareness: a reinforcement learning approach from 2D laser
  scans}.
\newblock \bibinfo{journal}{\emph{arXiv preprint arXiv:1911.03074}}
  (\bibinfo{year}{2019}).
\newblock


\bibitem[\protect\citeauthoryear{Kaelbling}{Kaelbling}{2020}]%
        {kaelbling2020foundation}
\bibfield{author}{\bibinfo{person}{Leslie~Pack Kaelbling}.}
  \bibinfo{year}{2020}\natexlab{}.
\newblock \showarticletitle{The foundation of efficient robot learning}.
\newblock \bibinfo{journal}{\emph{Science}} \bibinfo{volume}{369},
  \bibinfo{number}{6506} (\bibinfo{year}{2020}), \bibinfo{pages}{915--916}.
\newblock


\bibitem[\protect\citeauthoryear{Kanazawa, Kinugawa, and Kosuge}{Kanazawa
  et~al\mbox{.}}{2019}]%
        {kanazawa2019adaptive}
\bibfield{author}{\bibinfo{person}{Akira Kanazawa}, \bibinfo{person}{Jun
  Kinugawa}, {and} \bibinfo{person}{Kazuhiro Kosuge}.}
  \bibinfo{year}{2019}\natexlab{}.
\newblock \showarticletitle{Adaptive motion planning for a collaborative robot
  based on prediction uncertainty to enhance human safety and work efficiency}.
\newblock \bibinfo{journal}{\emph{IEEE Transactions on Robotics}}
  \bibinfo{volume}{35}, \bibinfo{number}{4} (\bibinfo{year}{2019}),
  \bibinfo{pages}{817--832}.
\newblock


\bibitem[\protect\citeauthoryear{Kar and Corcoran}{Kar and Corcoran}{2017}]%
        {kar2017review}
\bibfield{author}{\bibinfo{person}{Anuradha Kar} {and} \bibinfo{person}{Peter
  Corcoran}.} \bibinfo{year}{2017}\natexlab{}.
\newblock \showarticletitle{A review and analysis of eye-gaze estimation
  systems, algorithms and performance evaluation methods in consumer
  platforms}.
\newblock \bibinfo{journal}{\emph{IEEE Access}}  \bibinfo{volume}{5}
  (\bibinfo{year}{2017}), \bibinfo{pages}{16495--16519}.
\newblock


\bibitem[\protect\citeauthoryear{Karamouzas, Heil, Van~Beek, and
  Overmars}{Karamouzas et~al\mbox{.}}{2009}]%
        {karamouzas2009predictive}
\bibfield{author}{\bibinfo{person}{Ioannis Karamouzas}, \bibinfo{person}{Peter
  Heil}, \bibinfo{person}{Pascal Van~Beek}, {and} \bibinfo{person}{Mark~H
  Overmars}.} \bibinfo{year}{2009}\natexlab{}.
\newblock \showarticletitle{A predictive collision avoidance model for
  pedestrian simulation}. In \bibinfo{booktitle}{\emph{International workshop
  on motion in games}}. Springer, \bibinfo{pages}{41--52}.
\newblock


\bibitem[\protect\citeauthoryear{Karamouzas, Skinner, and Guy}{Karamouzas
  et~al\mbox{.}}{2014}]%
        {karamouzas2014universal}
\bibfield{author}{\bibinfo{person}{Ioannis Karamouzas}, \bibinfo{person}{Brian
  Skinner}, {and} \bibinfo{person}{Stephen~J Guy}.}
  \bibinfo{year}{2014}\natexlab{}.
\newblock \showarticletitle{Universal power law governing pedestrian
  interactions}.
\newblock \bibinfo{journal}{\emph{Physical review letters}}
  \bibinfo{volume}{113}, \bibinfo{number}{23} (\bibinfo{year}{2014}),
  \bibinfo{pages}{238701}.
\newblock


\bibitem[\protect\citeauthoryear{Karnan, Nair, Xiao, Warnell, Pirk, Toshev,
  Hart, Biswas, and Stone}{Karnan et~al\mbox{.}}{2022}]%
        {karnan2022socially}
\bibfield{author}{\bibinfo{person}{Haresh Karnan}, \bibinfo{person}{Anirudh
  Nair}, \bibinfo{person}{Xuesu Xiao}, \bibinfo{person}{Garrett Warnell},
  \bibinfo{person}{Soeren Pirk}, \bibinfo{person}{Alexander Toshev},
  \bibinfo{person}{Justin Hart}, \bibinfo{person}{Joydeep Biswas}, {and}
  \bibinfo{person}{Peter Stone}.} \bibinfo{year}{2022}\natexlab{}.
\newblock \showarticletitle{Socially Compliant Navigation Dataset (SCAND): A
  Large-Scale Dataset of Demonstrations for Social Navigation}.
\newblock \bibinfo{journal}{\emph{IEEE Robotics and Automation Letters}}
  (\bibinfo{year}{2022}).
\newblock


\bibitem[\protect\citeauthoryear{Karunarathne, Morales, Kanda, and
  Ishiguro}{Karunarathne et~al\mbox{.}}{2018}]%
        {karunarathne2018model}
\bibfield{author}{\bibinfo{person}{Deneth Karunarathne},
  \bibinfo{person}{Yoichi Morales}, \bibinfo{person}{Takayuki Kanda}, {and}
  \bibinfo{person}{Hiroshi Ishiguro}.} \bibinfo{year}{2018}\natexlab{}.
\newblock \showarticletitle{Model of side-by-side walking without the robot
  knowing the goal}.
\newblock \bibinfo{journal}{\emph{International Journal of Social Robotics}}
  \bibinfo{volume}{10}, \bibinfo{number}{4} (\bibinfo{year}{2018}),
  \bibinfo{pages}{401--420}.
\newblock


\bibitem[\protect\citeauthoryear{Khambhaita, Rios-Martinez, and
  Alami}{Khambhaita et~al\mbox{.}}{2016}]%
        {khambhaita2016head}
\bibfield{author}{\bibinfo{person}{Harmish Khambhaita}, \bibinfo{person}{Jorge
  Rios-Martinez}, {and} \bibinfo{person}{Rachid Alami}.}
  \bibinfo{year}{2016}\natexlab{}.
\newblock \showarticletitle{Head-Body Motion Coordination for Human Aware Robot
  Navigation}. In \bibinfo{booktitle}{\emph{9th International workshop on
  Human-Friendly Robotics (HFR 2016)}}. \bibinfo{pages}{8p}.
\newblock


\bibitem[\protect\citeauthoryear{Kim and Pineau}{Kim and Pineau}{2016}]%
        {kim2016socially}
\bibfield{author}{\bibinfo{person}{Beomjoon Kim} {and} \bibinfo{person}{Joelle
  Pineau}.} \bibinfo{year}{2016}\natexlab{}.
\newblock \showarticletitle{Socially adaptive path planning in human
  environments using inverse reinforcement learning}.
\newblock \bibinfo{journal}{\emph{International Journal of Social Robotics}}
  \bibinfo{volume}{8}, \bibinfo{number}{1} (\bibinfo{year}{2016}),
  \bibinfo{pages}{51--66}.
\newblock


\bibitem[\protect\citeauthoryear{Kirby, Simmons, and Forlizzi}{Kirby
  et~al\mbox{.}}{2009}]%
        {kirby2009companion}
\bibfield{author}{\bibinfo{person}{Rachel Kirby}, \bibinfo{person}{Reid
  Simmons}, {and} \bibinfo{person}{Jodi Forlizzi}.}
  \bibinfo{year}{2009}\natexlab{}.
\newblock \showarticletitle{Companion: A constraint-optimizing method for
  person-acceptable navigation}. In \bibinfo{booktitle}{\emph{RO-MAN 2009-The
  18th IEEE International Symposium on Robot and Human Interactive
  Communication}}. IEEE, \bibinfo{pages}{607--612}.
\newblock


\bibitem[\protect\citeauthoryear{Kit}{Kit}{2012}]%
        {kit2012change}
\bibfield{author}{\bibinfo{person}{Dmitry~Mark Kit}.}
  \bibinfo{year}{2012}\natexlab{}.
\newblock \emph{\bibinfo{title}{Change detection models for mobile cameras}}.
\newblock \bibinfo{thesistype}{Ph.D. Dissertation}.
\newblock


\bibitem[\protect\citeauthoryear{Kitagawa, Liu, and Kanda}{Kitagawa
  et~al\mbox{.}}{2021}]%
        {Ryo2021Omni}
\bibfield{author}{\bibinfo{person}{Ryo Kitagawa}, \bibinfo{person}{Yuyi Liu},
  {and} \bibinfo{person}{Takayuki Kanda}.} \bibinfo{year}{2021}\natexlab{}.
\newblock \showarticletitle{Human-Inspired Motion Planning for Omni-Directional
  Social Robots}. In \bibinfo{booktitle}{\emph{Proceedings of the 2021 ACM/IEEE
  International Conference on Human-Robot Interaction}}
  \emph{(\bibinfo{series}{HRI '21})}. \bibinfo{publisher}{Association for
  Computing Machinery}, \bibinfo{address}{New York, NY, USA},
  \bibinfo{pages}{34–42}.
\newblock
\showISBNx{9781450382892}
\urldef\tempurl%
\url{https://doi.org/10.1145/3434073.3444679}
\showDOI{\tempurl}


\bibitem[\protect\citeauthoryear{Kitazawa and Fujiyama}{Kitazawa and
  Fujiyama}{2010}]%
        {kitazawa2010pedestrian}
\bibfield{author}{\bibinfo{person}{Kay Kitazawa} {and} \bibinfo{person}{Taku
  Fujiyama}.} \bibinfo{year}{2010}\natexlab{}.
\newblock \showarticletitle{Pedestrian vision and collision avoidance behavior:
  Investigation of the information process space of pedestrians using an eye
  tracker}.
\newblock In \bibinfo{booktitle}{\emph{Pedestrian and evacuation dynamics
  2008}}. \bibinfo{publisher}{Springer}, \bibinfo{pages}{95--108}.
\newblock


\bibitem[\protect\citeauthoryear{Kostavelis and Gasteratos}{Kostavelis and
  Gasteratos}{2015}]%
        {kostavelis2015semantic}
\bibfield{author}{\bibinfo{person}{Ioannis Kostavelis} {and}
  \bibinfo{person}{Antonios Gasteratos}.} \bibinfo{year}{2015}\natexlab{}.
\newblock \showarticletitle{Semantic mapping for mobile robotics tasks: A
  survey}.
\newblock \bibinfo{journal}{\emph{Robotics and Autonomous Systems}}
  \bibinfo{volume}{66} (\bibinfo{year}{2015}), \bibinfo{pages}{86--103}.
\newblock


\bibitem[\protect\citeauthoryear{Kretzschmar, Spies, Sprunk, and
  Burgard}{Kretzschmar et~al\mbox{.}}{2016}]%
        {kretzschmar2016socially}
\bibfield{author}{\bibinfo{person}{Henrik Kretzschmar}, \bibinfo{person}{Markus
  Spies}, \bibinfo{person}{Christoph Sprunk}, {and} \bibinfo{person}{Wolfram
  Burgard}.} \bibinfo{year}{2016}\natexlab{}.
\newblock \showarticletitle{Socially compliant mobile robot navigation via
  inverse reinforcement learning}.
\newblock \bibinfo{journal}{\emph{The International Journal of Robotics
  Research}} \bibinfo{volume}{35}, \bibinfo{number}{11} (\bibinfo{year}{2016}),
  \bibinfo{pages}{1289--1307}.
\newblock


\bibitem[\protect\citeauthoryear{Kruse, Basili, Glasauer, and Kirsch}{Kruse
  et~al\mbox{.}}{2012}]%
        {kruse2012legible}
\bibfield{author}{\bibinfo{person}{Thibault Kruse}, \bibinfo{person}{Patrizia
  Basili}, \bibinfo{person}{Stefan Glasauer}, {and} \bibinfo{person}{Alexandra
  Kirsch}.} \bibinfo{year}{2012}\natexlab{}.
\newblock \showarticletitle{Legible robot navigation in the proximity of moving
  humans}. In \bibinfo{booktitle}{\emph{2012 IEEE Workshop on Advanced Robotics
  and its Social Impacts (ARSO)}}. IEEE, \bibinfo{pages}{83--88}.
\newblock


\bibitem[\protect\citeauthoryear{Kruse, Kirsch, Khambhaita, and Alami}{Kruse
  et~al\mbox{.}}{2014}]%
        {kruse2014evaluating}
\bibfield{author}{\bibinfo{person}{Thibault Kruse}, \bibinfo{person}{Alexandra
  Kirsch}, \bibinfo{person}{Harmish Khambhaita}, {and} \bibinfo{person}{Rachid
  Alami}.} \bibinfo{year}{2014}\natexlab{}.
\newblock \showarticletitle{Evaluating directional cost models in navigation}.
  In \bibinfo{booktitle}{\emph{Proceedings of the 2014 ACM/IEEE international
  conference on Human-robot interaction}}. \bibinfo{pages}{350--357}.
\newblock


\bibitem[\protect\citeauthoryear{Kruse, Pandey, Alami, and Kirsch}{Kruse
  et~al\mbox{.}}{2013}]%
        {kruse2013human}
\bibfield{author}{\bibinfo{person}{Thibault Kruse}, \bibinfo{person}{Amit~Kumar
  Pandey}, \bibinfo{person}{Rachid Alami}, {and} \bibinfo{person}{Alexandra
  Kirsch}.} \bibinfo{year}{2013}\natexlab{}.
\newblock \showarticletitle{Human-aware robot navigation: A survey}.
\newblock \bibinfo{journal}{\emph{Robotics and Autonomous Systems}}
  \bibinfo{volume}{61}, \bibinfo{number}{12} (\bibinfo{year}{2013}),
  \bibinfo{pages}{1726--1743}.
\newblock


\bibitem[\protect\citeauthoryear{Kuderer, Kretzschmar, Sprunk, and
  Burgard}{Kuderer et~al\mbox{.}}{2012}]%
        {kuderer2012feature}
\bibfield{author}{\bibinfo{person}{Markus Kuderer}, \bibinfo{person}{Henrik
  Kretzschmar}, \bibinfo{person}{Christoph Sprunk}, {and}
  \bibinfo{person}{Wolfram Burgard}.} \bibinfo{year}{2012}\natexlab{}.
\newblock \showarticletitle{Feature-based prediction of trajectories for
  socially compliant navigation.}. In \bibinfo{booktitle}{\emph{Robotics:
  science and systems}}.
\newblock


\bibitem[\protect\citeauthoryear{Li, Jiang, Ge, and Lee}{Li
  et~al\mbox{.}}{2018}]%
        {li2018role}
\bibfield{author}{\bibinfo{person}{Mingming Li}, \bibinfo{person}{Rui Jiang},
  \bibinfo{person}{Shuzhi~Sam Ge}, {and} \bibinfo{person}{Tong~Heng Lee}.}
  \bibinfo{year}{2018}\natexlab{}.
\newblock \showarticletitle{Role playing learning for socially concomitant
  mobile robot navigation}.
\newblock \bibinfo{journal}{\emph{CAAI Transactions on Intelligence
  Technology}} \bibinfo{volume}{3}, \bibinfo{number}{1} (\bibinfo{year}{2018}),
  \bibinfo{pages}{49--58}.
\newblock


\bibitem[\protect\citeauthoryear{Liang, Patel, Sathyamoorthy, and
  Manocha}{Liang et~al\mbox{.}}{2020}]%
        {liang2020crowdsteer}
\bibfield{author}{\bibinfo{person}{Jing Liang}, \bibinfo{person}{Utsav Patel},
  \bibinfo{person}{Adarsh~Jagan Sathyamoorthy}, {and} \bibinfo{person}{Dinesh
  Manocha}.} \bibinfo{year}{2020}\natexlab{}.
\newblock \showarticletitle{CrowdSteer: Realtime Smooth and Collision-Free
  Robot Navigation in Dense Crowd Scenarios Trained using High-Fidelity
  Simulation}.
\newblock \bibinfo{journal}{\emph{International Joint Conference on AI}}
  (\bibinfo{year}{2020}).
\newblock


\bibitem[\protect\citeauthoryear{Long, Fanl, Liao, Liu, Zhang, and Pan}{Long
  et~al\mbox{.}}{2018}]%
        {long2018towards}
\bibfield{author}{\bibinfo{person}{Pinxin Long}, \bibinfo{person}{Tingxiang
  Fanl}, \bibinfo{person}{Xinyi Liao}, \bibinfo{person}{Wenxi Liu},
  \bibinfo{person}{Hao Zhang}, {and} \bibinfo{person}{Jia Pan}.}
  \bibinfo{year}{2018}\natexlab{}.
\newblock \showarticletitle{Towards optimally decentralized multi-robot
  collision avoidance via deep reinforcement learning}. In
  \bibinfo{booktitle}{\emph{2018 IEEE International Conference on Robotics and
  Automation (ICRA)}}. IEEE, \bibinfo{pages}{6252--6259}.
\newblock


\bibitem[\protect\citeauthoryear{L{\'o}pez, Alvarez, and {\'A}lvarez}{L{\'o}pez
  et~al\mbox{.}}{2019}]%
        {lopez2019walking}
\bibfield{author}{\bibinfo{person}{Antonio~M L{\'o}pez},
  \bibinfo{person}{Juan~C Alvarez}, {and} \bibinfo{person}{Diego {\'A}lvarez}.}
  \bibinfo{year}{2019}\natexlab{}.
\newblock \showarticletitle{Walking turn prediction from upper body kinematics:
  A systematic review with implications for human-robot interaction}.
\newblock \bibinfo{journal}{\emph{Applied Sciences}} \bibinfo{volume}{9},
  \bibinfo{number}{3} (\bibinfo{year}{2019}), \bibinfo{pages}{361}.
\newblock


\bibitem[\protect\citeauthoryear{Loscos, Marchal, and Meyer}{Loscos
  et~al\mbox{.}}{2003}]%
        {loscos2003intuitive}
\bibfield{author}{\bibinfo{person}{Celine Loscos}, \bibinfo{person}{David
  Marchal}, {and} \bibinfo{person}{Alexandre Meyer}.}
  \bibinfo{year}{2003}\natexlab{}.
\newblock \showarticletitle{Intuitive crowd behavior in dense urban
  environments using local laws}. In \bibinfo{booktitle}{\emph{Proceedings of
  Theory and Practice of Computer Graphics, 2003.}} IEEE,
  \bibinfo{pages}{122--129}.
\newblock


\bibitem[\protect\citeauthoryear{Lu, Allan, and Smart}{Lu
  et~al\mbox{.}}{2013}]%
        {lu2013tuning}
\bibfield{author}{\bibinfo{person}{David~V Lu}, \bibinfo{person}{Daniel~B
  Allan}, {and} \bibinfo{person}{William~D Smart}.}
  \bibinfo{year}{2013}\natexlab{}.
\newblock \showarticletitle{Tuning cost functions for social navigation}. In
  \bibinfo{booktitle}{\emph{International Conference on Social Robotics}}.
  Springer, \bibinfo{pages}{442--451}.
\newblock


\bibitem[\protect\citeauthoryear{Lu, Woo, Faragasso, Yamashita, and Asama}{Lu
  et~al\mbox{.}}{2022}]%
        {lu2022socially}
\bibfield{author}{\bibinfo{person}{Xiaojun Lu}, \bibinfo{person}{Hanwool Woo},
  \bibinfo{person}{Angela Faragasso}, \bibinfo{person}{Atsushi Yamashita},
  {and} \bibinfo{person}{Hajime Asama}.} \bibinfo{year}{2022}\natexlab{}.
\newblock \showarticletitle{Socially aware robot navigation in crowds via deep
  reinforcement learning with resilient reward functions}.
\newblock \bibinfo{journal}{\emph{Advanced Robotics}} \bibinfo{volume}{36},
  \bibinfo{number}{8} (\bibinfo{year}{2022}), \bibinfo{pages}{388--403}.
\newblock


\bibitem[\protect\citeauthoryear{Luber, Spinello, Silva, and Arras}{Luber
  et~al\mbox{.}}{2012}]%
        {luber2012socially}
\bibfield{author}{\bibinfo{person}{Matthias Luber}, \bibinfo{person}{Luciano
  Spinello}, \bibinfo{person}{Jens Silva}, {and} \bibinfo{person}{Kai~O
  Arras}.} \bibinfo{year}{2012}\natexlab{}.
\newblock \showarticletitle{Socially-aware robot navigation: A learning
  approach}. In \bibinfo{booktitle}{\emph{2012 IEEE/RSJ International
  Conference on Intelligent Robots and Systems}}. IEEE,
  \bibinfo{pages}{902--907}.
\newblock


\bibitem[\protect\citeauthoryear{Lynch, Pettr{\'e}, Bruneau, Kulpa,
  Cr{\'e}tual, and Olivier}{Lynch et~al\mbox{.}}{2018}]%
        {lynch2018effect}
\bibfield{author}{\bibinfo{person}{Sean~D Lynch}, \bibinfo{person}{Julien
  Pettr{\'e}}, \bibinfo{person}{Julien Bruneau}, \bibinfo{person}{Richard
  Kulpa}, \bibinfo{person}{Armel Cr{\'e}tual}, {and}
  \bibinfo{person}{Anne-H{\'e}l{\`e}ne Olivier}.}
  \bibinfo{year}{2018}\natexlab{}.
\newblock \showarticletitle{Effect of virtual human gaze behaviour during an
  orthogonal collision avoidance walking task}. In
  \bibinfo{booktitle}{\emph{2018 IEEE Conference on Virtual Reality and 3D User
  Interfaces (VR)}}. IEEE, \bibinfo{pages}{136--142}.
\newblock


\bibitem[\protect\citeauthoryear{Ma, Lo, Song, Wang, Zhang, and Liao}{Ma
  et~al\mbox{.}}{2013}]%
        {ma2013modeling}
\bibfield{author}{\bibinfo{person}{J Ma}, \bibinfo{person}{Siu~Ming Lo},
  \bibinfo{person}{WG Song}, \bibinfo{person}{WL Wang}, \bibinfo{person}{J
  Zhang}, {and} \bibinfo{person}{GX Liao}.} \bibinfo{year}{2013}\natexlab{}.
\newblock \showarticletitle{Modeling pedestrian space in complex building for
  efficient pedestrian traffic simulation}.
\newblock \bibinfo{journal}{\emph{Automation in Construction}}
  \bibinfo{volume}{30} (\bibinfo{year}{2013}), \bibinfo{pages}{25--36}.
\newblock


\bibitem[\protect\citeauthoryear{Martins, Rocha, Pais, and Menezes}{Martins
  et~al\mbox{.}}{2019}]%
        {martins2019clusternav}
\bibfield{author}{\bibinfo{person}{Gon{\c{c}}alo~S Martins},
  \bibinfo{person}{Rui~P Rocha}, \bibinfo{person}{Fernando~J Pais}, {and}
  \bibinfo{person}{Paulo Menezes}.} \bibinfo{year}{2019}\natexlab{}.
\newblock \showarticletitle{Clusternav: Learning-based robust navigation
  operating in cluttered environments}. In \bibinfo{booktitle}{\emph{2019
  International Conference on Robotics and Automation (ICRA)}}. IEEE,
  \bibinfo{pages}{9624--9630}.
\newblock


\bibitem[\protect\citeauthoryear{Mavrogiannis, Baldini, Wang, Zhao, Trautman,
  Steinfeld, and Oh}{Mavrogiannis et~al\mbox{.}}{2021}]%
        {mavrogiannis2021core}
\bibfield{author}{\bibinfo{person}{Christoforos Mavrogiannis},
  \bibinfo{person}{Francesca Baldini}, \bibinfo{person}{Allan Wang},
  \bibinfo{person}{Dapeng Zhao}, \bibinfo{person}{Pete Trautman},
  \bibinfo{person}{Aaron Steinfeld}, {and} \bibinfo{person}{Jean Oh}.}
  \bibinfo{year}{2021}\natexlab{}.
\newblock \showarticletitle{Core challenges of social robot navigation: A
  survey}.
\newblock \bibinfo{journal}{\emph{arXiv preprint arXiv:2103.05668}}
  (\bibinfo{year}{2021}).
\newblock


\bibitem[\protect\citeauthoryear{Mavrogiannis, Hutchinson, Macdonald,
  Alves-Oliveira, and Knepper}{Mavrogiannis et~al\mbox{.}}{2019}]%
        {mavrogiannis2019effects}
\bibfield{author}{\bibinfo{person}{Christoforos Mavrogiannis},
  \bibinfo{person}{Alena~M Hutchinson}, \bibinfo{person}{John Macdonald},
  \bibinfo{person}{Patr{\'\i}cia Alves-Oliveira}, {and} \bibinfo{person}{Ross~A
  Knepper}.} \bibinfo{year}{2019}\natexlab{}.
\newblock \showarticletitle{Effects of distinct robot navigation strategies on
  human behavior in a crowded environment}. In \bibinfo{booktitle}{\emph{2019
  14th ACM/IEEE International Conference on Human-Robot Interaction (HRI)}}.
  IEEE, \bibinfo{pages}{421--430}.
\newblock


\bibitem[\protect\citeauthoryear{Mavrogiannis, Thomason, and
  Knepper}{Mavrogiannis et~al\mbox{.}}{2018}]%
        {mavrogiannis2018social}
\bibfield{author}{\bibinfo{person}{Christoforos~I Mavrogiannis},
  \bibinfo{person}{Wil~B Thomason}, {and} \bibinfo{person}{Ross~A Knepper}.}
  \bibinfo{year}{2018}\natexlab{}.
\newblock \showarticletitle{Social momentum: A framework for legible navigation
  in dynamic multi-agent environments}. In
  \bibinfo{booktitle}{\emph{Proceedings of the 2018 ACM/IEEE International
  Conference on Human-Robot Interaction}}. \bibinfo{pages}{361--369}.
\newblock


\bibitem[\protect\citeauthoryear{May, Dondrup, and Hanheide}{May
  et~al\mbox{.}}{2015}]%
        {may2015show}
\bibfield{author}{\bibinfo{person}{Alyxander~David May},
  \bibinfo{person}{Christian Dondrup}, {and} \bibinfo{person}{Marc Hanheide}.}
  \bibinfo{year}{2015}\natexlab{}.
\newblock \showarticletitle{Show me your moves! Conveying navigation intention
  of a mobile robot to humans}. In \bibinfo{booktitle}{\emph{2015 European
  Conference on Mobile Robots (ECMR)}}. IEEE, \bibinfo{pages}{1--6}.
\newblock


\bibitem[\protect\citeauthoryear{Mead and Matari{\'c}}{Mead and
  Matari{\'c}}{2016}]%
        {mead2016perceptual}
\bibfield{author}{\bibinfo{person}{Ross Mead} {and} \bibinfo{person}{Maja~J
  Matari{\'c}}.} \bibinfo{year}{2016}\natexlab{}.
\newblock \showarticletitle{Perceptual models of human-robot proxemics}. In
  \bibinfo{booktitle}{\emph{Experimental Robotics}}. Springer,
  \bibinfo{pages}{261--276}.
\newblock


\bibitem[\protect\citeauthoryear{Meng, Ratliff, Xiang, and Fox}{Meng
  et~al\mbox{.}}{2020}]%
        {meng2019scaling}
\bibfield{author}{\bibinfo{person}{Xiangyun Meng}, \bibinfo{person}{Nathan
  Ratliff}, \bibinfo{person}{Yu Xiang}, {and} \bibinfo{person}{Dieter Fox}.}
  \bibinfo{year}{2020}\natexlab{}.
\newblock \showarticletitle{Scaling Local Control to Large-Scale Topological
  Navigation}.
\newblock \bibinfo{journal}{\emph{IEEE International Conference on Robotics and
  Automation (ICRA)}} (\bibinfo{year}{2020}).
\newblock


\bibitem[\protect\citeauthoryear{M{\"o}ller, Furnari, Battiato, H{\"a}rm{\"a},
  and Farinella}{M{\"o}ller et~al\mbox{.}}{2021}]%
        {moller2021survey}
\bibfield{author}{\bibinfo{person}{Ronja M{\"o}ller}, \bibinfo{person}{Antonino
  Furnari}, \bibinfo{person}{Sebastiano Battiato}, \bibinfo{person}{Aki
  H{\"a}rm{\"a}}, {and} \bibinfo{person}{Giovanni~Maria Farinella}.}
  \bibinfo{year}{2021}\natexlab{}.
\newblock \showarticletitle{A survey on human-aware robot navigation}.
\newblock \bibinfo{journal}{\emph{Robotics and Autonomous Systems}}
  \bibinfo{volume}{145} (\bibinfo{year}{2021}), \bibinfo{pages}{103837}.
\newblock


\bibitem[\protect\citeauthoryear{Moussa{\"\i}d, Helbing, and
  Theraulaz}{Moussa{\"\i}d et~al\mbox{.}}{2011}]%
        {moussaid2011simple}
\bibfield{author}{\bibinfo{person}{Mehdi Moussa{\"\i}d}, \bibinfo{person}{Dirk
  Helbing}, {and} \bibinfo{person}{Guy Theraulaz}.}
  \bibinfo{year}{2011}\natexlab{}.
\newblock \showarticletitle{How simple rules determine pedestrian behavior and
  crowd disasters}.
\newblock \bibinfo{journal}{\emph{Proceedings of the National Academy of
  Sciences}} \bibinfo{volume}{108}, \bibinfo{number}{17}
  (\bibinfo{year}{2011}), \bibinfo{pages}{6884--6888}.
\newblock


\bibitem[\protect\citeauthoryear{Moussa{\"\i}d, Perozo, Garnier, Helbing, and
  Theraulaz}{Moussa{\"\i}d et~al\mbox{.}}{2010}]%
        {moussaid2010walking}
\bibfield{author}{\bibinfo{person}{Mehdi Moussa{\"\i}d},
  \bibinfo{person}{Niriaska Perozo}, \bibinfo{person}{Simon Garnier},
  \bibinfo{person}{Dirk Helbing}, {and} \bibinfo{person}{Guy Theraulaz}.}
  \bibinfo{year}{2010}\natexlab{}.
\newblock \showarticletitle{The walking behaviour of pedestrian social groups
  and its impact on crowd dynamics}.
\newblock \bibinfo{journal}{\emph{PloS one}} \bibinfo{volume}{5},
  \bibinfo{number}{4} (\bibinfo{year}{2010}), \bibinfo{pages}{e10047}.
\newblock


\bibitem[\protect\citeauthoryear{M{\"u}ller, Stachniss, Arras, and
  Burgard}{M{\"u}ller et~al\mbox{.}}{2008}]%
        {muller2008socially}
\bibfield{author}{\bibinfo{person}{J{\"o}rg M{\"u}ller},
  \bibinfo{person}{Cyrill Stachniss}, \bibinfo{person}{Kai~O Arras}, {and}
  \bibinfo{person}{Wolfram Burgard}.} \bibinfo{year}{2008}\natexlab{}.
\newblock \showarticletitle{Socially inspired motion planning for mobile robots
  in populated environments}. In \bibinfo{booktitle}{\emph{Proc. of
  International Conference on Cognitive Systems}}.
\newblock


\bibitem[\protect\citeauthoryear{Murakami, Feliciani, Nishiyama, and
  Nishinari}{Murakami et~al\mbox{.}}{2021}]%
        {murakami2021mutual}
\bibfield{author}{\bibinfo{person}{Hisashi Murakami}, \bibinfo{person}{Claudio
  Feliciani}, \bibinfo{person}{Yuta Nishiyama}, {and}
  \bibinfo{person}{Katsuhiro Nishinari}.} \bibinfo{year}{2021}\natexlab{}.
\newblock \showarticletitle{Mutual anticipation can contribute to
  self-organization in human crowds}.
\newblock \bibinfo{journal}{\emph{Science Advances}} \bibinfo{volume}{7},
  \bibinfo{number}{12} (\bibinfo{year}{2021}), \bibinfo{pages}{eabe7758}.
\newblock


\bibitem[\protect\citeauthoryear{Murakami, Kuno, Shimada, and Shirai}{Murakami
  et~al\mbox{.}}{2002}]%
        {murakami2002collision}
\bibfield{author}{\bibinfo{person}{Yoshifumi Murakami},
  \bibinfo{person}{Yoshinori Kuno}, \bibinfo{person}{Nobutaka Shimada}, {and}
  \bibinfo{person}{Yoshiaki Shirai}.} \bibinfo{year}{2002}\natexlab{}.
\newblock \showarticletitle{Collision avoidance by observing pedestrians' faces
  for intelligent wheelchairs}.
\newblock \bibinfo{journal}{\emph{Journal of the Robotics Society of Japan}}
  \bibinfo{volume}{20}, \bibinfo{number}{2} (\bibinfo{year}{2002}),
  \bibinfo{pages}{206--213}.
\newblock


\bibitem[\protect\citeauthoryear{Musse and Thalmann}{Musse and
  Thalmann}{1997}]%
        {musse1997model}
\bibfield{author}{\bibinfo{person}{Soraia~Raupp Musse} {and}
  \bibinfo{person}{Daniel Thalmann}.} \bibinfo{year}{1997}\natexlab{}.
\newblock \showarticletitle{A model of human crowd behavior: Group
  inter-relationship and collision detection analysis}.
\newblock In \bibinfo{booktitle}{\emph{Computer Animation and
  Simulation’97}}. \bibinfo{publisher}{Springer}, \bibinfo{pages}{39--51}.
\newblock


\bibitem[\protect\citeauthoryear{Nakauchi and Simmons}{Nakauchi and
  Simmons}{2002}]%
        {nakauchi2002social}
\bibfield{author}{\bibinfo{person}{Yasushi Nakauchi} {and}
  \bibinfo{person}{Reid Simmons}.} \bibinfo{year}{2002}\natexlab{}.
\newblock \showarticletitle{A social robot that stands in line}.
\newblock \bibinfo{journal}{\emph{Autonomous Robots}} \bibinfo{volume}{12},
  \bibinfo{number}{3} (\bibinfo{year}{2002}), \bibinfo{pages}{313--324}.
\newblock


\bibitem[\protect\citeauthoryear{Narayanan, Phillips, and Likhachev}{Narayanan
  et~al\mbox{.}}{2012}]%
        {narayanan2012anytime}
\bibfield{author}{\bibinfo{person}{Venkatraman Narayanan},
  \bibinfo{person}{Mike Phillips}, {and} \bibinfo{person}{Maxim Likhachev}.}
  \bibinfo{year}{2012}\natexlab{}.
\newblock \showarticletitle{Anytime safe interval path planning for dynamic
  environments}. In \bibinfo{booktitle}{\emph{2012 IEEE/RSJ International
  Conference on Intelligent Robots and Systems}}. IEEE,
  \bibinfo{pages}{4708--4715}.
\newblock


\bibitem[\protect\citeauthoryear{Nardi and Stachniss}{Nardi and
  Stachniss}{2020}]%
        {nardi2019long}
\bibfield{author}{\bibinfo{person}{Lorenzo Nardi} {and} \bibinfo{person}{Cyrill
  Stachniss}.} \bibinfo{year}{2020}\natexlab{}.
\newblock \showarticletitle{Long-term robot navigation in indoor environments
  estimating patterns in traversability changes}.
\newblock \bibinfo{journal}{\emph{ICRA}} (\bibinfo{year}{2020}).
\newblock


\bibitem[\protect\citeauthoryear{Norman}{Norman}{2009}]%
        {norman2009design}
\bibfield{author}{\bibinfo{person}{Don Norman}.}
  \bibinfo{year}{2009}\natexlab{}.
\newblock \bibinfo{booktitle}{\emph{The design of future things}}.
\newblock \bibinfo{publisher}{Basic books}.
\newblock


\bibitem[\protect\citeauthoryear{Nummenmaa, Hy{\"o}n{\"a}, and
  Hietanen}{Nummenmaa et~al\mbox{.}}{2009}]%
        {nummenmaa2009ll}
\bibfield{author}{\bibinfo{person}{Lauri Nummenmaa}, \bibinfo{person}{Jukka
  Hy{\"o}n{\"a}}, {and} \bibinfo{person}{Jari~K Hietanen}.}
  \bibinfo{year}{2009}\natexlab{}.
\newblock \showarticletitle{I'll walk this way: Eyes reveal the direction of
  locomotion and make passersby look and go the other way}.
\newblock \bibinfo{journal}{\emph{Psychological Science}} \bibinfo{volume}{20},
  \bibinfo{number}{12} (\bibinfo{year}{2009}), \bibinfo{pages}{1454--1458}.
\newblock


\bibitem[\protect\citeauthoryear{O'Callaghan, Singh, Alempijevic, and
  Ramos}{O'Callaghan et~al\mbox{.}}{2011}]%
        {o2011learning}
\bibfield{author}{\bibinfo{person}{Simon~T O'Callaghan},
  \bibinfo{person}{Surya~PN Singh}, \bibinfo{person}{Alen Alempijevic}, {and}
  \bibinfo{person}{Fabio~T Ramos}.} \bibinfo{year}{2011}\natexlab{}.
\newblock \showarticletitle{Learning navigational maps by observing human
  motion patterns}. In \bibinfo{booktitle}{\emph{2011 IEEE International
  Conference on Robotics and Automation}}. IEEE, \bibinfo{pages}{4333--4340}.
\newblock


\bibitem[\protect\citeauthoryear{Ohki, Nagatani, and Yoshida}{Ohki
  et~al\mbox{.}}{2010}]%
        {ohki2010collision}
\bibfield{author}{\bibinfo{person}{Takeshi Ohki}, \bibinfo{person}{Keiji
  Nagatani}, {and} \bibinfo{person}{Kazuya Yoshida}.}
  \bibinfo{year}{2010}\natexlab{}.
\newblock \showarticletitle{Collision avoidance method for mobile robot
  considering motion and personal spaces of evacuees}. In
  \bibinfo{booktitle}{\emph{2010 IEEE/RSJ International Conference on
  Intelligent Robots and Systems}}. IEEE, \bibinfo{pages}{1819--1824}.
\newblock


\bibitem[\protect\citeauthoryear{Okal and Arras}{Okal and Arras}{2014}]%
        {okal2014towards}
\bibfield{author}{\bibinfo{person}{Billy Okal} {and} \bibinfo{person}{Kai~O
  Arras}.} \bibinfo{year}{2014}\natexlab{}.
\newblock \showarticletitle{Towards group-level social activity recognition for
  mobile robots}. In \bibinfo{booktitle}{\emph{IROS Assistance and Service
  Robotics in a Human Environments Workshop}}.
\newblock


\bibitem[\protect\citeauthoryear{Okal and Arras}{Okal and Arras}{2016}]%
        {okal2016learning}
\bibfield{author}{\bibinfo{person}{Billy Okal} {and} \bibinfo{person}{Kai~O
  Arras}.} \bibinfo{year}{2016}\natexlab{}.
\newblock \showarticletitle{Learning socially normative robot navigation
  behaviors with bayesian inverse reinforcement learning}. In
  \bibinfo{booktitle}{\emph{2016 IEEE International Conference on Robotics and
  Automation (ICRA)}}. IEEE, \bibinfo{pages}{2889--2895}.
\newblock


\bibitem[\protect\citeauthoryear{Pacchierotti, Christensen, and
  Jensfelt}{Pacchierotti et~al\mbox{.}}{2006}]%
        {pacchierotti2006design}
\bibfield{author}{\bibinfo{person}{Elena Pacchierotti},
  \bibinfo{person}{Henrik~I Christensen}, {and} \bibinfo{person}{Patric
  Jensfelt}.} \bibinfo{year}{2006}\natexlab{}.
\newblock \showarticletitle{Design of an office-guide robot for social
  interaction studies}. In \bibinfo{booktitle}{\emph{2006 IEEE/RSJ
  International Conference on Intelligent Robots and Systems}}. IEEE,
  \bibinfo{pages}{4965--4970}.
\newblock


\bibitem[\protect\citeauthoryear{Pandey and Alami}{Pandey and Alami}{2010}]%
        {pandey2010framework}
\bibfield{author}{\bibinfo{person}{Amit~Kumar Pandey} {and}
  \bibinfo{person}{Rachid Alami}.} \bibinfo{year}{2010}\natexlab{}.
\newblock \showarticletitle{A framework towards a socially aware mobile robot
  motion in human-centered dynamic environment}. In
  \bibinfo{booktitle}{\emph{2010 IEEE/RSJ International Conference on
  Intelligent Robots and Systems}}. IEEE, \bibinfo{pages}{5855--5860}.
\newblock


\bibitem[\protect\citeauthoryear{Papadakis, Rives, and Spalanzani}{Papadakis
  et~al\mbox{.}}{2014}]%
        {papadakis2014adaptive}
\bibfield{author}{\bibinfo{person}{Panagiotis Papadakis},
  \bibinfo{person}{Patrick Rives}, {and} \bibinfo{person}{Anne Spalanzani}.}
  \bibinfo{year}{2014}\natexlab{}.
\newblock \showarticletitle{Adaptive spacing in human-robot interactions}. In
  \bibinfo{booktitle}{\emph{2014 IEEE/RSJ International Conference on
  Intelligent Robots and Systems}}. IEEE, \bibinfo{pages}{2627--2632}.
\newblock


\bibitem[\protect\citeauthoryear{Park, Rojas, and Yang}{Park
  et~al\mbox{.}}{2013}]%
        {park2013collision}
\bibfield{author}{\bibinfo{person}{Jin~Hyoung Park},
  \bibinfo{person}{Francisco~Arturo Rojas}, {and} \bibinfo{person}{Hyun~Seung
  Yang}.} \bibinfo{year}{2013}\natexlab{}.
\newblock \showarticletitle{A collision avoidance behavior model for crowd
  simulation based on psychological findings}.
\newblock \bibinfo{journal}{\emph{Computer Animation and Virtual Worlds}}
  \bibinfo{volume}{24}, \bibinfo{number}{3-4} (\bibinfo{year}{2013}),
  \bibinfo{pages}{173--183}.
\newblock


\bibitem[\protect\citeauthoryear{Patla, Adkin, and Ballard}{Patla
  et~al\mbox{.}}{1999}]%
        {Patla1999}
\bibfield{author}{\bibinfo{person}{A.~E. Patla}, \bibinfo{person}{A. Adkin},
  {and} \bibinfo{person}{T. Ballard}.} \bibinfo{year}{1999}\natexlab{}.
\newblock \showarticletitle{Online steering: coordination and control of body
  center of mass, head and body reorientation}.
\newblock \bibinfo{journal}{\emph{Experimental Brain Research}}
  \bibinfo{volume}{129}, \bibinfo{number}{4} (\bibinfo{date}{01 Dec}
  \bibinfo{year}{1999}), \bibinfo{pages}{629--634}.
\newblock
\showISSN{1432-1106}
\urldef\tempurl%
\url{https://doi.org/10.1007/s002210050932}
\showDOI{\tempurl}


\bibitem[\protect\citeauthoryear{Pfeiffer, Schwesinger, Sommer, Galceran, and
  Siegwart}{Pfeiffer et~al\mbox{.}}{2016}]%
        {pfeiffer2016predicting}
\bibfield{author}{\bibinfo{person}{Mark Pfeiffer}, \bibinfo{person}{Ulrich
  Schwesinger}, \bibinfo{person}{Hannes Sommer}, \bibinfo{person}{Enric
  Galceran}, {and} \bibinfo{person}{Roland Siegwart}.}
  \bibinfo{year}{2016}\natexlab{}.
\newblock \showarticletitle{Predicting actions to act predictably: Cooperative
  partial motion planning with maximum entropy models}. In
  \bibinfo{booktitle}{\emph{2016 IEEE/RSJ International Conference on
  Intelligent Robots and Systems (IROS)}}. IEEE, \bibinfo{pages}{2096--2101}.
\newblock


\bibitem[\protect\citeauthoryear{Pirk, Lee, Xiao, Takayama, Francis, and
  Toshev}{Pirk et~al\mbox{.}}{2022}]%
        {pirk2022protocol}
\bibfield{author}{\bibinfo{person}{S{\"o}ren Pirk}, \bibinfo{person}{Edward
  Lee}, \bibinfo{person}{Xuesu Xiao}, \bibinfo{person}{Leila Takayama},
  \bibinfo{person}{Anthony Francis}, {and} \bibinfo{person}{Alexander Toshev}.}
  \bibinfo{year}{2022}\natexlab{}.
\newblock \showarticletitle{A Protocol for Validating Social Navigation
  Policies}.
\newblock \bibinfo{journal}{\emph{arXiv preprint arXiv:2204.05443}}
  (\bibinfo{year}{2022}).
\newblock


\bibitem[\protect\citeauthoryear{Povinelli, Bierschwale, and Cech}{Povinelli
  et~al\mbox{.}}{1999}]%
        {povinelli1999comprehension}
\bibfield{author}{\bibinfo{person}{Daniel~J Povinelli},
  \bibinfo{person}{Donna~T Bierschwale}, {and} \bibinfo{person}{Claude~G
  Cech}.} \bibinfo{year}{1999}\natexlab{}.
\newblock \showarticletitle{Comprehension of seeing as a referential act in
  young children, but not juvenile chimpanzees}.
\newblock \bibinfo{journal}{\emph{British Journal of Developmental Psychology}}
  \bibinfo{volume}{17}, \bibinfo{number}{1} (\bibinfo{year}{1999}),
  \bibinfo{pages}{37--60}.
\newblock


\bibitem[\protect\citeauthoryear{Prassler, Bank, Kluge, and Hagele}{Prassler
  et~al\mbox{.}}{2002}]%
        {prassler2002key}
\bibfield{author}{\bibinfo{person}{Erwin Prassler}, \bibinfo{person}{Dirk
  Bank}, \bibinfo{person}{Boris Kluge}, {and} \bibinfo{person}{M Hagele}.}
  \bibinfo{year}{2002}\natexlab{}.
\newblock \showarticletitle{Key technologies in robot assistants: Motion
  coordination between a human and a mobile robot}.
\newblock \bibinfo{journal}{\emph{Transactions on Control, Automation and
  Systems Engineering}} \bibinfo{volume}{4}, \bibinfo{number}{1}
  (\bibinfo{year}{2002}), \bibinfo{pages}{56--61}.
\newblock


\bibitem[\protect\citeauthoryear{Ratsamee, Mae, Ohara, Takubo, and
  Arai}{Ratsamee et~al\mbox{.}}{2013}]%
        {ratsamee2013human}
\bibfield{author}{\bibinfo{person}{Photchara Ratsamee},
  \bibinfo{person}{Yasushi Mae}, \bibinfo{person}{Kenichi Ohara},
  \bibinfo{person}{Tomohito Takubo}, {and} \bibinfo{person}{Tatsuo Arai}.}
  \bibinfo{year}{2013}\natexlab{}.
\newblock \showarticletitle{Human--robot collision avoidance using a modified
  social force model with body pose and face orientation}.
\newblock \bibinfo{journal}{\emph{International Journal of Humanoid Robotics}}
  \bibinfo{volume}{10}, \bibinfo{number}{01} (\bibinfo{year}{2013}),
  \bibinfo{pages}{1350008}.
\newblock


\bibitem[\protect\citeauthoryear{Reig, Luria, Wang, Oltman, Carter, Steinfeld,
  Forlizzi, and Zimmerman}{Reig et~al\mbox{.}}{2020}]%
        {reig2020not}
\bibfield{author}{\bibinfo{person}{Samantha Reig}, \bibinfo{person}{Michal
  Luria}, \bibinfo{person}{Janet~Z Wang}, \bibinfo{person}{Danielle Oltman},
  \bibinfo{person}{Elizabeth~Jeanne Carter}, \bibinfo{person}{Aaron Steinfeld},
  \bibinfo{person}{Jodi Forlizzi}, {and} \bibinfo{person}{John Zimmerman}.}
  \bibinfo{year}{2020}\natexlab{}.
\newblock \showarticletitle{Not Some Random Agent: Multi-person interaction
  with a personalizing service robot}. In \bibinfo{booktitle}{\emph{Proceedings
  of the 2020 ACM/IEEE International Conference on Human-Robot Interaction}}.
  \bibinfo{pages}{289--297}.
\newblock


\bibitem[\protect\citeauthoryear{Reynolds}{Reynolds}{1999}]%
        {reynolds1999steering}
\bibfield{author}{\bibinfo{person}{Craig~W Reynolds}.}
  \bibinfo{year}{1999}\natexlab{}.
\newblock \showarticletitle{Steering behaviors for autonomous characters}. In
  \bibinfo{booktitle}{\emph{Game developers conference}},
  Vol.~\bibinfo{volume}{1999}. Citeseer, \bibinfo{pages}{763--782}.
\newblock


\bibitem[\protect\citeauthoryear{Rios-Martinez, Renzaglia, Spalanzani,
  Martinelli, and Laugier}{Rios-Martinez et~al\mbox{.}}{2012}]%
        {rios2012navigating}
\bibfield{author}{\bibinfo{person}{Jorge Rios-Martinez},
  \bibinfo{person}{Alessandro Renzaglia}, \bibinfo{person}{Anne Spalanzani},
  \bibinfo{person}{Agostino Martinelli}, {and} \bibinfo{person}{Christian
  Laugier}.} \bibinfo{year}{2012}\natexlab{}.
\newblock \showarticletitle{Navigating between people: A stochastic
  optimization approach}. In \bibinfo{booktitle}{\emph{2012 IEEE International
  Conference on Robotics and Automation}}. IEEE, \bibinfo{pages}{2880--2885}.
\newblock


\bibitem[\protect\citeauthoryear{Rios-Martinez, Spalanzani, and
  Laugier}{Rios-Martinez et~al\mbox{.}}{2015}]%
        {rios2015proxemics}
\bibfield{author}{\bibinfo{person}{Jorge Rios-Martinez}, \bibinfo{person}{Anne
  Spalanzani}, {and} \bibinfo{person}{Christian Laugier}.}
  \bibinfo{year}{2015}\natexlab{}.
\newblock \showarticletitle{From proxemics theory to socially-aware navigation:
  A survey}.
\newblock \bibinfo{journal}{\emph{International Journal of Social Robotics}}
  \bibinfo{volume}{7}, \bibinfo{number}{2} (\bibinfo{year}{2015}),
  \bibinfo{pages}{137--153}.
\newblock


\bibitem[\protect\citeauthoryear{Saran, Majumdar, Short, Thomaz, and
  Niekum}{Saran et~al\mbox{.}}{2018}]%
        {saran2018human}
\bibfield{author}{\bibinfo{person}{Akanksha Saran}, \bibinfo{person}{Srinjoy
  Majumdar}, \bibinfo{person}{Elaine~Schaertl Short}, \bibinfo{person}{Andrea
  Thomaz}, {and} \bibinfo{person}{Scott Niekum}.}
  \bibinfo{year}{2018}\natexlab{}.
\newblock \showarticletitle{Human gaze following for human-robot interaction}.
  In \bibinfo{booktitle}{\emph{2018 IEEE/RSJ International Conference on
  Intelligent Robots and Systems (IROS)}}. IEEE, \bibinfo{pages}{8615--8621}.
\newblock


\bibitem[\protect\citeauthoryear{Saran, Short, Thomaz, and Niekum}{Saran
  et~al\mbox{.}}{2020}]%
        {saran2020understanding}
\bibfield{author}{\bibinfo{person}{Akanksha Saran},
  \bibinfo{person}{Elaine~Schaertl Short}, \bibinfo{person}{Andrea Thomaz},
  {and} \bibinfo{person}{Scott Niekum}.} \bibinfo{year}{2020}\natexlab{}.
\newblock \showarticletitle{Understanding teacher gaze patterns for robot
  learning}. In \bibinfo{booktitle}{\emph{Conference on Robot Learning}}. PMLR,
  \bibinfo{pages}{1247--1258}.
\newblock


\bibitem[\protect\citeauthoryear{Sciutti, Bisio, Nori, Metta, Fadiga, Pozzo,
  and Sandini}{Sciutti et~al\mbox{.}}{2012}]%
        {sciutti2012measuring}
\bibfield{author}{\bibinfo{person}{Alessandra Sciutti}, \bibinfo{person}{Ambra
  Bisio}, \bibinfo{person}{Francesco Nori}, \bibinfo{person}{Giorgio Metta},
  \bibinfo{person}{Luciano Fadiga}, \bibinfo{person}{Thierry Pozzo}, {and}
  \bibinfo{person}{Giulio Sandini}.} \bibinfo{year}{2012}\natexlab{}.
\newblock \showarticletitle{Measuring human-robot interaction through motor
  resonance}.
\newblock \bibinfo{journal}{\emph{International Journal of Social Robotics}}
  \bibinfo{volume}{4}, \bibinfo{number}{3} (\bibinfo{year}{2012}),
  \bibinfo{pages}{223--234}.
\newblock


\bibitem[\protect\citeauthoryear{Senft, Satake, and Kanda}{Senft
  et~al\mbox{.}}{2020}]%
        {senft2020would}
\bibfield{author}{\bibinfo{person}{Emmanuel Senft}, \bibinfo{person}{Satoru
  Satake}, {and} \bibinfo{person}{Takayuki Kanda}.}
  \bibinfo{year}{2020}\natexlab{}.
\newblock \showarticletitle{Would You Mind Me if I Pass by You?
  Socially-Appropriate Behaviour for an Omni-based Social Robot in Narrow
  Environment}. In \bibinfo{booktitle}{\emph{Proceedings of the 2020 ACM/IEEE
  International Conference on Human-Robot Interaction}}.
  \bibinfo{pages}{539--547}.
\newblock


\bibitem[\protect\citeauthoryear{Shi, Satake, Kanda, and Ishiguro}{Shi
  et~al\mbox{.}}{2018}]%
        {shi2018robot}
\bibfield{author}{\bibinfo{person}{Chao Shi}, \bibinfo{person}{Satoru Satake},
  \bibinfo{person}{Takayuki Kanda}, {and} \bibinfo{person}{Hiroshi Ishiguro}.}
  \bibinfo{year}{2018}\natexlab{}.
\newblock \showarticletitle{A robot that distributes flyers to pedestrians in a
  shopping mall}.
\newblock \bibinfo{journal}{\emph{International Journal of Social Robotics}}
  \bibinfo{volume}{10}, \bibinfo{number}{4} (\bibinfo{year}{2018}),
  \bibinfo{pages}{421--437}.
\newblock


\bibitem[\protect\citeauthoryear{Shit}{Shit}{2020}]%
        {shit2020precise}
\bibfield{author}{\bibinfo{person}{Rathin~Chandra Shit}.}
  \bibinfo{year}{2020}\natexlab{}.
\newblock \showarticletitle{Precise localization for achieving next-generation
  autonomous navigation: State-of-the-art, taxonomy and future prospects}.
\newblock \bibinfo{journal}{\emph{Computer Communications}}
  \bibinfo{volume}{160} (\bibinfo{year}{2020}), \bibinfo{pages}{351--374}.
\newblock


\bibitem[\protect\citeauthoryear{Shrestha, Onishi, Kobayashi, Kamezaki, and
  Sugano}{Shrestha et~al\mbox{.}}{2018}]%
        {shrestha2018communicating}
\bibfield{author}{\bibinfo{person}{Moondeep~C Shrestha},
  \bibinfo{person}{Tomoya Onishi}, \bibinfo{person}{Ayano Kobayashi},
  \bibinfo{person}{Mitsuhiro Kamezaki}, {and} \bibinfo{person}{Shigeki
  Sugano}.} \bibinfo{year}{2018}\natexlab{}.
\newblock \showarticletitle{Communicating directional intent in robot
  navigation using projection indicators}. In \bibinfo{booktitle}{\emph{2018
  27th IEEE International Symposium on Robot and Human Interactive
  Communication (RO-MAN)}}. IEEE, \bibinfo{pages}{746--751}.
\newblock


\bibitem[\protect\citeauthoryear{{Shrestha}, {Onishi}, {Kobayashi}, {Kamezaki},
  and {Sugano}}{{Shrestha} et~al\mbox{.}}{2018}]%
        {8525528}
\bibfield{author}{\bibinfo{person}{M.~C. {Shrestha}}, \bibinfo{person}{T.
  {Onishi}}, \bibinfo{person}{A. {Kobayashi}}, \bibinfo{person}{M. {Kamezaki}},
  {and} \bibinfo{person}{S. {Sugano}}.} \bibinfo{year}{2018}\natexlab{}.
\newblock \showarticletitle{Communicating Directional Intent in Robot
  Navigation using Projection Indicators}. In \bibinfo{booktitle}{\emph{2018
  27th IEEE International Symposium on Robot and Human Interactive
  Communication (RO-MAN)}}. \bibinfo{pages}{746--751}.
\newblock
\showISSN{1944-9437}
\urldef\tempurl%
\url{https://doi.org/10.1109/ROMAN.2018.8525528}
\showDOI{\tempurl}


\bibitem[\protect\citeauthoryear{Siegwart, Nourbakhsh, and Scaramuzza}{Siegwart
  et~al\mbox{.}}{2011}]%
        {siegwart2011introduction}
\bibfield{author}{\bibinfo{person}{Roland Siegwart},
  \bibinfo{person}{Illah~Reza Nourbakhsh}, {and} \bibinfo{person}{Davide
  Scaramuzza}.} \bibinfo{year}{2011}\natexlab{}.
\newblock \bibinfo{booktitle}{\emph{Introduction to autonomous mobile robots}}.
\newblock \bibinfo{publisher}{MIT press}.
\newblock


\bibitem[\protect\citeauthoryear{Silva and Fraichard}{Silva and
  Fraichard}{2017}]%
        {silva2017human}
\bibfield{author}{\bibinfo{person}{Grimaldo Silva} {and}
  \bibinfo{person}{Thierry Fraichard}.} \bibinfo{year}{2017}\natexlab{}.
\newblock \showarticletitle{Human robot motion: A shared effort approach}. In
  \bibinfo{booktitle}{\emph{2017 European Conference on Mobile Robots (ECMR)}}.
  IEEE, \bibinfo{pages}{1--6}.
\newblock


\bibitem[\protect\citeauthoryear{Singh, Miller, Newn, Velloso, Vetere, and
  Sonenberg}{Singh et~al\mbox{.}}{2020}]%
        {singh2020combining}
\bibfield{author}{\bibinfo{person}{Ronal Singh}, \bibinfo{person}{Tim Miller},
  \bibinfo{person}{Joshua Newn}, \bibinfo{person}{Eduardo Velloso},
  \bibinfo{person}{Frank Vetere}, {and} \bibinfo{person}{Liz Sonenberg}.}
  \bibinfo{year}{2020}\natexlab{}.
\newblock \showarticletitle{Combining gaze and AI planning for online human
  intention recognition}.
\newblock \bibinfo{journal}{\emph{Artificial Intelligence}}
  (\bibinfo{year}{2020}), \bibinfo{pages}{103275}.
\newblock


\bibitem[\protect\citeauthoryear{Sisbot, Marin-Urias, Alami, and Simeon}{Sisbot
  et~al\mbox{.}}{2007}]%
        {sisbot2007human}
\bibfield{author}{\bibinfo{person}{Emrah~Akin Sisbot}, \bibinfo{person}{Luis~F
  Marin-Urias}, \bibinfo{person}{Rachid Alami}, {and} \bibinfo{person}{Thierry
  Simeon}.} \bibinfo{year}{2007}\natexlab{}.
\newblock \showarticletitle{A human aware mobile robot motion planner}.
\newblock \bibinfo{journal}{\emph{IEEE Transactions on Robotics}}
  \bibinfo{volume}{23}, \bibinfo{number}{5} (\bibinfo{year}{2007}),
  \bibinfo{pages}{874--883}.
\newblock


\bibitem[\protect\citeauthoryear{{Smith}, {Ba}, {Odobez}, and
  {Gatica-Perez}}{{Smith} et~al\mbox{.}}{2008}]%
        {4359373}
\bibfield{author}{\bibinfo{person}{K. {Smith}}, \bibinfo{person}{S.~O. {Ba}},
  \bibinfo{person}{J. {Odobez}}, {and} \bibinfo{person}{D. {Gatica-Perez}}.}
  \bibinfo{year}{2008}\natexlab{}.
\newblock \showarticletitle{Tracking the Visual Focus of Attention for a
  Varying Number of Wandering People}.
\newblock \bibinfo{journal}{\emph{IEEE Transactions on Pattern Analysis and
  Machine Intelligence}} \bibinfo{volume}{30}, \bibinfo{number}{7}
  (\bibinfo{date}{July} \bibinfo{year}{2008}), \bibinfo{pages}{1212--1229}.
\newblock
\showISSN{0162-8828}
\urldef\tempurl%
\url{https://doi.org/10.1109/TPAMI.2007.70773}
\showDOI{\tempurl}


\bibitem[\protect\citeauthoryear{Soproni, Mikl{\'o}si, Top{\'a}l, and
  Cs{\'a}nyi}{Soproni et~al\mbox{.}}{2001}]%
        {soproni2001comprehension}
\bibfield{author}{\bibinfo{person}{Krisztina Soproni},
  \bibinfo{person}{{\'A}d{\'a}m Mikl{\'o}si}, \bibinfo{person}{J{\'o}zsef
  Top{\'a}l}, {and} \bibinfo{person}{Vilmos Cs{\'a}nyi}.}
  \bibinfo{year}{2001}\natexlab{}.
\newblock \showarticletitle{Comprehension of human communicative signs in pet
  dogs (Canis familiaris).}
\newblock \bibinfo{journal}{\emph{Journal of comparative psychology}}
  \bibinfo{volume}{115}, \bibinfo{number}{2} (\bibinfo{year}{2001}),
  \bibinfo{pages}{122}.
\newblock


\bibitem[\protect\citeauthoryear{Stiefelhagen, Finke, Yang, and
  Waibel}{Stiefelhagen et~al\mbox{.}}{1999}]%
        {stiefelhagen1999gaze}
\bibfield{author}{\bibinfo{person}{Rainer Stiefelhagen},
  \bibinfo{person}{Michael Finke}, \bibinfo{person}{Jie Yang}, {and}
  \bibinfo{person}{Alex Waibel}.} \bibinfo{year}{1999}\natexlab{}.
\newblock \showarticletitle{From gaze to focus of attention}. In
  \bibinfo{booktitle}{\emph{International Conference on Advances in Visual
  Information Systems}}. Springer, \bibinfo{pages}{765--772}.
\newblock


\bibitem[\protect\citeauthoryear{Strassner and Langer}{Strassner and
  Langer}{2005}]%
        {strassner2005virtual}
\bibfield{author}{\bibinfo{person}{Johannes Strassner} {and}
  \bibinfo{person}{Marion Langer}.} \bibinfo{year}{2005}\natexlab{}.
\newblock \showarticletitle{Virtual humans with personalized perception and
  dynamic levels of knowledge}.
\newblock \bibinfo{journal}{\emph{Computer Animation and Virtual Worlds}}
  \bibinfo{volume}{16}, \bibinfo{number}{3-4} (\bibinfo{year}{2005}),
  \bibinfo{pages}{331--342}.
\newblock


\bibitem[\protect\citeauthoryear{Svenstrup, Bak, and Andersen}{Svenstrup
  et~al\mbox{.}}{2010}]%
        {svenstrup2010trajectory}
\bibfield{author}{\bibinfo{person}{Mikael Svenstrup}, \bibinfo{person}{Thomas
  Bak}, {and} \bibinfo{person}{Hans~J{\o}rgen Andersen}.}
  \bibinfo{year}{2010}\natexlab{}.
\newblock \showarticletitle{Trajectory planning for robots in dynamic human
  environments}. In \bibinfo{booktitle}{\emph{2010 IEEE/RSJ International
  Conference on Intelligent Robots and Systems}}. IEEE,
  \bibinfo{pages}{4293--4298}.
\newblock


\bibitem[\protect\citeauthoryear{Swofford, Peruzzi, Tsoi, Thompson,
  Mart{\'\i}n-Mart{\'\i}n, Savarese, and V{\'a}zquez}{Swofford
  et~al\mbox{.}}{2020}]%
        {swofford2020improving}
\bibfield{author}{\bibinfo{person}{Mason Swofford}, \bibinfo{person}{John
  Peruzzi}, \bibinfo{person}{Nathan Tsoi}, \bibinfo{person}{Sydney Thompson},
  \bibinfo{person}{Roberto Mart{\'\i}n-Mart{\'\i}n}, \bibinfo{person}{Silvio
  Savarese}, {and} \bibinfo{person}{Marynel V{\'a}zquez}.}
  \bibinfo{year}{2020}\natexlab{}.
\newblock \showarticletitle{Improving Social Awareness Through DANTE: Deep
  Affinity Network for Clustering Conversational Interactants}.
\newblock \bibinfo{journal}{\emph{Proceedings of the ACM on Human-Computer
  Interaction}} \bibinfo{volume}{4}, \bibinfo{number}{CSCW1}
  (\bibinfo{year}{2020}), \bibinfo{pages}{1--23}.
\newblock


\bibitem[\protect\citeauthoryear{Syrdal, Otero, and Dautenhahn}{Syrdal
  et~al\mbox{.}}{2008}]%
        {syrdal2008video}
\bibfield{author}{\bibinfo{person}{Dag~Sverre Syrdal}, \bibinfo{person}{Nuno
  Otero}, {and} \bibinfo{person}{Kerstin Dautenhahn}.}
  \bibinfo{year}{2008}\natexlab{}.
\newblock \showarticletitle{Video prototyping in human-robot interaction:
  Results from a qualitative study}. In \bibinfo{booktitle}{\emph{Proceedings
  of the 15th European conference on Cognitive ergonomics: the ergonomics of
  cool interaction}}. \bibinfo{pages}{1--8}.
\newblock


\bibitem[\protect\citeauthoryear{Szafir, Mutlu, and Fong}{Szafir
  et~al\mbox{.}}{2015}]%
        {Szafir:2015:CDF:2696454.2696475}
\bibfield{author}{\bibinfo{person}{Daniel Szafir}, \bibinfo{person}{Bilge
  Mutlu}, {and} \bibinfo{person}{Terry Fong}.} \bibinfo{year}{2015}\natexlab{}.
\newblock \showarticletitle{Communicating Directionality in Flying Robots}. In
  \bibinfo{booktitle}{\emph{Proceedings of the Tenth Annual ACM/IEEE
  International Conference on Human-Robot Interaction}}
  \emph{(\bibinfo{series}{HRI '15})}. \bibinfo{publisher}{ACM},
  \bibinfo{address}{New York, NY, USA}, \bibinfo{pages}{19--26}.
\newblock
\showISBNx{978-1-4503-2883-8}
\urldef\tempurl%
\url{https://doi.org/10.1145/2696454.2696475}
\showDOI{\tempurl}


\bibitem[\protect\citeauthoryear{Tai, Zhang, Liu, and Burgard}{Tai
  et~al\mbox{.}}{2018}]%
        {tai2018socially}
\bibfield{author}{\bibinfo{person}{Lei Tai}, \bibinfo{person}{Jingwei Zhang},
  \bibinfo{person}{Ming Liu}, {and} \bibinfo{person}{Wolfram Burgard}.}
  \bibinfo{year}{2018}\natexlab{}.
\newblock \showarticletitle{Socially compliant navigation through raw depth
  inputs with generative adversarial imitation learning}. In
  \bibinfo{booktitle}{\emph{2018 IEEE International Conference on Robotics and
  Automation (ICRA)}}. IEEE, \bibinfo{pages}{1111--1117}.
\newblock


\bibitem[\protect\citeauthoryear{Tamura, Fukuzawa, and Asama}{Tamura
  et~al\mbox{.}}{2010}]%
        {tamura2010smooth}
\bibfield{author}{\bibinfo{person}{Yusuke Tamura}, \bibinfo{person}{Tomohiro
  Fukuzawa}, {and} \bibinfo{person}{Hajime Asama}.}
  \bibinfo{year}{2010}\natexlab{}.
\newblock \showarticletitle{Smooth collision avoidance in human-robot
  coexisting environment}. In \bibinfo{booktitle}{\emph{2010 IEEE/RSJ
  International Conference on Intelligent Robots and Systems}}. IEEE,
  \bibinfo{pages}{3887--3892}.
\newblock


\bibitem[\protect\citeauthoryear{Thrun, Beetz, Bennewitz, Burgard, Cremers,
  Dellaert, Fox, Haehnel, Rosenberg, Roy, et~al\mbox{.}}{Thrun
  et~al\mbox{.}}{2000}]%
        {thrun2000probabilistic}
\bibfield{author}{\bibinfo{person}{Sebastian Thrun}, \bibinfo{person}{Michael
  Beetz}, \bibinfo{person}{Maren Bennewitz}, \bibinfo{person}{Wolfram Burgard},
  \bibinfo{person}{Armin~B Cremers}, \bibinfo{person}{Frank Dellaert},
  \bibinfo{person}{Dieter Fox}, \bibinfo{person}{Dirk Haehnel},
  \bibinfo{person}{Chuck Rosenberg}, \bibinfo{person}{Nicholas Roy},
  {et~al\mbox{.}}} \bibinfo{year}{2000}\natexlab{}.
\newblock \showarticletitle{Probabilistic algorithms and the interactive museum
  tour-guide robot minerva}.
\newblock \bibinfo{journal}{\emph{The International Journal of Robotics
  Research}} \bibinfo{volume}{19}, \bibinfo{number}{11} (\bibinfo{year}{2000}),
  \bibinfo{pages}{972--999}.
\newblock


\bibitem[\protect\citeauthoryear{Topp and Christensen}{Topp and
  Christensen}{2005}]%
        {topp2005tracking}
\bibfield{author}{\bibinfo{person}{Elin~Anna Topp} {and}
  \bibinfo{person}{Henrik~I Christensen}.} \bibinfo{year}{2005}\natexlab{}.
\newblock \showarticletitle{Tracking for following and passing persons}. In
  \bibinfo{booktitle}{\emph{2005 IEEE/RSJ International Conference on
  Intelligent Robots and Systems}}. IEEE, \bibinfo{pages}{2321--2327}.
\newblock


\bibitem[\protect\citeauthoryear{Torta, Cuijpers, and Juola}{Torta
  et~al\mbox{.}}{2013}]%
        {torta2013design}
\bibfield{author}{\bibinfo{person}{Elena Torta}, \bibinfo{person}{Raymond~H
  Cuijpers}, {and} \bibinfo{person}{James~F Juola}.}
  \bibinfo{year}{2013}\natexlab{}.
\newblock \showarticletitle{Design of a parametric model of personal space for
  robotic social navigation}.
\newblock \bibinfo{journal}{\emph{International Journal of Social Robotics}}
  \bibinfo{volume}{5}, \bibinfo{number}{3} (\bibinfo{year}{2013}),
  \bibinfo{pages}{357--365}.
\newblock


\bibitem[\protect\citeauthoryear{Treuille, Cooper, and Popovi{\'c}}{Treuille
  et~al\mbox{.}}{2006}]%
        {treuille2006continuum}
\bibfield{author}{\bibinfo{person}{Adrien Treuille}, \bibinfo{person}{Seth
  Cooper}, {and} \bibinfo{person}{Zoran Popovi{\'c}}.}
  \bibinfo{year}{2006}\natexlab{}.
\newblock \showarticletitle{Continuum crowds}.
\newblock \bibinfo{journal}{\emph{ACM Transactions on Graphics (TOG)}}
  \bibinfo{volume}{25}, \bibinfo{number}{3} (\bibinfo{year}{2006}),
  \bibinfo{pages}{1160--1168}.
\newblock
\newblock
\shownote{\url{https://howtorts.github.io/2014/01/09/continuum-crowds.html}.}


\bibitem[\protect\citeauthoryear{Truc, Singamaneni, Sidobre, Ivaldi, and
  Alami}{Truc et~al\mbox{.}}{2022}]%
        {truc2022khaos}
\bibfield{author}{\bibinfo{person}{J{\'e}r{\^o}me Truc},
  \bibinfo{person}{Phani-Teja Singamaneni}, \bibinfo{person}{Daniel Sidobre},
  \bibinfo{person}{Serena Ivaldi}, {and} \bibinfo{person}{Rachid Alami}.}
  \bibinfo{year}{2022}\natexlab{}.
\newblock \showarticletitle{KHAOS: a Kinematic Human Aware Optimization-based
  System for Reactive Planning of Flying-Coworker}. In
  \bibinfo{booktitle}{\emph{ICRA 2022}}.
\newblock


\bibitem[\protect\citeauthoryear{Truong and Ngo}{Truong and Ngo}{2016}]%
        {truong2016dynamic}
\bibfield{author}{\bibinfo{person}{Xuan-Tung Truong} {and}
  \bibinfo{person}{Trung-Dung Ngo}.} \bibinfo{year}{2016}\natexlab{}.
\newblock \showarticletitle{Dynamic social zone based mobile robot navigation
  for human comfortable safety in social environments}.
\newblock \bibinfo{journal}{\emph{International Journal of Social Robotics}}
  \bibinfo{volume}{8}, \bibinfo{number}{5} (\bibinfo{year}{2016}),
  \bibinfo{pages}{663--684}.
\newblock


\bibitem[\protect\citeauthoryear{Tsoi, Hussein, Espinoza, Ruiz, and
  V{\'a}zquez}{Tsoi et~al\mbox{.}}{2020}]%
        {tsoi2020sean}
\bibfield{author}{\bibinfo{person}{Nathan Tsoi}, \bibinfo{person}{Mohamed
  Hussein}, \bibinfo{person}{Jeacy Espinoza}, \bibinfo{person}{Xavier Ruiz},
  {and} \bibinfo{person}{Marynel V{\'a}zquez}.}
  \bibinfo{year}{2020}\natexlab{}.
\newblock \showarticletitle{SEAN: Social Environment for Autonomous
  Navigation}. In \bibinfo{booktitle}{\emph{Proceedings of the 8th
  International Conference on Human-Agent Interaction}}.
  \bibinfo{pages}{281--283}.
\newblock


\bibitem[\protect\citeauthoryear{Unhelkar, P{\'e}rez-D'Arpino, Stirling, and
  Shah}{Unhelkar et~al\mbox{.}}{2015}]%
        {unhelkar2015human}
\bibfield{author}{\bibinfo{person}{Vaibhav~V Unhelkar},
  \bibinfo{person}{Claudia P{\'e}rez-D'Arpino}, \bibinfo{person}{Leia
  Stirling}, {and} \bibinfo{person}{Julie~A Shah}.}
  \bibinfo{year}{2015}\natexlab{}.
\newblock \showarticletitle{Human-robot co-navigation using anticipatory
  indicators of human walking motion}. In \bibinfo{booktitle}{\emph{2015 IEEE
  International Conference on Robotics and Automation (ICRA)}}. IEEE,
  \bibinfo{pages}{6183--6190}.
\newblock


\bibitem[\protect\citeauthoryear{Van Den~Berg, Guy, Lin, and Manocha}{Van
  Den~Berg et~al\mbox{.}}{2011}]%
        {van2011reciprocal}
\bibfield{author}{\bibinfo{person}{Jur Van Den~Berg},
  \bibinfo{person}{Stephen~J Guy}, \bibinfo{person}{Ming Lin}, {and}
  \bibinfo{person}{Dinesh Manocha}.} \bibinfo{year}{2011}\natexlab{}.
\newblock \showarticletitle{Reciprocal n-body collision avoidance}.
\newblock In \bibinfo{booktitle}{\emph{Robotics research}}.
  \bibinfo{publisher}{Springer}, \bibinfo{pages}{3--19}.
\newblock


\bibitem[\protect\citeauthoryear{Vasquez, Okal, and Arras}{Vasquez
  et~al\mbox{.}}{2014}]%
        {vasquez2014inverse}
\bibfield{author}{\bibinfo{person}{Dizan Vasquez}, \bibinfo{person}{Billy
  Okal}, {and} \bibinfo{person}{Kai~O Arras}.} \bibinfo{year}{2014}\natexlab{}.
\newblock \showarticletitle{Inverse reinforcement learning algorithms and
  features for robot navigation in crowds: an experimental comparison}. In
  \bibinfo{booktitle}{\emph{2014 IEEE/RSJ International Conference on
  Intelligent Robots and Systems}}. IEEE, \bibinfo{pages}{1341--1346}.
\newblock


\bibitem[\protect\citeauthoryear{Verma and Ranga}{Verma and Ranga}{2021}]%
        {verma2021multi}
\bibfield{author}{\bibinfo{person}{Janardan~Kumar Verma} {and}
  \bibinfo{person}{Virender Ranga}.} \bibinfo{year}{2021}\natexlab{}.
\newblock \showarticletitle{Multi-robot coordination analysis, taxonomy,
  challenges and future scope}.
\newblock \bibinfo{journal}{\emph{Journal of Intelligent \& Robotic Systems}}
  \bibinfo{volume}{102}, \bibinfo{number}{1} (\bibinfo{year}{2021}),
  \bibinfo{pages}{1--36}.
\newblock


\bibitem[\protect\citeauthoryear{Vinciarelli, Pantic, and Bourlard}{Vinciarelli
  et~al\mbox{.}}{2009}]%
        {vinciarelli2009social}
\bibfield{author}{\bibinfo{person}{Alessandro Vinciarelli},
  \bibinfo{person}{Maja Pantic}, {and} \bibinfo{person}{Herv{\'e} Bourlard}.}
  \bibinfo{year}{2009}\natexlab{}.
\newblock \showarticletitle{Social signal processing: Survey of an emerging
  domain}.
\newblock \bibinfo{journal}{\emph{Image and vision computing}}
  \bibinfo{volume}{27}, \bibinfo{number}{12} (\bibinfo{year}{2009}),
  \bibinfo{pages}{1743--1759}.
\newblock


\bibitem[\protect\citeauthoryear{Walter}{Walter}{1950}]%
        {walter1950imitation}
\bibfield{author}{\bibinfo{person}{W~Grey Walter}.}
  \bibinfo{year}{1950}\natexlab{}.
\newblock \showarticletitle{An imitation of life}.
\newblock \bibinfo{journal}{\emph{Scientific american}} \bibinfo{volume}{182},
  \bibinfo{number}{5} (\bibinfo{year}{1950}), \bibinfo{pages}{42--45}.
\newblock


\bibitem[\protect\citeauthoryear{Watanabe, Ikeda, Morales, Shinozawa,
  Miyashita, and Hagita}{Watanabe et~al\mbox{.}}{2015}]%
        {watanabe2015communicating}
\bibfield{author}{\bibinfo{person}{Atsushi Watanabe}, \bibinfo{person}{Tetsushi
  Ikeda}, \bibinfo{person}{Yoichi Morales}, \bibinfo{person}{Kazuhiko
  Shinozawa}, \bibinfo{person}{Takahiro Miyashita}, {and}
  \bibinfo{person}{Norihiro Hagita}.} \bibinfo{year}{2015}\natexlab{}.
\newblock \showarticletitle{Communicating robotic navigational intentions}. In
  \bibinfo{booktitle}{\emph{2015 IEEE/RSJ International Conference on
  Intelligent Robots and Systems (IROS)}}. IEEE, \bibinfo{pages}{5763--5769}.
\newblock


\bibitem[\protect\citeauthoryear{Xiao, Liu, Warnell, and Stone}{Xiao
  et~al\mbox{.}}{2022}]%
        {xiao2020motion}
\bibfield{author}{\bibinfo{person}{Xuesu Xiao}, \bibinfo{person}{Bo Liu},
  \bibinfo{person}{Garrett Warnell}, {and} \bibinfo{person}{Peter Stone}.}
  \bibinfo{year}{2022}\natexlab{}.
\newblock \showarticletitle{Motion planning and control for mobile robot
  navigation using machine learning: a survey}.
\newblock \bibinfo{journal}{\emph{Autonomous Robots}} (\bibinfo{year}{2022}),
  \bibinfo{pages}{1--29}.
\newblock


\bibitem[\protect\citeauthoryear{Yan, Jouandeau, and Cherif}{Yan
  et~al\mbox{.}}{2013}]%
        {yan2013survey}
\bibfield{author}{\bibinfo{person}{Zhi Yan}, \bibinfo{person}{Nicolas
  Jouandeau}, {and} \bibinfo{person}{Arab~Ali Cherif}.}
  \bibinfo{year}{2013}\natexlab{}.
\newblock \showarticletitle{A survey and analysis of multi-robot coordination}.
\newblock \bibinfo{journal}{\emph{International Journal of Advanced Robotic
  Systems}} \bibinfo{volume}{10}, \bibinfo{number}{12} (\bibinfo{year}{2013}),
  \bibinfo{pages}{399}.
\newblock


\bibitem[\protect\citeauthoryear{Yao, Zhang, and Oh}{Yao et~al\mbox{.}}{2019}]%
        {yao2019following}
\bibfield{author}{\bibinfo{person}{Xinjie Yao}, \bibinfo{person}{Ji Zhang},
  {and} \bibinfo{person}{Jean Oh}.} \bibinfo{year}{2019}\natexlab{}.
\newblock \showarticletitle{Following Social Groups: Socially Compliant
  Autonomous Navigation in Dense Crowds}.
\newblock \bibinfo{journal}{\emph{arXiv preprint arXiv:1911.12063}}
  (\bibinfo{year}{2019}).
\newblock


\bibitem[\protect\citeauthoryear{Yedidsion, Deans, Sheehan, Chillara, Hart,
  Stone, and Mooney}{Yedidsion et~al\mbox{.}}{2019}]%
        {yedidsion2019optimal}
\bibfield{author}{\bibinfo{person}{Harel Yedidsion},
  \bibinfo{person}{Jacqueline Deans}, \bibinfo{person}{Connor Sheehan},
  \bibinfo{person}{Mahathi Chillara}, \bibinfo{person}{Justin Hart},
  \bibinfo{person}{Peter Stone}, {and} \bibinfo{person}{Raymond~J Mooney}.}
  \bibinfo{year}{2019}\natexlab{}.
\newblock \showarticletitle{Optimal Use of Verbal Instructions for Multi-robot
  Human Navigation Guidance}. In \bibinfo{booktitle}{\emph{International
  Conference on Social Robotics}}. Springer, \bibinfo{pages}{133--143}.
\newblock


\end{thebibliography}

\end{document}